\makeatletter
\@namedef{ver@everyshi.sty}{}
\makeatother
\documentclass[10pt,twocolumn,letterpaper]{article}

\usepackage[pagenumbers]{cvpr}              %

\usepackage{graphicx}
\usepackage{amsmath}
\usepackage{amssymb}
\usepackage{booktabs}
\usepackage[T1]{fontenc}

\usepackage{epsfig}
\usepackage{pifont}

\usepackage{subcaption}

\usepackage{etoolbox}
\usepackage{tabularx}
\usepackage{enumerate}
\usepackage{array,multirow}
\usepackage{makecell}
\usepackage{duckuments}

\usepackage{tikz}

\usepackage{color}
\usepackage{enumitem}

\definecolor{citecolor}{RGB}{34,139,34}

\usepackage[pagebackref,breaklinks,colorlinks]{hyperref}

\usepackage[capitalize]{cleveref}
\crefname{section}{Sec.}{Secs.}
\Crefname{section}{Section}{Sections}
\Crefname{table}{Table}{Tables}
\crefname{table}{Tab.}{Tabs.}

\definecolor{ubpubColor}{rgb}{0.43, 0.5, 0.5}
\definecolor{backboneColor}{rgb}{0.423, 0.325, 0.365}
\definecolor{fpnColor}{rgb}{0.255, 0.498, 0.416}

\newcommand{\PAR}[1]{\vskip4pt \noindent {\bf #1~}}
\newcommand{\MPAR}[1]{\vskip4pt \noindent {\it #1~}}

\newcommand{\UNPUB}[1]{\textcolor{ubpubColor}{#1}}

\newcommand*{\first}[1]{\textcolor{red}{\textbf{#1}}\@\xspace}
\newcommand*{\second}[1]{\textcolor{green}{\textbf{#1}}\@\xspace}
\newcommand*{\third}[1]{\textcolor{blue}{\textbf{#1}}\@\xspace}
\newcommand*{\fourth}[1]{\textcolor{orange}{\textbf{#1}}\@\xspace}
\newcommand*{\fifth}[1]{\textcolor{violet}{\textbf{#1}}\@\xspace}

\newtoggle{comments}
\toggletrue{comments}
\iftoggle{comments}{%
\newcommand{\lau}[1]{\textcolor{magenta}{\textbf{Laura: }{#1}}}
\newcommand{\alj}[1]{{\leavevmode\color{blue}#1}}
\newcommand{\achal}[1]{{\leavevmode\color{orange}[\textbf{Achal:} #1]}}
\newcommand{\cs}[1]{\textcolor{green}{\textbf{CS: }{#1}}}
\newcommand{\deva}[1]{{\leavevmode\color{orange}[\textbf{Deva:} #1]}}
\newcommand{\bigb}[1]{{\leavevmode\color{orange}[\textbf{BigB:} #1]}}
\newcommand{\bastian}[1]{{\leavevmode\color{orange}[\textbf{Bastian:} #1]}}
\newcommand{\laura}[1]{{\leavevmode\color{orange}[\textbf{Laura:} #1]}}
\newcommand{\jono}[1]{{\leavevmode\color{green}[\textbf{Jono:} #1]}}
}{%
\newcommand{\lau}[1]{}
\newcommand{\alj}[1]{}
\newcommand{\achal}[1]{}
\newcommand{\cs}[1]{}
\newcommand{\deva}[1]{}
\newcommand{\bigb}[1]{}
\newcommand{\bastian}[1]{}
\newcommand{\laura}[1]{}
\newcommand{\jono}[1]{}
}
\newcommand{\embd}[1]{$\mathlarger{\mathlarger{\varepsilon}}$}

\newcommand{\cbox}[1]{\tikz[baseline=-0.5ex]\draw[#1, line width=3, ](0,0) -- (0.2, 0);}

\definecolor{scorec}{RGB}{31,119,180} %
\definecolor{objc}{RGB}{255,131,22} %
\definecolor{bgc}{RGB}{56,165,56} %
\definecolor{ensmc}{RGB}{214,39,40} %

 \newcommand\blfootnote[1]{%
  \begingroup
  \renewcommand\thefootnote{}\footnote{#1}%
  \addtocounter{footnote}{-1}%
  \endgroup
}

\newcommand*{\known}{\emph{known}\@\xspace}
\newcommand*{\unknown}{\emph{unknown}\@\xspace}

\def\cvprPaperID{2221} %
\def\confName{CVPR}
\def\confYear{2022}

\begin{document}

\title{\vspace{-1.8em}Opening up Open World Tracking\vspace{-0.6em}}

\author{Yang Liu$^{1,}$\textcolor{red}{\footnotemark[1]}%
\quad
Idil Esen Zulfikar$^{2,}$\textcolor{red}{\footnotemark[1]}%
\quad
Jonathon Luiten$^{2,3,}$\textcolor{red}{\footnotemark[1]}%
\quad
Achal Dave$^{3,}$\textcolor{red}{\footnotemark[1]}%

\\
\quad
Deva Ramanan$^3$
\quad
Bastian Leibe$^2$
\quad
Aljo\u sa O\u sep$^{1,3}$
\quad
Laura Leal-Taix\'{e}$^1$\\
$^1$\small{Technical University of Munich, Germany} \quad $^2$\small{RWTH Aachen University, Germany} \quad $^3$\small{Carnegie Mellon University, USA} \\ 
\footnotesize{$^1${\tt \{yang14.liu, aljosa.osep, leal.taixe\}@tum.de}} \quad \quad \quad
\footnotesize{$^3${\tt \{achald, deva\}@cs.cmu.edu} }\\
\footnotesize{$^2${\tt \{zulfikar, luiten, leibe\}@vision.rwth-aachen.de}} \\
\small{\texttt{\url{openworldtracking.github.io}}}
}

\twocolumn[{%
\renewcommand\twocolumn[1][]{#1}%
\maketitle
\begin{center}
\vspace{-0.8cm}
\captionsetup{type=figure}
\includegraphics[width=0.999\linewidth]{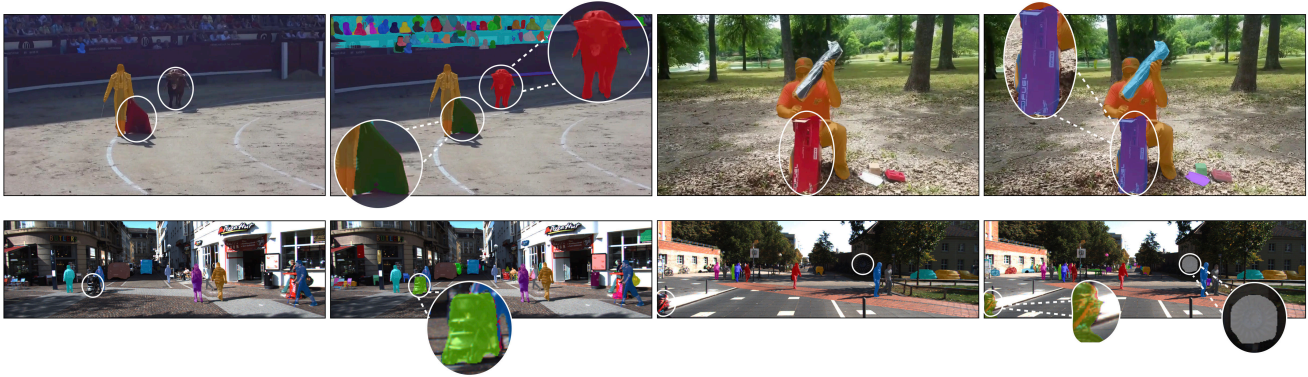}
\vspace{-19pt}
\captionof{figure}{\textit{Each pair left:} The standard approach to multi-object tracking is to detect, track and possibly segment objects that correspond to specific, pre-defined semantic classes, such as cars and pedestrians~\cite{Voigtlaender19CVPR}. \textit{Each pair right:} The output of our tracking baseline, that can track objects, such as child stroller, that was not labeled in the model training set. 
The significant contribution of this paper is the first benchmark, designed for studying the performance of object trackers in such open world conditions, in which trackers are only given a partial knowledge about the visual world, embracing the fact that \textit{one could never train object detectors for every possible semantic class}.}
\label{fig:teaser}
\vspace{-0.1cm}
\end{center}%
}]
\blfootnote{\textcolor{red}{*} These authors contributed equally to this work.}
\blfootnote{Accepted to CVPR 2022 as an Oral Presentation.}

\vspace{-5pt}

\begin{abstract}
\vspace{-5pt}
Tracking and detecting any object, including ones never-seen-before during model training, is a crucial but elusive capability of autonomous systems. An autonomous agent that is blind to never-seen-before objects poses a safety hazard when operating in the real world -- and yet this is how almost all current systems work. One of the main obstacles towards advancing tracking \emph{any} object is that this task is notoriously difficult to evaluate. A benchmark that would allow us to perform an apples-to-apples comparison of existing efforts is a crucial first step towards advancing this important research field.
This paper addresses this evaluation deficit and lays out the landscape and evaluation methodology for detecting and tracking both \textit{known} and \textit{unknown} objects in the open-world setting. We propose a new benchmark, \textit{TAO-OW: Tracking Any Object in an Open World}, analyze existing efforts in multi-object tracking, and construct a baseline for this task while highlighting future challenges. 
We hope to open a new front in multi-object tracking research that will hopefully bring us a step closer to intelligent systems that can operate safely in the real world. 

\end{abstract}

\section{Introduction}
\label{sec:intro}

\noindent \emph{Understanding common scenarios is easy.} Vision systems, trained on millions of examples of cars and pedestrians, work pretty well at detecting these objects, determining what and where they are, and tracking them through a scene.
\noindent \emph{Understanding never-seen-before scenarios is \textbf{extremely} hard.} What happens when a plane lands on the road in front an autonomous vehicle? Or a new children's toy is thrown onto the road? How will current vision systems be able to handle these previously unseen and unknown situations? Will a system designed to detect and track potentially hazardous objects pick up on these at all? Or will they be completely ignored with disastrous consequences (such as a vehicle hitting the \textit{child stroller} in Fig.~\ref{fig:teaser}, \textit{bottom-left})? 

Tracking and detection methods work reasonably well for objects that have a huge amount of data collected on them. But without building systems that can deal with never-seen-before objects, vision systems will never be safe enough to work in the real world and collecting more data can never scale up to address the infinite variety of possible unknown things that can happen.
Many anecdotal examples indicate that current vision systems perform poorly in previously unseen scenarios~\cite{petrovic2020traffic}, but we cannot quantitatively measure this phenomenon, or even evaluate progress, because there are no benchmarks on which to evaluate. 

In this paper we present a new benchmark (\textbf{TAO-OW}: \textit{Tracking Any Object in an Open World}) for measuring detection and tracking performance in an open-world setting. %
Closed-world multi-object tracking benchmarks~\cite{dendorfer20ijcv, Geiger12CVPR, dave20ECCV} and methods~\cite{Leibe08IJCV, Bergmann19ICCV, Voigtlaender19CVPR} focus on tracking object classes that belong to a predefined set of frequently observed classes. 
In contrast, in our Open-World Tracking (OWT) task, \textit{all} object must be tracked, and methods are specifically evaluated on how well they can track object classes that they weren't allowed to train on (\unknown objects), as well as objects which were in the training set (\known objects).

\begin{figure}[t]
\centering
\includegraphics[width=1.0\linewidth]{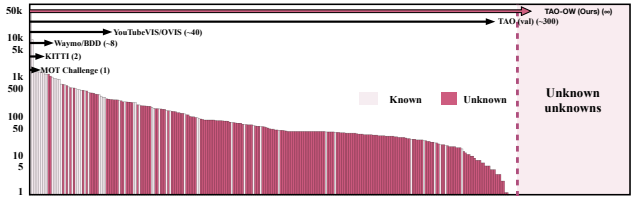}
\vspace{-18pt}
\caption{\textbf{TAO-OW Benchmark} class distribution in the validation set, showing \known classes for which training data is given, and the \unknown classes which serve as a proxy for the infinite variety (\emph{unknown unknowns}) of objects which may appear in an open-world. Note the y-axis is \textbf{log-scaled}.}
\label{fig:the_tail}
\end{figure}

Open-World evaluation is inherently difficult. One has to restrict the set of objects that algorithms are allowed to train on. These \known objects should be varied and diverse enough to represent the set of objects that could typically be expected to have data collected for, but there should be plentiful examples of further \unknown objects, not presented as labeled samples to the models being evaluated. 
We base our work upon the recently introduced TAO dataset~\cite{dave20ECCV}\footnote{License available at \url{taodataset.org}.}, which contains a large corpus of videos from many diverse scenarios such as driving, movies, and everyday scenes. Such a wide diversity is important in order to be able to capture a wide range of \unknown objects. For \known classes we use the $80$ classes from COCO \cite{Lin14ECCV}, which cover a wide range of common objects, while leaving over $700$ \unknown object categories to evaluate the performance of algorithms on objects for which they have not been trained. %
In Fig.~\ref{fig:the_tail} we show our TAO-OW benchmark, with its inherently long tailed distribution of object categories, its \known and \unknown split, and a comparison to previous tracking benchmarks \cite{dendorfer20ijcv,Geiger12CVPR,sun20CVPR,bdd100k,Xu18ECCV,qi2021occluded}, which are all limited to closed-world evaluation on a small number of categories.

Another inherent difficulty with open-world evaluation is dealing with the fact that it is impossible to exhaustively annotate the complete set of objects which should be detected and tracked (by definition, we do not want to penalize trackers for tracking unknown, unannotated objects). To tackle this issue, we propose a new evaluation metric called Open-World Tracking Accuracy (OWTA) which naturally decomposes detection and tracking evaluation components allowing the evaluation of tracking accuracy in the setting where extra unannotated object detections are not penalized. Such evaluation is enabled by the constraint that proposed objects must be supplied as non-overlapping segmentation masks. 

Armed with our Open-World Tracking benchmark and evaluation methodology, we analyze several methods which have attempted this task but have lacked a common evaluation protocol~\cite{Osep18ICRA, Dave19ICCVW, luiten20WACV}. A significant contribution of this paper is our thorough analysis of a wide variety of approaches. 
This analysis leads us to propose an open-world tracking approach which currently performs the best on our Open-World Tracking Benchmark, while also performing very competitively on previous closed-world benchmarks, even though it was not designed or tuned for these.

In summary, the \textbf{main contribution} of this work is to open up a new direction in vision-based multi-object tracking that goes beyond current closed-world benchmarks.
We formalize the Open-World Tracking problem, (i) propose a benchmark with a suitable recall-based evaluation to measure progress, (ii) analyze existing design paradigms, providing a large collection of baselines based on state-of-the-art approaches from the closed-world setting, and (iii) present a strong method which works well for both open- and closed-world tracking.
Our experiments show that closed-world detectors work surprisingly well for \textit{localizing} even unknown objects.
However, \textit{tracking} unknown objects remains more challenging than known objects.

\section{Related Work}

\PAR{Related tasks and benchmarks.} Multi-object tracking (MOT) is a challenging task which involves localizing objects in both space and time, often in dense, crowded environments.
Existing MOT datasets focus on closed-set tracking on video~\cite{Geiger12CVPR,dendorfer20ijcv,Wen15arxiv,Yang19ICCV} or LiDAR streams~\cite{Caesar20CVPR, sun20CVPR}. 
Recent efforts move towards pixel-precise segmentation of tracked objects in video~\cite{Milan15CVPR,Voigtlaender19CVPR, Yang19ICCV, kim20cvpr, weber21arxiv} or LiDAR sequences~\cite{aygun21cvpr}, and study performance in the long tail of object classes~\cite{dave20ECCV}. 
Closer to our work is unsupervised video object segmentation (UVOS)~\cite{Caelles19arXiv} and motion segmentation~\cite{irani1998unified,shi1998motion,bideau2016detailed}, where multiple objects that are present throughout the video and exhibit dominant motion need to be tracked and segmented. %
However, almost all classes in these benchmarks exist in COCO, and almost all methods \cite{luiten20WACV,Athar20eccv} achieve excellent performance by training on COCO. Our work explicitly evaluates on classes \emph{beyond} COCO.

\PAR{Multi-object tracking.}
Early methods in vision-based tracking~\cite{Wren97TPAMI, Paragios00TPAMI, Haritaoglu00TPAMI} and robotic perception~\cite{Teichman11ICRA, Moosmann13ICRA} utilized class-agnostic, bottom-up segmentation as a tracking cue, \eg, based on LiDAR point cloud clustering~\cite{Moosmann09IVS, Wang12ICRA} or background modeling and foreground grouping~\cite{JainNagel79TPAMI, Wren97TPAMI, Stauffer00PAMI}. 
A step forward in vision-based MOT was the \textit{tracking-by-detection} paradigm, which relies on pre-trained object detectors. Early effort focused on developing strong data association techniques~\cite{Pirsiavash11CVPR, Leal11ICCVW, Milan16TPAMI, Brendel11CVPR, Leibe08TPAMI, Reid79TAC} and hand-crafting appearance~\cite{Milan14TPAMI, Mitzel10ECCV, Geiger14TPAMI} and motion cues~\cite{Choi15ICCV, leal14cvpr}. 
More recent efforts are largely data-driven, learning strong appearance models~\cite{LealTaixe16CVPRW, Kim15ICCV}, learning to regress targets~\cite{Bergmann19ICCV} and to associate detections using graph neural networks~\cite{Braso20CVPR}. 
This progress in closed-set multi-object tracking is largely thanks to efforts in releasing new datasets, benchmarks, and evaluation metrics. However, MOT is currently only evaluated in well-controlled, closed-set domains, where object classes are known a priori and are present in training sets.

\PAR{Beyond closed-world tracking.}
Tracking-by-detection approaches have been generalized to generic objects~\cite{Dave19ICCVW, Osep18ICRA, Osep20ICRA} and UVOS~\cite{Dave19ICCVW, luiten20WACV, Luiten18ACCV, xie19cvpr}, using object proposal methods trained in a category-agnostic manner~\cite{Pinheiro16ECCV, He17ICCV}.
However, until recently, there was no suitable evaluation methodology for the open-world domain, making it unclear how such methods generalized to arbitrary objects. 

Recent \textbf{parallel work}~\cite{wang2021unidentified} focuses on labeling a variety of object classes in human-centric Kinetics400 dataset~\cite{kay2017kinetics}. 
This work focuses on data collection and proposes to use the existing closed-world Track mAP~\cite{Yang19ICCV} metric for evaluation. This metric has recently been heavily criticized~\cite{luiten20ijcv} due poor interpretability, lack of sensitivity and a lack of error-type differentiability, which are especially problematic for evaluating tracking in the open world. Furthermore, by default, this metric requires exhaustively labeling all objects, which is infeasible in practice.
The data is also limited to human-centric activities. In contrast, we study open-world tracking in a significantly more diverse setting including videos from multiple different domains, which is crucial for studying open-world problems, resulting in less bias and more generalization (\eg avoiding that objects always appear in the center of frames).
Finally, we analyze prior work on open-world tracking and identify building blocks of these methods to perform a thorough evaluation of these efforts and devise a new baseline, shown to work very well in both, open- and closed-world conditions.

\PAR{Open-set recognition, detection and segmentation.}
Open-set recognition methods~\cite{scheirer12PAMI, scheirer14PAMI, jain14ECCV, bendale16CVPR} focus on minimizing the confusion between \known object classes, presented to the model during the training, and \unknown object classes, that may (only) appear in the open world. Object detection has recently been studied in open-set conditions \cite{miller18ICRA, dhamija18boult}. %
By contrast, open-world recognition methods, as defined by~\cite{bendale15CVPR, liu19CVPR} must explicitly recognize \unknown object instances that were not observed during training, and update object detectors to recognize these unknown instances. Learning to detect \unknown objects in automotive scenarios was tackled in~\cite{osep18ECCVW}, where object detectors were re-trained using clusters of \unknown object tracks~\cite{Osep18ICRA, Osep19ICRA}, mined from video. Similarly, \cite{hwang21CVPR} learns to detect \unknown object instances by sampling and clustering object proposals from the \textit{void} regions from labeled images and using these clusters as pseudo-labels during model training. 
\textit{Joseph \etal}~\cite{joseph21CVPR} propose an extension to Faster R-CNN~\cite{Ren15NIPS} for distinguishing known/unknown classes by adding a contrastive objective, that maximizes the margin between \known and \unknown objects in feature space. %
Unlike these previous works, we do not study how to minimize the confusion between \known or \unknown semantic classes or tackle incremental learning. We study how well we can identify and track objects from both \known and \unknown classes, and we do not require semantic interpretation of tracked objects. Instead, we advocate for the view that any-object tracking is a fundamental problem that should precede recognition. We see our work as a basis for applying such techniques to the video domain that intelligent agents observe.

\section{Opening up Open World Tracking}
\label{sec:owtdef}

Current trackers are limited to specific object classes, such as people or cars, that are labeled in training datasets (which we refer to as \known objects).
We wish to additionally evaluate trackers on \unknown objects, which were not labeled in the training set.
An open-world tracker must segment and track \textit{all} objects (both \known and \unknown) in videos.
Evaluating trackers in this setting is notoriously challenging.
First, densely labeling every object in a video is prohibitively expensive.
Virtually \textit{no} real-world dataset labels all objects, typically limiting the labeling cost by labeling only a subset of classes (\eg KITTI~\cite{Geiger12CVPR}, MOTChallenge~\cite{dendorfer20ijcv}) or instances (\eg TAO~\cite{dave20ECCV}).
Second, defining a generic but consistent notion of an \textit{object} is difficult~\cite{Alexe12TPAMI}.

We address these two challenges simultaneously by relying on a \textit{recall}-based evaluation, inspired by early work on object proposal evaluation~\cite{endres2010category,Alexe12TPAMI} and also adopted for zero-shot object detection~\cite{Bansal18ECCV} and open-world LiDAR segmentation~\cite{wong20corl}. 
Although a precise definition of an \textit{object} is difficult to specify, people have a general notion of what an object is and can label arbitrary objects in a scene~\cite{Gupta19CVPR}. 
Therefore, we can obtain positive object instances as those on which multiple human annotators reach a consensus that \textit{something is an object}. 
This allows us to measure how many ground truth instances a tracker can \textit{recall}.%

Defining the notion of a false positive (FP) is non-trivial as we can only expect a \textit{subset} of objects to be labeled. 
If we consider unlabeled regions as non-objects (FPs), we may be penalizing the tracking system for tracking regions that could still be considered to be valid objects. 
See Fig.~\ref{fig:non_gt_tracking} for an example of objects not labeled in the TAO~\cite{dave20ECCV} dataset, but correctly tracked by our baseline tracker.

\label{sec:metric}

\newcommand{\mysize}{1.0}
\newcommand{\rulesep}{\unskip\ \vrule\ }
\begin{figure}[t]
    \centering
    \setlength{\fboxsep}{0.32pt}%
    \begin{subfigure}[b]{0.32\linewidth}
        \includegraphics[width=\mysize\linewidth]{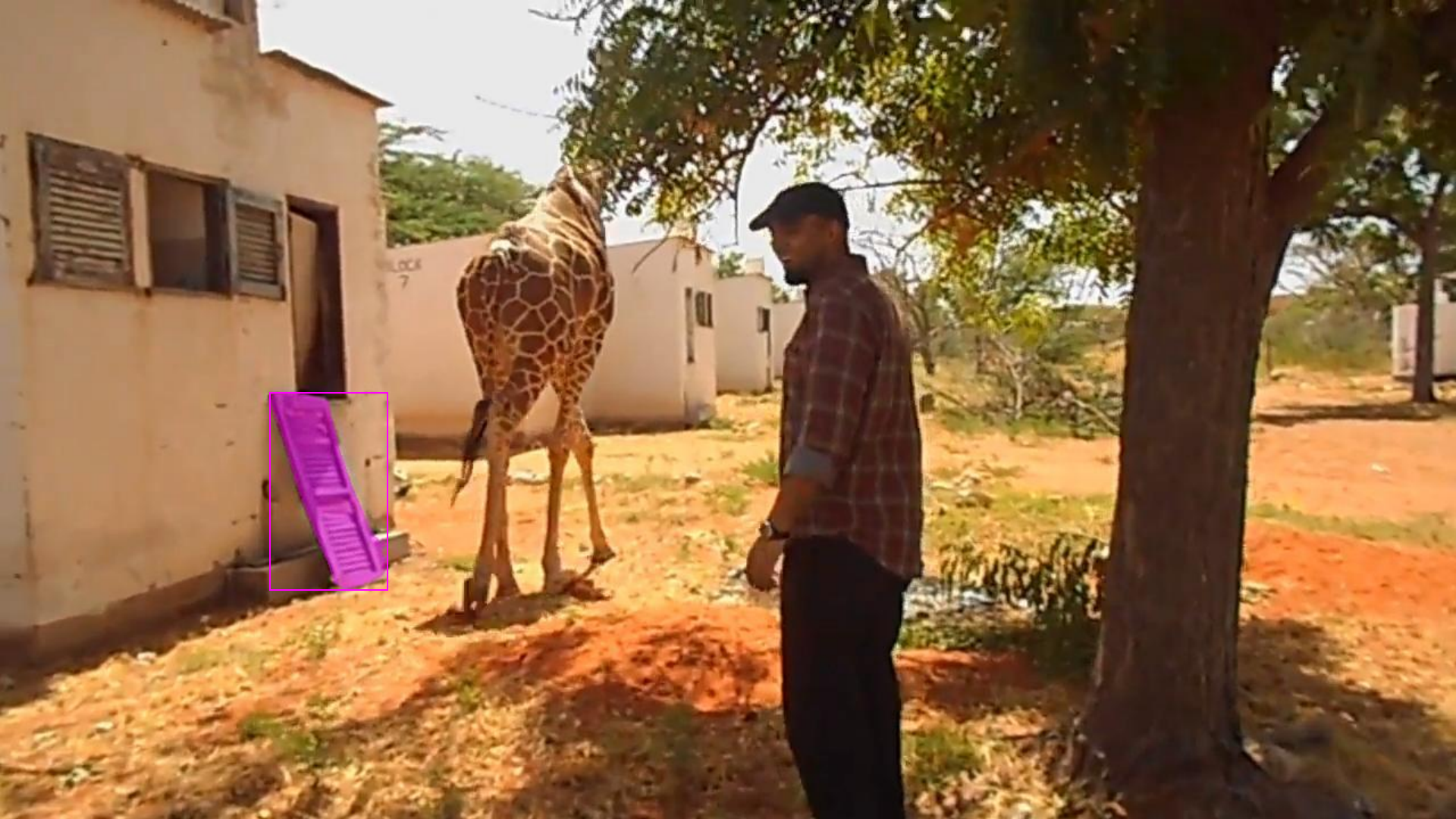}
        \vspace{-16pt}
    \end{subfigure}
    \begin{subfigure}[b]{0.32\linewidth}
        \includegraphics[width=\mysize\linewidth]{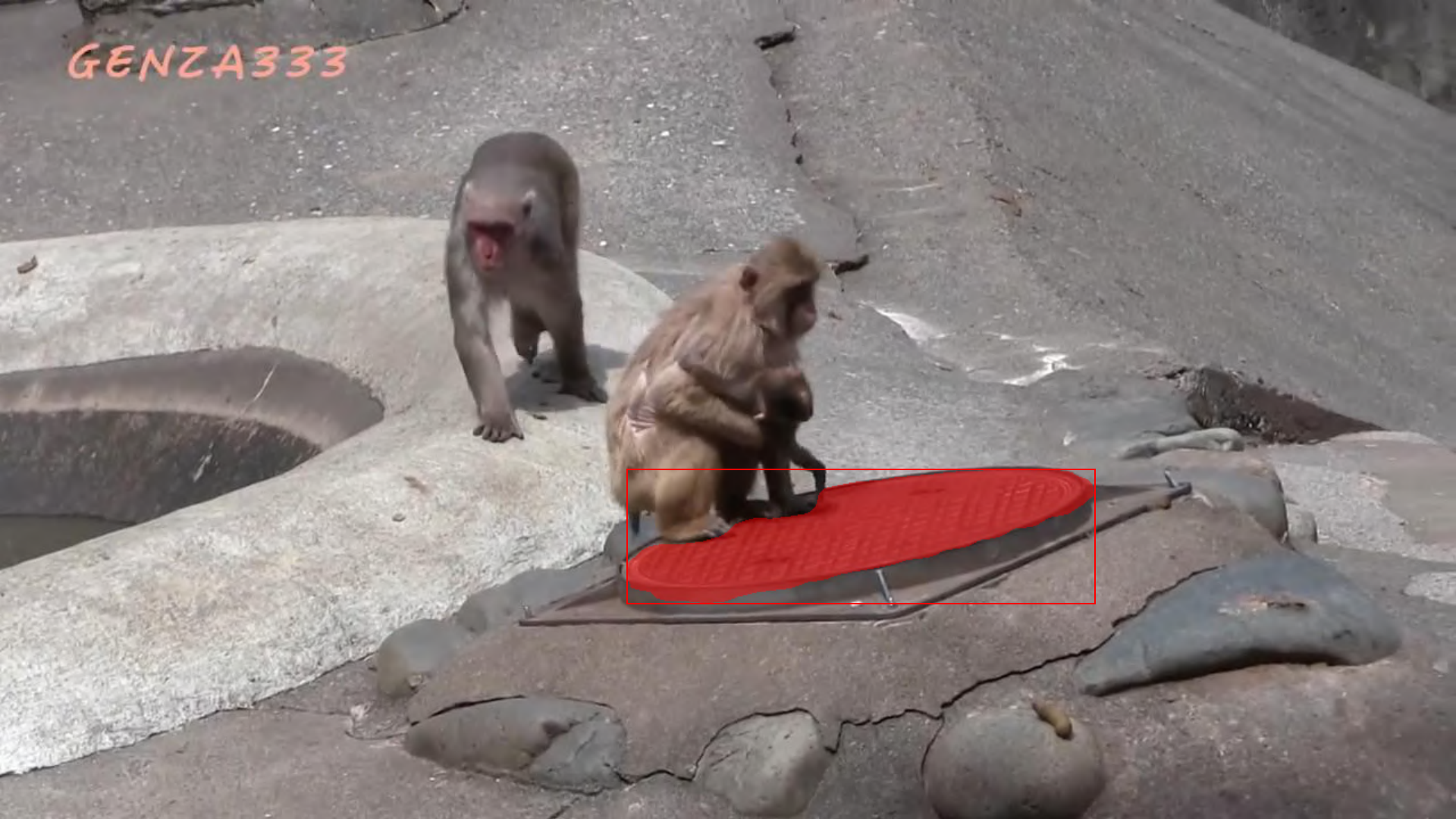}
        \vspace{-16pt}
    \end{subfigure} 
    \begin{subfigure}[b]{0.32\linewidth}
        \includegraphics[width=\mysize\linewidth]{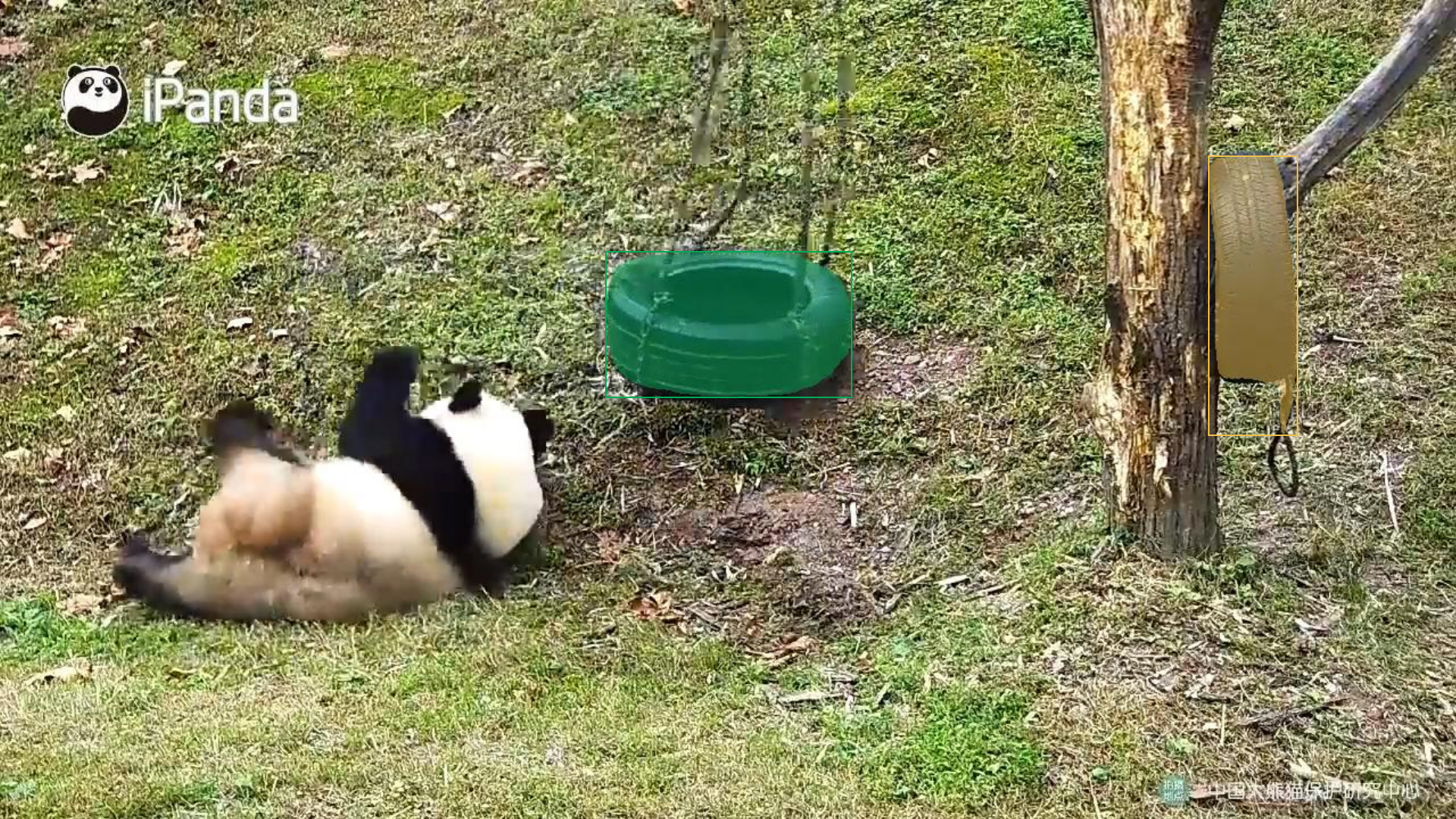}
        \vspace{-16pt}
    \end{subfigure}    
    \vspace{5pt}
	
    \begin{subfigure}[b]{0.32\linewidth}
        \includegraphics[width=\mysize\linewidth]{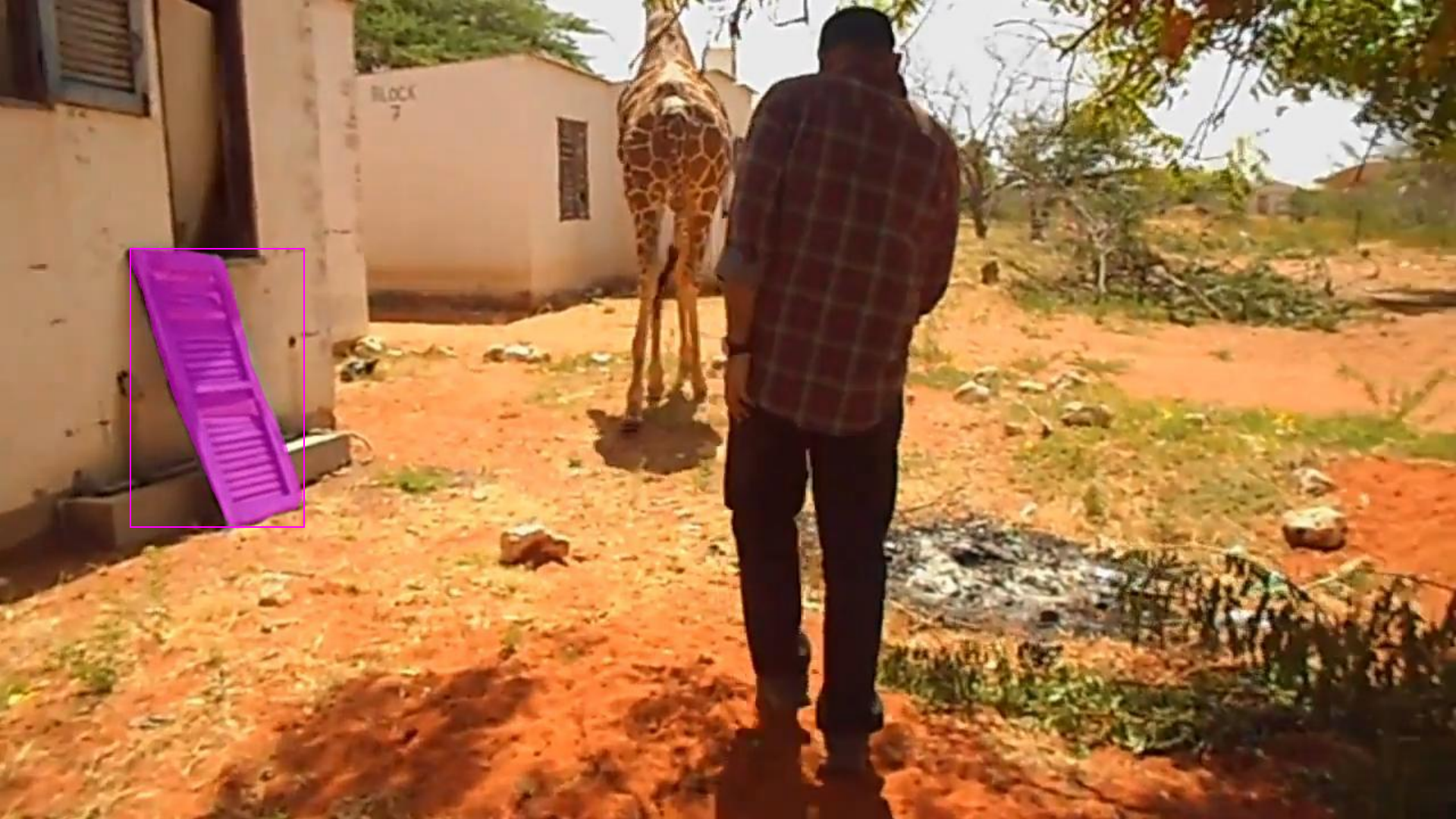}
        \vspace{-16pt}
    \end{subfigure}
        \begin{subfigure}[b]{0.32\linewidth}
        \includegraphics[width=\mysize\linewidth]{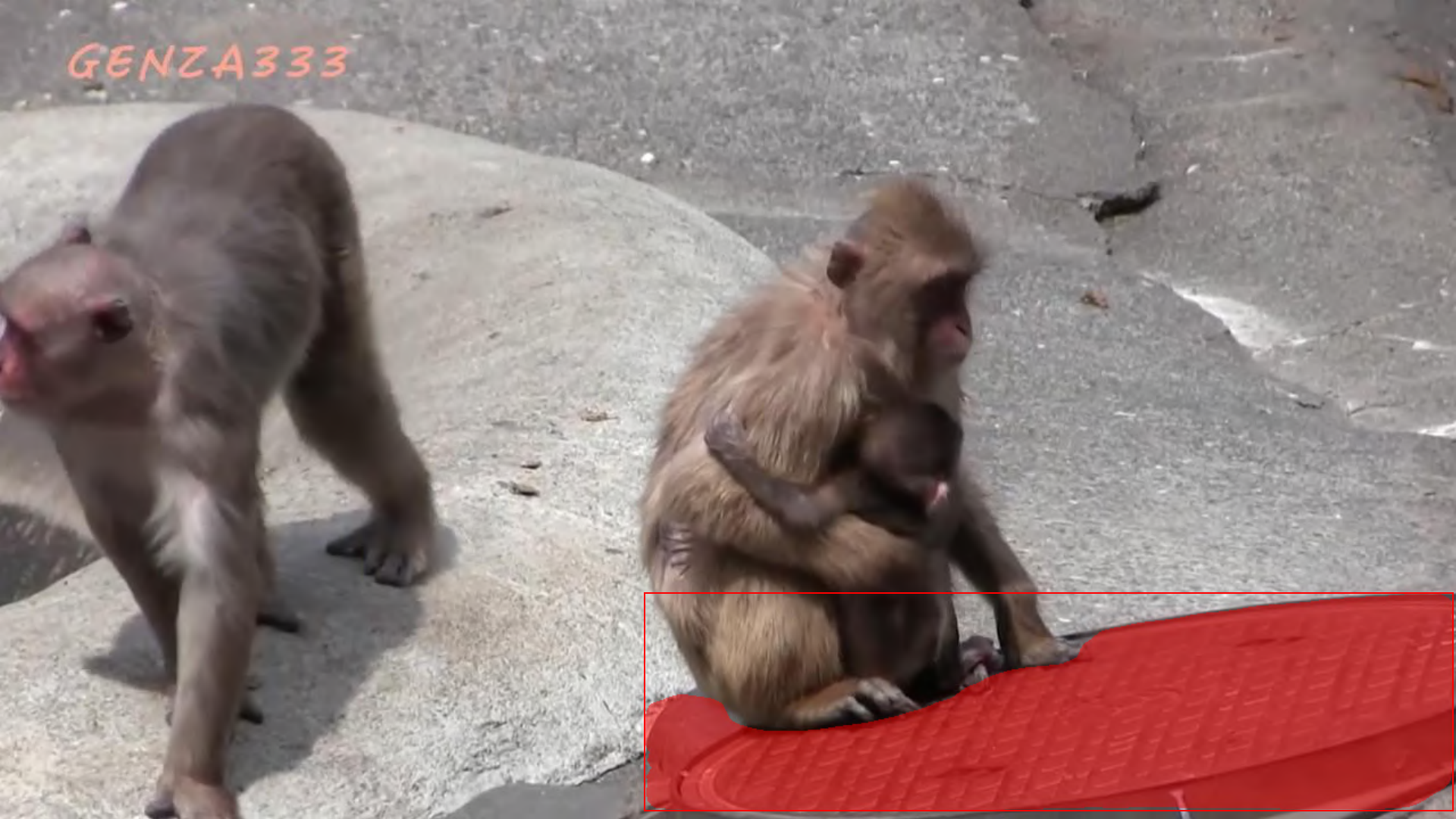}
        \vspace{-16pt}
    \end{subfigure}
    \begin{subfigure}[b]{0.32\linewidth}
     \includegraphics[width=\mysize\linewidth]{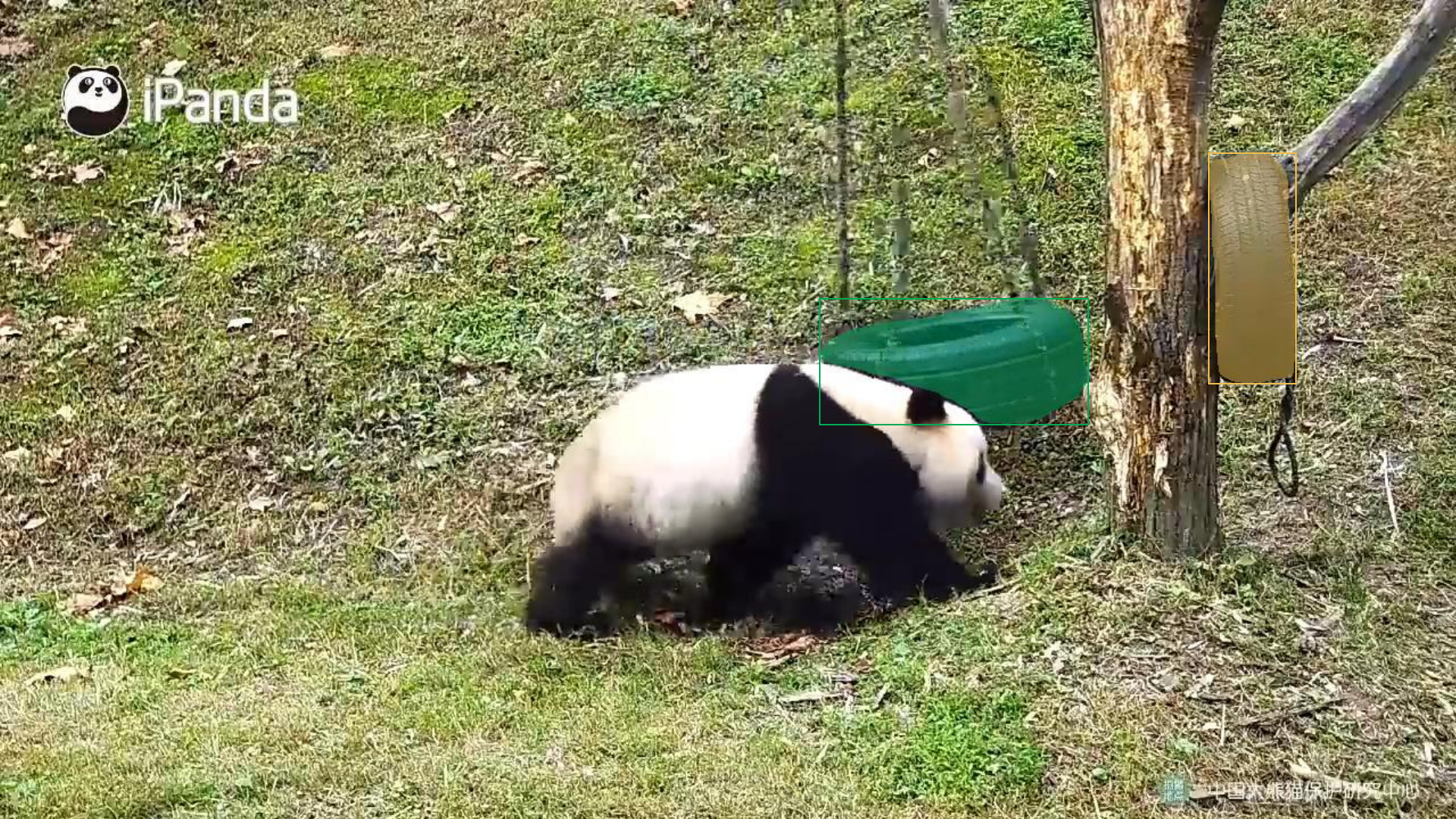}
        \vspace{-16pt}
    \end{subfigure} 
    \vspace{-5pt}
    \caption{\textbf{Unknown unknowns.} Examples of unlabeled objects outside of the TAO~\cite{dave20ECCV} vocabulary which are correctly tracked by our tracker.}
    \label{fig:non_gt_tracking}
\end{figure}

~

\PAR{Open-World Tracking Accuracy (OWTA).} 
We propose the OWTA (Open-World Tracking Accuracy) metric for this task, which is a generalization of the recently proposed HOTA metric \cite{luiten20ijcv} for closed-world tracking. OWTA consists of two intuitive terms, the \textit{association accuracy (AssA)} and \textit{detection recall (DetRe)}. Both terms are evaluated with respect to localization threshold $\alpha$, and the final OWTA metric is integrated over localization thresholds $\alpha$: 
\begin{align*}
\textrm{OWTA}_{\alpha} & = \sqrt{\textrm{DetRe}_{\alpha} \cdot \textrm{AssA}_{\alpha}} \,,\;\;
\textrm{DetRe}_{\alpha}  = \frac{|\textrm{TP}_{\alpha}|}{|\textrm{TP}_{\alpha}| + |\textrm{FN}_{\alpha}|}.
\end{align*}
The recall term \textit{DetRe} does not penalize false positives. This recall-based evaluation is inspired by prior work for evaluating tasks in the open-world such as zero-shot object detection or LiDAR instance segmentation~\cite{Bansal18ECCV, wong20corl}. 

The association accuracy \textit{AssA} term was recently introduced in~\cite{luiten20ijcv}. It measures the number of frames in which the predicted track overlaps with the matched ground truth track. 
For each true positive detection in a predicted track $p_t$ which is matched to a ground truth track $g_t$, \textit{AssA} computes the number of TP associations (TPA, detections in $p_t$ which overlap with $g_t$), FP associations (FPA, detections in $p_t$ which do not overlap with $g_t$), and FN associations (FNA, ground truth annotations in $g_t$ which do not overlap with $p_t$).
\textit{AssA} is evaluated as intersection-over-union over TPA, FPA and FNA sets, and averaged over TPs:
$$
\begin{aligned}
\textrm{AssA}_{\alpha} &= \frac{1}{|\textrm{TP}_{\alpha}|} \sum\limits_{\scriptsize{c \, \in \, \textrm{TP}_{\alpha}}}\frac{\textrm{TPA}_{\alpha}(c)}{\textrm{TPA}_{\alpha}(c) + \textrm{FPA}_{\alpha}(c) + \textrm{FNA}_{\alpha}(c)}.\\
\end{aligned}
$$
The adoption of this \textit{association} term is built on the insight that it is class-agnostic and does not require a densely labeled dataset. %
This is possible because the FPA term in \emph{AssA} is not affected by FP tracks that are not matched to ground truth. Such a factorization is not possible with other metrics such as Track mAP \cite{Yang19ICCV} and IDF1 \cite{ristani2016performance}.

Note that at test time, we require methods to output tracks as non-overlapping masks, such that each pixel in each frame must be uniquely assigned to a track or the background. 
Thus, to achieve high recall a method must correctly group and track pixels over time. A trivial solution that would (theoretically) predict infinitely many tracks is not possible, as the prediction of any track implies that no other track can occupy the same pixels. 
This also aligns our OWT task with the current trend in tracking research to focus on tracking objects at a pixel-accurate segmentation level, and to move away from coarse bounding-box level tracking. Our task can be understood as an Open-World version of MOTS (Multi-Object Tracking and Segmentation~\cite{Voigtlaender19CVPR}) or VIS (Video Instance Segmentation~\cite{Yang19ICCV}).

\section{TAO-OW Benchmark}
\label{sec:owtbench}

Defining a precise and reliable benchmark is critical for enabling progress. Therefore, we propose the Tracking Any Object in an Open World (TAO-OW) benchmark.

\PAR{Dataset.}
Unlike most existing MOT benchmarks~\cite{bdd100k, sun20CVPR, qi2021occluded, Yang19ICCV, Voigtlaender19CVPR}, 
the recently introduced TAO~\cite{dave20ECCV} dataset covers a wide range of classes .
TAO contains almost $3,000$ videos (including $593$ train, $988$ validation and $1,419$ test), comprising $100,000$ annotated frames and $800$ object categories.
Importantly, TAO is annotated without pre-defining object classes: annotators were asked to tag \textit{any objects that move in the video}.
This results in a long-tailed class distribution (see Fig.~\ref{fig:the_tail}), which serves as a proxy for the wide range of objects that could appear in the real world.
If we can build trackers that can track every object in this large video corpus, we can expect them to generalize to a large variety of unconstrained and \emph{open-world} scenarios. 

By default, TAO focuses on a \emph{closed-world} setting, where all classes are defined with examples that are given during training.
We re-purpose this data for the open-world setting, by holding out certain classes from training, while still evaluating on them. We also evaluate on a further $143$ classes which are only present in the test set and not the validation set, which we refer to as \emph{unknown unknowns}.
This enables evaluation in open-world conditions for classes that were not used for validating model parameters. 

\renewcommand{\mysize}{1.0}
\begin{figure}[t]
    \centering
    \setlength{\fboxsep}{0.5pt}%
    \begin{subfigure}[b]{0.49\linewidth}
        \includegraphics[width=\mysize\linewidth]{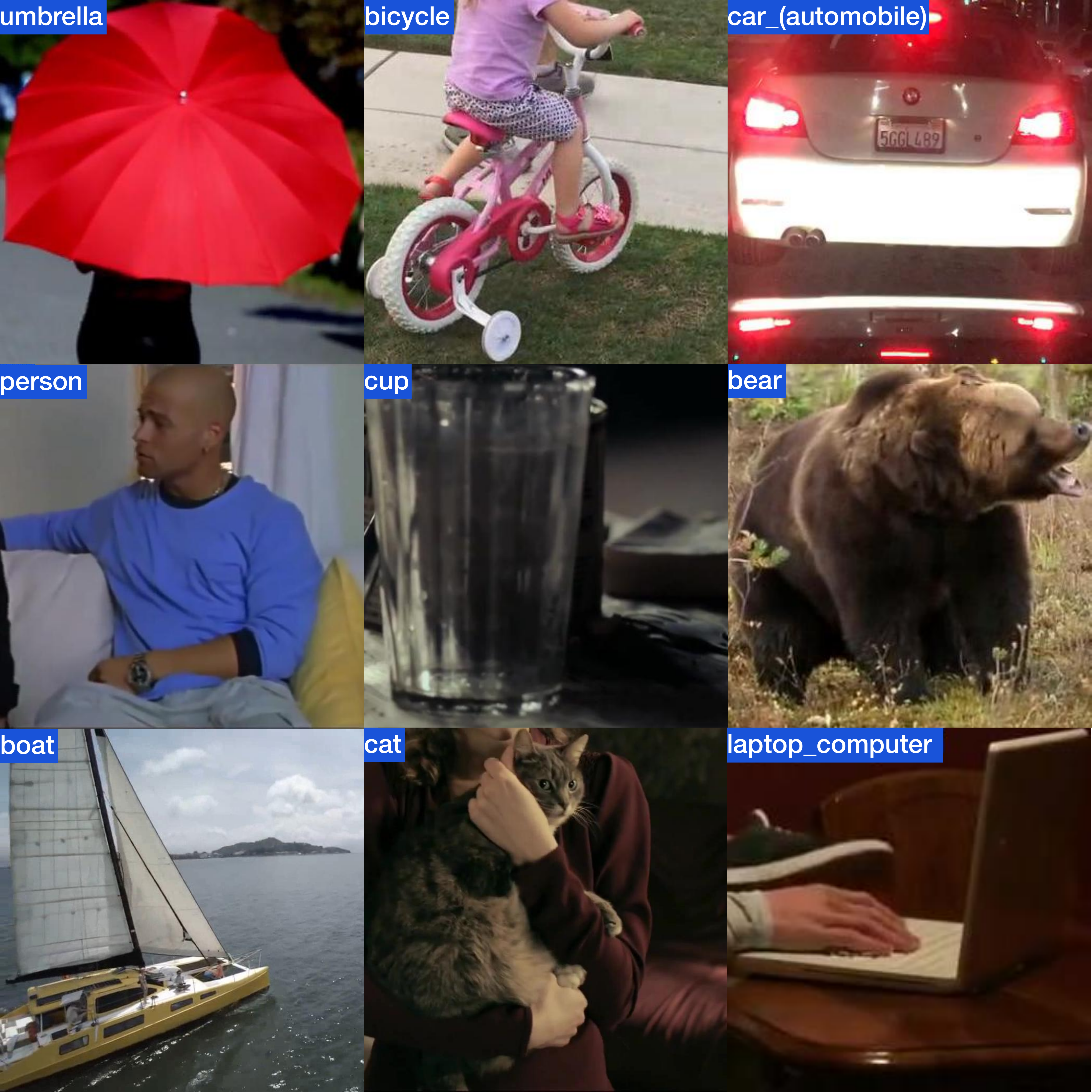}
    \end{subfigure}
    \setlength{\fboxsep}{0.5pt}%
    \begin{subfigure}[b]{0.49\linewidth}
        \includegraphics[width=\mysize\linewidth]{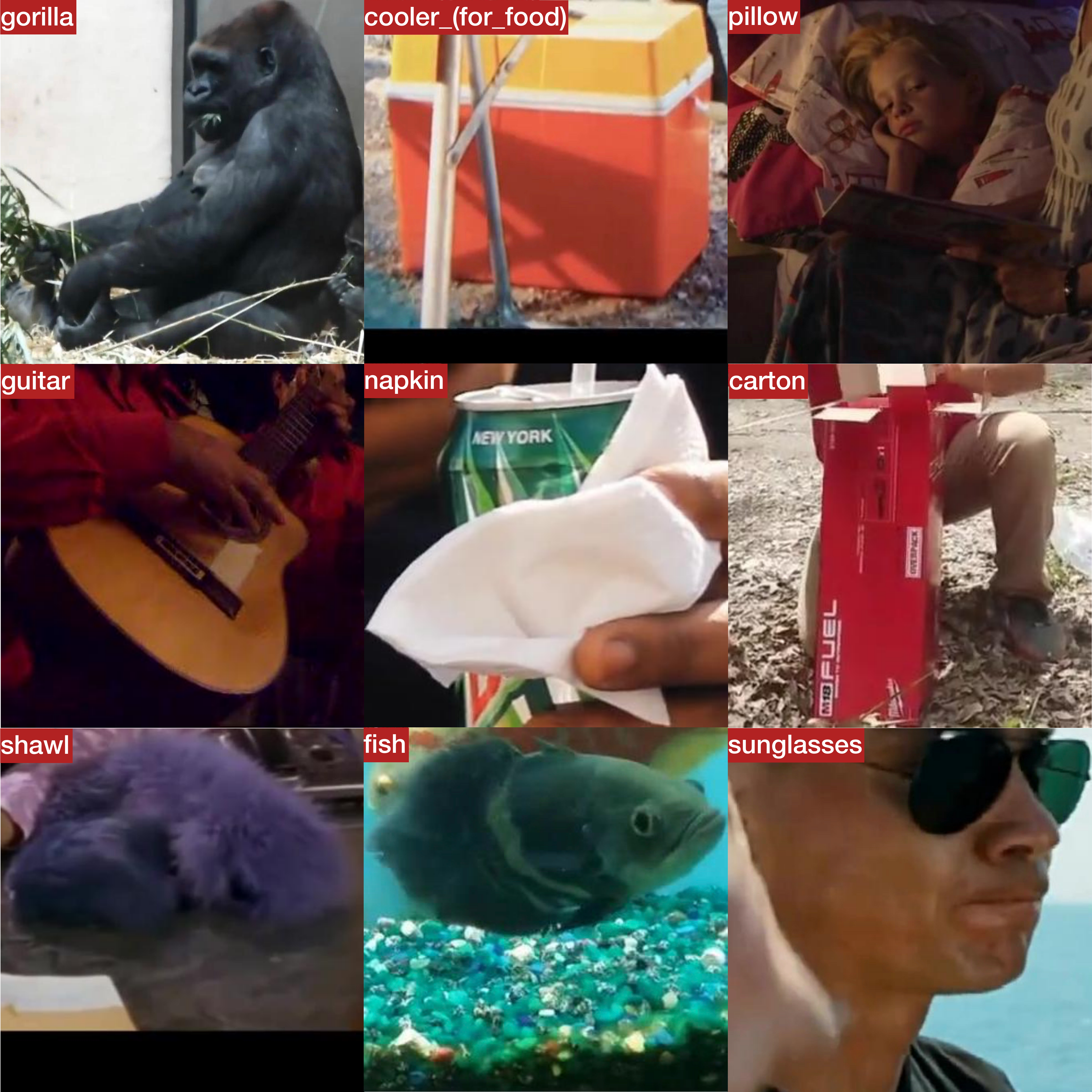}
    \end{subfigure}
    \vspace{-7pt}
    \caption{Examples of \known object categories (left) and \unknown object categories (right).}
    \label{fig:qualitative-tao}
\end{figure}

\PAR{The \known and the \unknown.}
When selecting a set of classes for \known and \unknown sets there are several factors to consider: (i) there should be a large enough and varied enough amount of data covering the \known classes, such that we can train models capable of generalizing to a wider set of classes; (ii) there should be adequate number of \unknown classes remaining to perform a thorough analysis of tracking results for these;  %
and finally (iii), the \known classes should contain classes commonly used in closed-world MOT, so trackers trained for the closed-world can be directly evaluated in the open-world setting. 
Thus, we define classes from the COCO~\cite{Lin14ECCV} dataset as \known, containing 80 common classes, including people, animals, vehicles, hand-held objects and appliances.

\begin{figure}[t]
    \centering
    \includegraphics[width=0.33\linewidth]{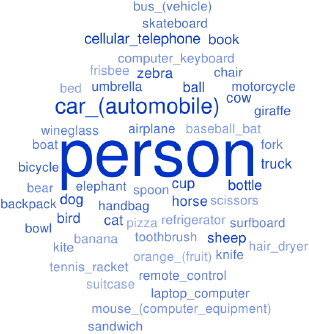}
    \includegraphics[width=0.65\linewidth]{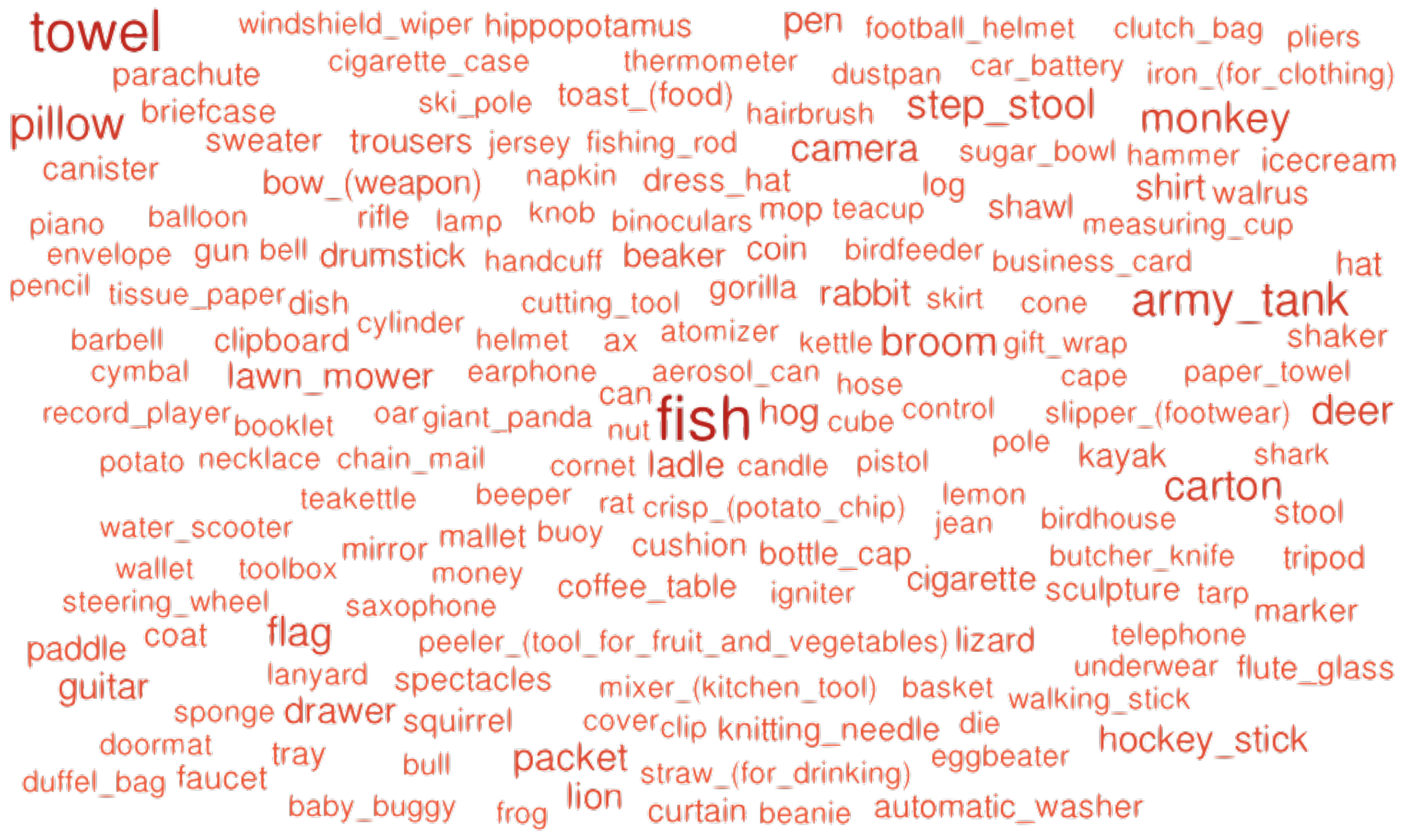}
    \vspace{-5pt}
    \caption{\textbf{TAO-OW classes.} Word cloud showing \textcolor{blue}{\known} (left) and \textcolor{red}{\unknown} (right) classes in our TAO-OW benchmark, with word-size proportional to frequency.}%
    \label{fig:word_clouds}
\end{figure}

TAO validation contains 52 of COCO's 80 classes, with a total of $87,358$ distinct object tracks -- we label these as the \known set of object tracks. 
TAO validation contains an additional 209 classes which do not overlap with COCO, consisting of $20,522$ distinct object tracks.
Of these \unknown object classes not present in the COCO dataset, the most common are \textit{fish}, \textit{towel} and \textit{pillow} with $1274$, $1128$ and $688$ tracks, respectively. 
This \unknown set includes many interesting and worthwhile-to-track classes; some of the authors' favorites include \textit{walrus}, \textit{ice cream}, \textit{drum}, \textit{frog}, \textit{gift wrap} and \textit{binoculars}. Fig.~\ref{fig:qualitative-tao} shows visual examples of videos for both \known and \unknown objects. Fig.~\ref{fig:word_clouds} presents a word cloud of all \known and \unknown objects in the TAO-OW validation set, where the word size is proportional to the number of annotated tracks per class.
To ensure evaluation is not biased by classes similar to \known classes, 
we identify $41$ related classes and mark them as `distractors', as done in closed-world tracking benchmarks~\cite{dendorfer20ijcv, Geiger12CVPR}. These are not used for evaluation. Examples include \textit{cab} (a special case of \textit{car}) and \textit{water bottle} (a special case of \textit{bottle}). We provide details in the supplementary.  

\PAR{Additional considerations.}
TAO is not densely labeled, there are many objects with no annotations. This requires special handling for \emph{closed-set} tracking, where metrics would penalize trackers for correctly predicting unannotated objects.
However, this does not affect OWT, which uses a recall-based OWTA metric (see Sec.\ref{sec:metric}). 
Also, TAO labels objects with bounding boxes, not segmentation masks, while OWT requires methods to produce mask results. Since the ground truth boxes are non-amodal (only cover the visible part of objects), we can evaluate by converting masks to bounding boxes during evaluation. 

\renewcommand{\mysize}{1.0}
\renewcommand{\rulesep}{\unskip\ \vrule\ }

~

\PAR{TAO-OW dataset split.} We provide a train, validation and test split for the TAO-OW dataset, which are adapted from the original TAO dataset. For training we only retain annotations from \known classes and remove all other objects. The validation set contains all objects from TAO but are further labeled as either \known or \unknown depending on if they match a COCO class. The test set contains all objects, with classes which were present in the validation set labeled as \known or \unknown respectively, and remaining classes labeled as \emph{unknown unknowns}. 
Since the \emph{unknown} classes in the validation set can be used to validate design decisions, in order to test models in a truly `hold out' scenario, we require that the test set contains classes which are not present in the validation set. These are the \emph{unknown unknown} classes. Only by such separation, can we consider our test set as a valid proxy for the real open-world, beyond all classes in both the train and validation sets.

\section{Designing Open-World Trackers}

\label{sec:methodology}

No benchmark is complete without well thought through and well-designed baselines. 
The most closely related methods~\cite{Osep18ICRA,Dave19ICCVW,luiten20WACV} are not directly applicable to the TAO-OW domain: \cite{Osep18ICRA} require stereo video, \cite{Dave19ICCVW} assume objects move, while ~\cite{luiten20WACV} assumes that all objects are present in every frame.%
Therefore, a significant contribution of this paper is the analysis of the principles underlying these methods to distill a unified framework for open-world trackers.

\begin{figure}[t]
    \centering
    \includegraphics[width=1.0\linewidth]{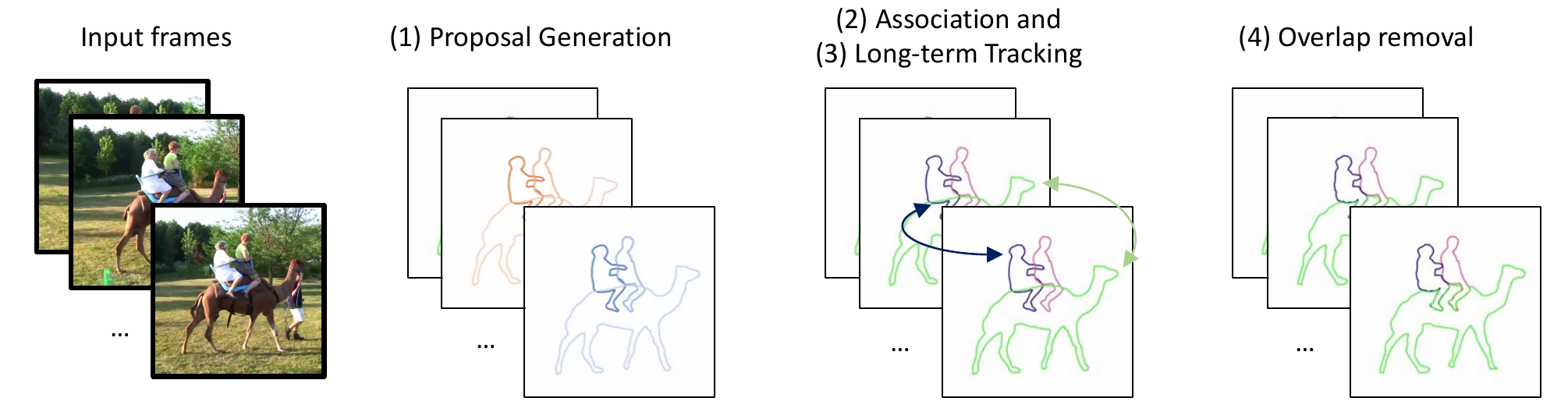}
    \vspace{-20pt}
    \caption{Open-world tracking baseline (OWTB) is inspired by tracking-by-detection pipeline: we (1) obtain object proposals, (2) compute cross-frame association scores, that are used to (3) form and manage tracks, and finally, (4) ensure that conflicts with tracks occupying same space-time volume are resolved.}
    \label{fig:pipeline}
\end{figure}

To devise a strong baseline in such a challenging setting, we first study the anatomy of \textit{tracking-by-detection} (TBD) methodology which has been the dominant MOT approach for years~\cite{dendorfer20ijcv}, and study how it can be adapted for the task of OWT tracking. 
We observe that standard TBD can be decompose into four stages (Fig.~\ref{fig:pipeline}): (1) First, we need to obtain per-image object proposals. This is followed by (2) short-term (cross-frame) proposal similarity estimation, a direct cue for data association; (3) Based on estimated similarities we need to associate proposals and manage tracks, and finally, (4) we need to determine for each pixel a unique track-to-pixel assignment. 
In the following we carefully analyze each stage, using the best-performing decisions as input for later stages to reduce the exponential design space.

\subsection{Proposal Generation (1)} 
\label{sec:exp_proposal_gen}

Following \textit{tracking-by-detection} design we first need to obtain image-level evidence the presence of potential objects. We build on intuition~\cite{Osep18ICRA,Dave19ICCVW, dhamija20WACV} that learned object proposal mechanisms, such as Region Proposal Network~\cite{Ren15NIPS, Pinheiro16ECCV, Pinheiro16NIPS} are explicitly trained to distinguish object-like regions from the \textit{background} and can thus generalize beyond object classes observed in the training set, as already shown in~\cite{Pinheiro16ECCV, Pinheiro16NIPS}. %
We base our analysis on the Mask R-CNN~\cite{He17ICCV} and study how well it generalizes to \textit{unknown} objects. We train our model using labels for 80 classes and first study its performance on TAO-OW's \known and \unknown classes separately. We evaluate this detector as a proposal generator by using a low score threshold and consider the top 1000 proposals output by the model.

\PAR{Proposal recall.} Tab.~\ref{tab:recall_at_1000} shows the recall for both \known and \unknown objects of different sizes when disabling non-maximum suppression and evaluating all $1000$ proposals.
Object sizes are relative to the image size: Large (\emph{ratio} $\geq$ $0.3$), Medium ($0.03$ $\leq$ \emph{ratio} $<$ $0.3$), Small (\emph{ratio} $<$ $0.03$).
The model performs well for large \known and \unknown objects, but significantly worse for small \unknown objects. This indicates that the proposals generalize well to \unknown objects when such objects are large and obvious but are not able to find these objects as well when they are small.

\begin{figure}[t]
    \centering
    \includegraphics[width=0.33\linewidth]{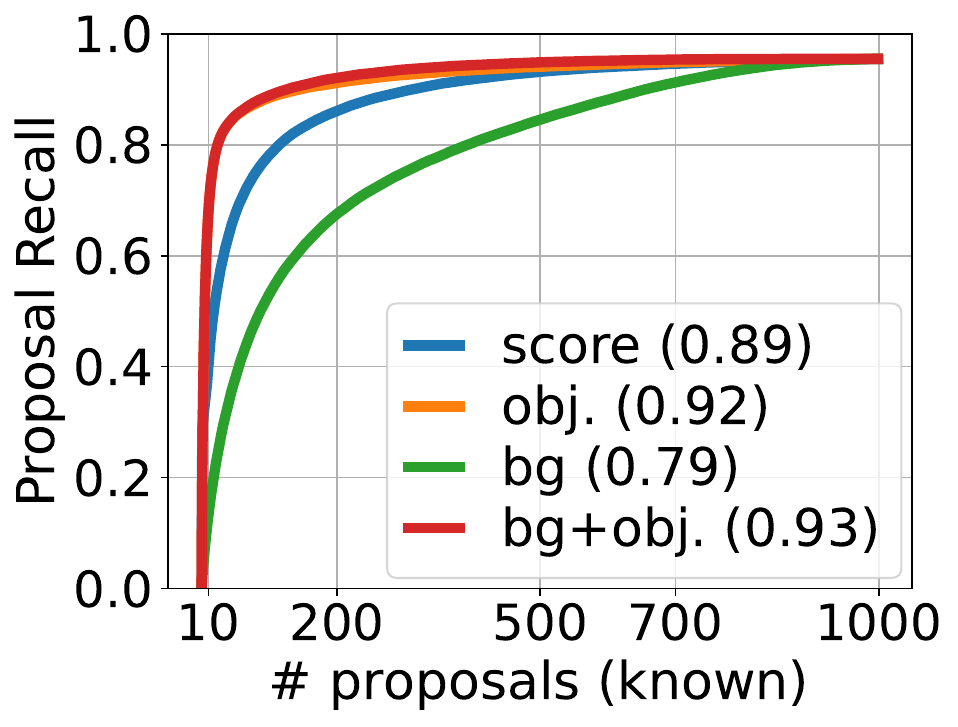}
    \includegraphics[width=0.315\linewidth]{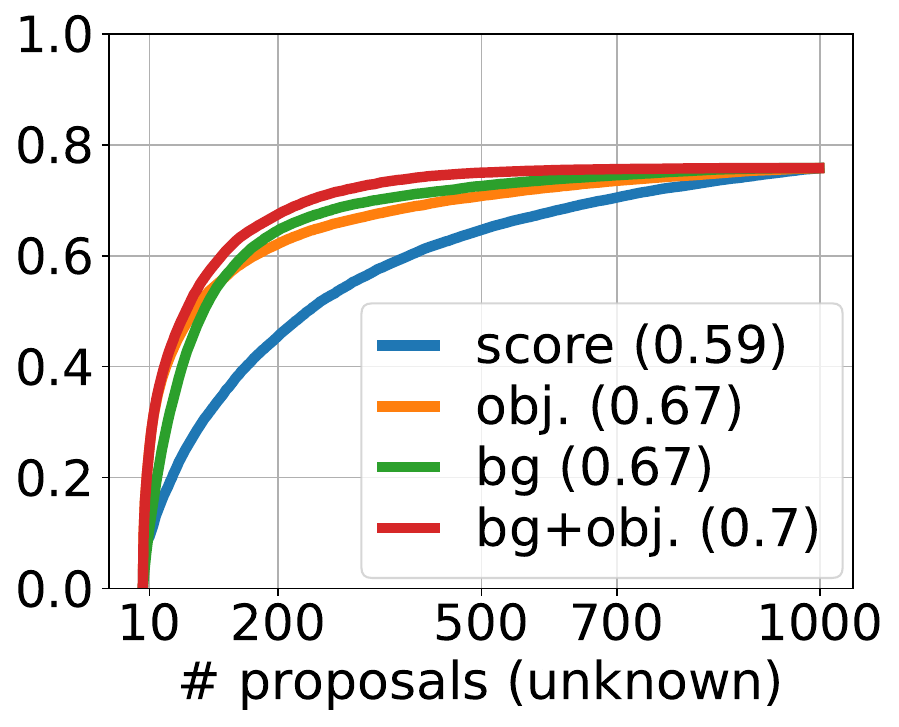}
    \includegraphics[width=0.33\linewidth]{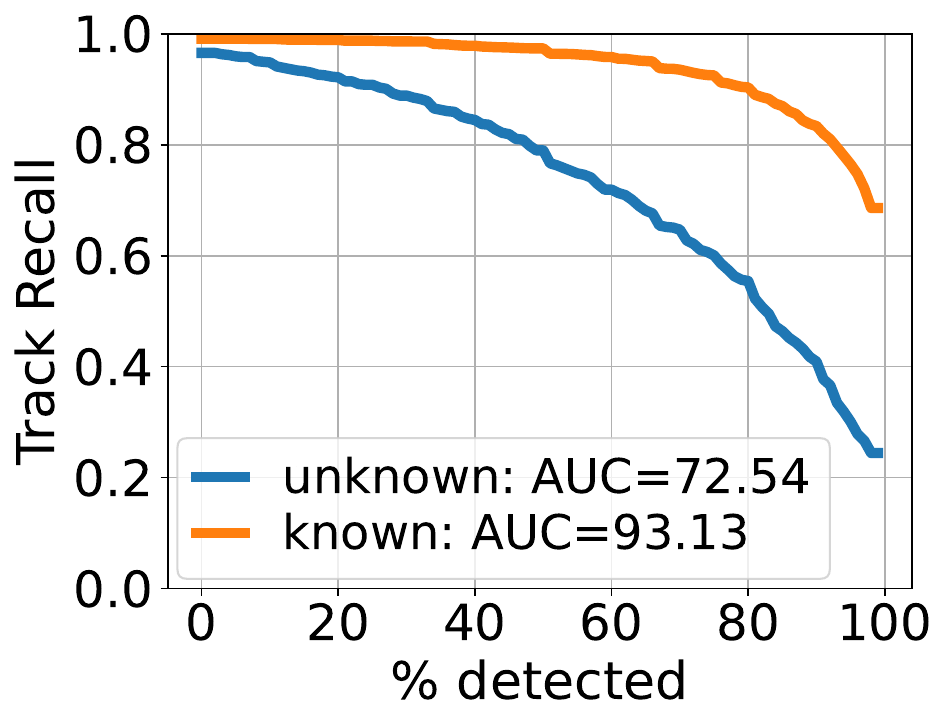}
    \vspace{-20pt}
    \caption{\textbf{Recall Analysis.} Proposal generation recall vs number of proposals for different scoring methods at IoU threshold $0.5$ for \textit{(left)} \known objects and \textit{(center)} \unknown objects. \textit{Right}: Track recall at varying \% objects correctly recalled: \eg, 50\% detected means at least half of the track must be correctly localized.}
    \label{fig:recall}
\end{figure}

Since using all $1000$ proposals as a tracking cue is not feasible, we next investigate how to distinguish \unknown objects from the background clutter. In Fig.~\ref{fig:recall} (left) we show detection recall vs. number of object proposals for several different scoring strategies and display area under the curve. \\
\noindent Fig.~\ref{fig:recall} show that the most confident \known class prediction score (score, \cbox{scorec}) is not a very reliable ranking cue ($0.89$ AUC for \known and $0.59$ AUC for the \unknown). The objectness score (obj., \cbox{objc}) estimated by the RPN provides a significantly better cue ($0.92$ AUC for \known and $0.67$ AUC for the \unknown). 
The background score (bg, \cbox{bgc}), estimated as score for none of the classes, \eg the `$c+1$'th class for a $c$ class detector, is reliable cue for \unknown objects ($0.67$ AUC), but not for \known objects ($0.79 AUC$). 
We obtain most reliable cue by combining the background and objectness scores (obj.+bg, \cbox{ensmc}) using the arithmetic mean ($0.93$ AUC for \known and $0.7$ AUC for the \unknown). We use this scoring function for the remainder of our experiments. 
In conclusion, 2-stage object detectors such as Mask R-CNN generalize quite well to \unknown classes, suggesting they inherently have both an `any object' detector built in (the RPN) and an object vs. non-object classifier. %

\begin{table}[t]
\scriptsize{
\centering
\tabcolsep=0.11cm
    \begin{tabular}{r|c|ccc}
    \toprule
     & Overall & Small size & Medium size & Large size \\\midrule
    \known/\unknown  & \textcolor{blue}{95.4}/\textcolor{red}{75.5} & \textcolor{blue}{91.4}/\textcolor{red}{66.1} & \textcolor{blue}{98.4}/\textcolor{red}{85.9} & \textcolor{blue}{99.7}/\textcolor{red}{98.2}  \\
    \bottomrule
    \end{tabular}
    \vspace{-7pt}
\caption{\textbf{Recall/size Analysis.} Recall for varying object sizes (1k proposals/image).
While models work well for known objects, and \textit{large} unknown objects, they struggle on smaller unknown objects.}
\label{tab:recall_at_1000}
}
\end{table}

\PAR{Track recall.} In addition to proposal recall, we are interested in how well \textit{tracks} are recalled.
Fig.~\ref{fig:recall} (right) shows the percentage of tracks recalled over different minimum relative track lengths. %
Almost every \unknown object ($97\%$) is recalled at least once during its track and over $80\%$ of objects are recalled more than half the time. 
Around $25\%$ of the \unknown objects are recalled in every frame. %

\subsection{Association Similarity (2)} 
Tracking requires estimating proposal \textit{similarity} across frames to maintain object identity.
Since this short-term association based on similarity is critical for accurate long-term tracking, we evaluate it in a controlled setting.
We pose short-term association as a \textit{relative classification} problem: Given a proposal corresponding to a specific query object in frame $t$, how well can the method identify the object among $N$ candidate proposals in frame $t+k$? We set $k$ to correspond to a 1 second gap, and systematically evaluate several different approaches proposed in the community~\cite{Bergmann19ICCV,Osep18ICRA,Dave19ICCVW,luiten20WACV,bewley16icip}. We outline our analysis in Tab.~\ref{tab:proposal_similarity_eval}. Note that all our methods are restricted to training on \known classes and have not seen \unknown classes during training.

\MPAR{Appearance-free.} We start with simple measures that ignore image content, relying only on intersection-over-union (IoU) in the `Appearance-free' block.
This includes `box IoU', `mask IoU', and `generalized IoU` (GIoU)~\cite{rezatofighi19CVPR}.
We evaluate a strategy (`box IoU w/ assoc. thresh') of only propagating through proposals which have box IoU over a threshold ($0.75$) with the previous frame, skipping frames with low quality matches.
We also use a Kalman Filter (KF) to forecast the box in frame $t+k$~\cite{bewley16icip,dave20ECCV,Dave19ICCVW} (`KF, Box IoU'), following parameters from~\cite{bewley16icip}, before computing IoU.

\newcommand{\myhline}{\cmidrule{2-5}}
\newcommand{\cmark}{\ding{51}}%
\newcommand{\xmark}{\ding{55}}%
\begin{table}[t]
\center
\scriptsize{
\tabcolsep=0.02cm
    \resizebox{1.05\columnwidth}{!}{
    \begin{tabular}{@{}c@{}}
    \begin{tabular}[t]{c l c  c  c}
    \toprule
     & Method & Inter. & Known  & Unkn. \\  %
     \cmidrule{1-5}
        \parbox[t]{3mm}{\multirow{8}{*}[-1em]{\rotatebox[origin=c]{90}{\textbf{Appearance-free}}} }
        &\multirow{2}{*}{Box IoU} & \cmark & \fourth{86.4} & 70.7 \\   
        & &  \xmark & 73.2 &  72.6 \\  
        \myhline
        &\multirow{2}{*}{Mask IoU}  & \cmark & 71.6 & 39.5  \\   
        &  &  \xmark &64.4 &  45.3 \\   
        \myhline
        &\multirow{2}{*}{GIoU}  & \cmark & \fourth{86.4} & 70.5 \\   
        &  &  \xmark & 73.0 &   70.9 \\   
        \myhline
        & B. IoU + thresh & \cmark & 81.0 & \fifth{74.7} \\  
        \myhline
        & KF, Box IoU & \cmark & 84.9 &  69.1 \\

        \myhline
        \end{tabular}
        \hspace{0.1cm}
        \begin{tabular}[t]{c l c  c  c}
        \toprule
     & Method & Inter. & Known  & Unkn. \\ 
        
        \cmidrule{1-5}
        \parbox[t]{3mm}{\multirow{3}{*}[-0.3em]{\rotatebox[origin=c]{90}{\textbf{Regress.}}} }
        &\multirow{2}{*}{Regression} & \cmark &\first{88.2} & 65.9  \\
        &  &  \xmark &74.7 & 70.3  \\
        \myhline
        & KF, Regression &  \cmark & \second{87.2} & 65.5  \\
        \myhline
        \cmidrule{1-5}
        \parbox[t]{3mm}{\multirow{7}{*}[-0.8em]{\rotatebox[origin=c]{90}{\textbf{Opt. Flow}}}}      
        & \multirow{2}{*}{Box IoU} & \cmark & \third{87.0} &  67.6  \\   
        & & \xmark& 80.1 &  \second{76.8}  \\   
        \myhline
        & \multirow{2}{*}{Mask IoU} &  \cmark&73.3 & 40.8  \\   
        & & \xmark &68.4 &  47.9  \\
        \myhline
        & GIoU &  \xmark & 80.3 &  \third{75.9} \\   
        \myhline
        &  \multirow{2}{*}{Opt. Fl. + Regr.} &  \cmark&\first{88.2} &  65.9  \\
        & & \xmark& 81.4 &  \fourth{75.3}  \\   
		\end{tabular}
		\hspace{0.1cm}
        \begin{tabular}[t]{c l c  c  c}
        \toprule
        \hspace{0.4cm} & Method & Inter. F & Known  & Unkn. \\ 
		
		\cmidrule{1-5}
        \multirow{6}{*}[-0.6em]{\rotatebox[origin=c]{90}{\textbf{Re-Identification}}}
        & \multirow{2}{*}{MaskRCNN \emph{euclidean}}  & \cmark & 74.3  & 63.5\\   
        & & \xmark &75.3 &  73.3 \\   
		\myhline
		& \multirow{2}{*}{MaskRCNN \emph{cosine}}& \cmark & 73.0 &  64.4  \\   
        & & \xmark &75.0 & 74.2  \\   
        \myhline
        & \UNPUB{PReMVOS \emph{euclidean} *} & \xmark&\UNPUB{82.3\rlap{*}} &  \UNPUB{77.1\rlap{*}}  \\   
        & \UNPUB{PReMVOS \emph{cosine} *} & \xmark &\UNPUB{82.7\rlap{*}} &  \UNPUB{77.5\rlap{*}} \\ 
        \cmidrule{1-5}
		\multirow{2}{*}[0.7em]{\rotatebox[origin=c]{90}{\textbf{Mix}}}    
        & \makecell[l]{Flow-Box IoU  + \\ MaskRCNN \emph{cosine}} & \xmark & \fifth{86.3} & \first{81.9} \\ 

    \myhline
    \end{tabular}
    \\
    \bottomrule
    \end{tabular}
    }
    \vspace{0.1cm}
    \vspace{-10pt}
\caption{\textbf{Association Similarity Ablation.} Top-1 accuracy on 1FPS proposal association classification for various approaches - see text. Best performing methods colored: \first{1st}, \second{2nd}, \third{3rd}, \fourth{4th}, \fifth{5th}. The Inter. column indicates whether `intermediate frames' were used. \UNPUB{*Non open-world oracle (trained on \unknown classes)}}
\label{tab:proposal_similarity_eval}
}
\end{table}

\MPAR{Regression.}
To incorporate appearance information in the motion estimation, we re-purpose an object detector's regressor~\cite{Bergmann19ICCV} to regress the box in frame $t+k$ (`Regression').
We also consider combining this with the KF, by using the KF forecast as input to the regressor (`KF, Regression').
\MPAR{Flow-based.} Next, we use optical flow to estimate proposal motion, following~\cite{luiten20WACV, luiten20RAL}.
We use optical flow to warp a proposal from one frame to the other and use this warped proposal with varying `IoU' criteria. We also use this flow warp as input to the `Regression' approach described above.
\MPAR{Re-Identification.}
We further investigate appearance-based re-identification (ReID) for similarity estimation, ensuring that
the ReID is trained only on \known classes.
We re-purpose the classification layer embedding (1024D) from our detector for the ReID(`MaskRCNN').
We also evaluate a ``non-open-world oracle'' ReID which is not limited to training on \known objects~\cite{Luiten18ACCV} (`PReMVOS').
\MPAR{Intermediate frames.} As TAO is annotated at $1$ FPS, we evaluate in two settings: direct (comparing frames $1$ second apart directly) and continuous, where the similarity is propagated through intermediate frames (\ie, we estimate similarity in one frame, select the most similar proposal, and repeat for all intermediate frames; denoted `Inter. frames'). 
\PAR{Discussion.}
We find that `box IoU' performs well for both \known (86.4) and \unknown (70.7) objects, matching GIoU and outperforming `mask IoU', which is sensitive to occlusions and articulated motion.
Using a regressor trained on \known objects (`Regression') improves association for \known objects (88.2) but degrades for \unknown objects (65.9).
Using a Kalman Filter does not improve accuracy over Box IoU, and matches `Regression' when used with the regressor. 
Using intermediate frames generally improves \known accuracy, but harms \unknown accuracy.
This is because detectors have low recall for \unknown classes, which makes dense propagation prone to drifting when propagating proposals across frames. We add `B. IoU + thresh' entry that skips frames containing low quality matches and increases \unknown accuracy to 74.7.
Optical flow improves the results in almost all cases, and appearance-based ReID with Mask R-CNN features slightly improve results for \known and \unknown.
The `oracle' PReMVOS ReID \cite{Luiten18ACCV} improves \known, but only slightly improves \unknown over flow methods.
The most promising method uses optical flow and box IoU.
We hypothesize that this method may be improved by using Mask R-CNN embeddings and evaluate a simple average of the similarity of these two approaches (`Mix').
This outperforms other approaches for \unknown by a large margin. Not using intermediate frames works well (and is about $30\times$ faster), therefore we ignore intermediate frames for the rest of our analysis. 

\begin{table*}[ht!]
\scriptsize{
\centering
\tabcolsep=0.05cm

    \begin{tabular}{l|c|rcccccccccccccccccc}
    \toprule
    & & & \multicolumn{5}{c}{Known} & & \multicolumn{5}{c}{Unknown} & & \multicolumn{5}{c}{Unknown-Unknown} \\
    & Method & & OWTA & D.Re & A.Acc & A.Re & A.Pr & & OWTA & D.Re & A.Acc & A.Re & A.Pr & & OWTA & D.Re & A.Acc & A.Re & A.Pr\\    %
    \cmidrule{1-2}
    \cmidrule{4-8}
    \cmidrule{10-14}
    \cmidrule{16-20}
    \parbox[t]{3mm}{\multirow{6}{*}{\rotatebox[origin=c]{90}{\textbf{Val}}} }
    & SORT~\cite{bewley16icip} & & 46.6 & 67.4 & 33.7 & 39.7 & 56.4 &  & 33.9 & 43.4 & 30.3 & 34.2 & 57.5 & & -- & -- & -- & -- & --  \\
    & Tracktor~\cite{Bergmann19ICCV} & & 57.9 & 80.2 & 42.6 & 43.6 & 94.4&  & 22.8 & 54.0 & 10.0 & 10.4 & 96.6 & & -- & -- & -- & -- & --  \\  
    & \textbf{OWTB (Ours)} & & 60.2 & 77.2 & 47.4 & 59.1 & 57.9 & & 39.2 & 46.9 & 34.5 & 42.6 & 48.9 & & -- & -- & -- & -- & --  \\
   & \UNPUB{OWTB (w/o N.O.) \textcolor{red}{$^\dagger$}}  & & \UNPUB{60.8} & \UNPUB{82.0} & \UNPUB{45.5} & \UNPUB{57.3} & \UNPUB{56.3} & & \UNPUB{42.4} & \UNPUB{58.9} & \UNPUB{31.5} & \UNPUB{39.5} & \UNPUB{46.8} & & -- & -- & -- & -- & --  \\
    & \UNPUB{AOA\textcolor{red}{*$^\dagger$}~\cite{Du_2020_TAO}} & & \UNPUB{52.8} & \UNPUB{72.5} & \UNPUB{39.1} & \UNPUB{48.8} & \UNPUB{53.6} & &  \UNPUB{49.7} & \UNPUB{74.7} & \UNPUB{33.4} & \UNPUB{41.1} & \UNPUB{51.1}  & & -- & -- & -- & -- & --  \\
    & \UNPUB{SORT-TAO\textcolor{red}{*$^\dagger$}~\cite{dave20ECCV}}  & & \UNPUB{54.2} & \UNPUB{74.0} & \UNPUB{40.6} & \UNPUB{45.0} & \UNPUB{67.3} & & \UNPUB{39.9} & \UNPUB{68.8} & \UNPUB{24.1} & \UNPUB{28.9} & \UNPUB{51.6} & & -- & -- & -- & --  & -- \\
    \hline
    \parbox[t]{3mm}{\multirow{3}{*}{\rotatebox[origin=c]{90}{\textbf{Test}}} }
    & SORT~\cite{bewley16icip} & & 46.6 & 67.1 & 33.7 & 39.5 & 56.0 &  & 32.0 & 42.2 & 26.0 & 30.3 & 53.7 & & 34.3 & 44.7 & 28.2 & 32.5 & 56.5  \\
    & Tracktor~\cite{Bergmann19ICCV} & & 57.9 & 79.7 & 42.9 & 43.9 & 94.5 &  & 23.8 & 53.8 & 11.0 & 11.4 & 96.2 & & 26.3 & 57.9 & 12.4 & 12.8 & 96.2  \\  
    & \textbf{OWTB (Ours)} & & 60.3 & 76.8 & 47.8 & 59.4 & 58.1 &  
    & 38.5 & 45.9 & 33.8 & 42.4 & 49.0 &   
    & 41.5 & 48.9 & 36.5 & 45.4 & 52.3 \\
    \bottomrule
    \end{tabular}
\vspace{-7pt}
\caption{
\textbf{Results of our OWTB on the TAO-OW val. and test set}. We report results in terms of our proposed OWTA metric, and additionally compare methods in terms of Detection Recall (D.Re), Association Accuracy (A.Acc), Association Recall (A.Re) and Association Precision (A.Pr). 
On the val set we compare our final Open-World Tracking Baseline (OWTB) to previous SOTA trackers on TAO-OW val. 
For the test set, \emph{Unknown} classes are the same as those present in the val set, while \emph{Unknown-Unknown} classes are further \unknown classes only present in the test set. \UNPUB{\textcolor{red}{*}: Non open-world (trained on \unknown classes), \textcolor{red}{$^\dagger$}: contains overlapping results.}} 
\label{tab:test_res}
}
\end{table*}

\subsection{Long-term Tracking (3)} 
\label{sec:long-term-track}
After obtaining object proposals and determining a method for calculating similarity between proposals over time, we must now combine all the proposals together into long-term tracks.
We compare  (i) simple Hungarian matching, (ii) Hungarian matching with a keep-alive mechanism to keep tracks alive through occlusions or missing detections~\cite{Dave19ICCVW}, and (iii) UnOVOST~\cite{luiten20WACV} that first builds tracklets using Hungarian matching, and then merges these tracklets in a second offline step. 
We observe that while \textit{keep alive} strategy ($39.7$ OWTA for \unknown) increases association recall over simple Hungarian matching ($39.8$ OWTA for \unknown), however it does so at the loss of the association precision. \textit{Offline tracklet merging} outperforms the two alternative strategies ($40.2$ OWTA for \unknown). We provide detailed results and analysis in the supplementary. %

\subsection{Overlap Removal (4)} 
\label{sec:overlap_removl}
In open-world tracking scenarios we need to rely on object proposals for tracking. Thus we can hypothesize several possible explanations of the observed evidence (\ie, object proposals overlap). However, OWTB tracking task requires unique assignment of pixels in the video to one of objects or background. First strategy (\textit{`non-overlap then track`}) resolves overlaps on the proposal level, and then performs tracking. 
The second approach (\textit{`track then non-overlap`}) follows~\cite{Osep18ICRA} and performs tracking first on the set of (possibly) overlapping proposals. Each track as a whole is then scored using the mean score of each proposal in a track and track suppression is performed within the video volume. 
Intuitively, the second approach should perform better as it can account for temporal context, however, the association problem becomes significantly more complex. We observe that differences between the two approaches are marginal (we provide detailed results and analysis in the supplementary). The simpler \textit{`non-overlap then track`} approach produces slightly better results. This is different to findings reported in~\cite{Osep18ICRA} and where (i) this strategy benefits from depth information and (ii) relies on quadratic pseudo-boolean optimization~\cite{Leibe08TPAMI} that is infeasible to apply at this scale.

\section{Evaluation}
\label{sec:eval_test}
After analyzing several design choices for open-world tracking, we settle on a tracker that uses both optical flow and re-id similarity scoring and combines these into final tracks using tracklet merging. We call our final tracker OWTB (Open-World Tracking Baseline).

Tab.~\ref{tab:test_res} reports the final results of our OWTB tracker on the TAO-OW validation set. First, we compare OWTB to SORT~\cite{bewley16icip} and Tracktor~\cite{Bergmann19ICCV}, both using same set of input proposals as OWTB (see Sec.~\ref{sec:exp_proposal_gen}). As can be seen, OWTB performs significantly better compared to SORT in terms of detection recall ($+9.6$ for \known and $+3.9$ for \unknown), association accuracy ($+13.1$ for \known and $+3.6$ for \unknown) and, consequentially, OWTA ($+13.2$ for \known and $+4.9$ for \unknown). This highlights that (i) a better tracking mechanism leads to a larger number of \unknown objects being tracked, and (ii) that our method produces longer tracks in these challenging scenarios due to a higher association recall. Association precision slightly drops (for \unknown objects) compared to SORT, which is not surprising, as we are tracking a larger number of objects. Tracktor~\cite{Bergmann19ICCV} almost doesn't incorrectly merge objects at all (high A.Pr), but also doesn't merge them correctly very often either (low A.Re), resulting in an overall worse A.Acc.~score than OWTB, especially for \unknown objects. Tracktor, however, gets a boost in D.Re over OWTB because it is able to produce more proposals in each frame than the ones given to OWTB.

As an oracle, we compare to two methods (AOA \cite{Du_2020_TAO} and SORT-TAO \cite{dave20ECCV}) that are state-of-the-art on closed-world TAO.
These comparisons are for reference only, as these methods are trained on \unknown classes, and thus are not open-world trackers.
They also do not produce non-overlapping results.
To analyze the impact of the non-overlapping constraint, we also evaluate OWTB without forcing non-overlaps.
This results in slightly better scores for both \known and \unknown classes, with the detection-recall improving significantly, while the association recall and precision slightly drop.
OWTB performs much better than both previous SOTA closed-world trackers for \known classes due to its strong design.
However, it falls behind for \unknown objects, since both oracle methods have been trained to both detect and track these unknown objects, resulting in higher detection recall and association accuracy.
This highlights the open challenges in open-world tracking.

\PAR{TAO-OW Test-set Evaluation.}
We evaluate and analyze OWTB on the TAO-OW validation set, where we have chosen the best design decisions via tuning. To test our tracker truly in the open-world, we further evaluate OWTB on the TAO-OW test set, which contains both \known, \unknown and \emph{unknown unknown} classes (unknown classes that are not present in the validation set).
Tab.~\ref{tab:test_res} shows that our OWTB performs similarly between the validation and test sets for \known and \unknown objects. Most importantly, our method performs similarly on the set of \emph{unknown unknown} classes (unique to the test set) as it does on classes present in validation. 
We hope our result will be the first of many on the TAO-OW benchmark for Open-World Tracking.

\PAR{OWTB vs. previous closed-world trackers.}
Does a tracker designed for performing well in open-world tracking also work in the traditional closed-world scenario? To test this, we evaluate our OWTB on two previous tracking benchmarks, DAVIS unsupervised~\cite{Caelles19arXiv} and KITTI-MOTS~\cite{Voigtlaender19CVPR}.
We summarize our findings and provide detailed results discussion in the supplementary. For DAVIS we use our own proposal-generation method and rank second (65.5 $\mathcal{J}$ \& $\mathcal{F}$), outperforming several recent methods~\cite{Ventura19CVPR,song2018pyramid,wang2019learning,gowda2020alba,zhou2020matnet,Athar20eccv}, except UnOVOST, all fine-tuned for segmenting dominantly moving regions (67.9 $\mathcal{J}$ \& $\mathcal{F}$). On KITTI-MOTS we use public detections supplied by the benchmark and compare to TrackR-CNN~\cite{Voigtlaender19CVPR} ($56.5/41.9$ HOTA) and recent PointTrack~\cite{xu20ECCV} ($61.9/54.4$ HOTA), that use the same detection set. Our OWTB outperforms both methods for the \textit{car} class ($64.0$ HOTA) and ranks second for the \textit{pedestrian} class ($52.7$ HOTA) 
All other methods are specifically tuned for the benchmarks and the classes, reinforcing the strong generalization capability of our method.

\vspace{-0.1cm}
\section{Conclusion}

With this paper, we hope to light a spark in the heart of the tracking community, by opening up the potential of Open-World Tracking. We have defined Open-World Tracking as a task, motivated its importance, and presented a benchmark and precise evaluation procedure.
We propose a general paradigm for tackling open-world tracking and analyze a wide range of design decisions within this paradigm.
Our analysis results in a tracker that works well for \textit{both} open-world and closed-world tracking. We are excited to announce that Open-World Tracking is open for business.

\footnotesize{\PAR{Acknowledgments.} The authors would like to thank Alexander Hermans and Patrick Dendorfer for their helpful comments on our manuscript. For partial funding of this project IEZ, JL and BL would like to acknowledge the ERC Consolidator Grant DeeViSe (ERC-2017-COG-773161). YL, AO and LLT would like
to acknowledge the Humboldt Foundation through the Sofja Kovalevskaja Award. AD, DR and LLT would like to acknowledge the
CMU Argo AI Center for Autonomous Vehicle Research. We thank Patrick Dendorfer and Mark Weber for help with setting up the benchmark.}

\normalsize

{\small
\bibliographystyle{ieee_fullname}
\bibliography{abbrev_short,refs}

\begin{thebibliography}{10}\itemsep=-1pt

\bibitem{Alexe12TPAMI}
Bogdan Alexe, Thomas Deselaers, and Vittorio Ferrari.
\newblock Measuring the objectness of image windows.
\newblock {\em PAMI}, 34(11):2189--2202, 2012.

\bibitem{Athar20eccv}
Ali Athar, Sabarinath Mahadevan, Aljosa Osep, Laura Leal-Taix{\'e}, and Bastian
  Leibe.
\newblock Stem-seg: Spatio-temporal embeddings for instance segmentation in
  videos.
\newblock In {\em ECCV}, 2020.

\bibitem{aygun21cvpr}
Mehmet Ayg{\"u}n, Aljo{\v{s}}a O{\v{s}}ep, Mark Weber, Maxim Maximov, Cyrill
  Stachniss, Jens Behley, and Laura Leal-Taix{\'e}.
\newblock 4d panoptic lidar segmentation.
\newblock In {\em CVPR}, 2021.

\bibitem{Bansal18ECCV}
Ankan Bansal, Karan Sikka, Gaurav Sharma, Rama Chellappa, and Ajay Divakaran.
\newblock Zero-shot object detection.
\newblock In {\em ECCV}, 2018.

\bibitem{bendale15CVPR}
Abhijit Bendale and Terrance Boult.
\newblock Towards open world recognition.
\newblock In {\em CVPR}, 2015.

\bibitem{bendale16CVPR}
Abhijit Bendale and Terrance~E Boult.
\newblock Towards open set deep networks.
\newblock In {\em CVPR}, 2016.

\bibitem{Bergmann19ICCV}
Philipp Bergmann, Tim Meinhardt, and Laura Leal-Taix{\'e}.
\newblock Tracking without bells and whistles.
\newblock In {\em ICCV}, 2019.

\bibitem{bewley16icip}
Alex Bewley, Zongyuan Ge, Lionel Ott, Fabio Ramos, and Ben Upcroft.
\newblock Simple online and realtime tracking.
\newblock In {\em ICIP}, 2016.

\bibitem{bideau2016detailed}
Pia Bideau and Erik Learned-Miller.
\newblock A detailed rubric for motion segmentation.
\newblock {\em arXiv preprint arXiv:1610.10033}, 2016.

\bibitem{Braso20CVPR}
Guillem Braso and Laura Leal-Taixe.
\newblock Learning a neural solver for multiple object tracking.
\newblock In {\em CVPR}, 2020.

\bibitem{Brendel11CVPR}
William Brendel, Mohamed~R. Amer, and Sinisa Todorovic.
\newblock Multi object tracking as maximum weight independent set.
\newblock {\em CVPR}, 2011.

\bibitem{MPI-Sintel}
D.~J. Butler, J. Wulff, G.~B. Stanley, and M.~J. Black.
\newblock A naturalistic open source movie for optical flow evaluation.
\newblock In {\em ECCV}, 2012.

\bibitem{Caelles19arXiv}
Sergi Caelles, Jordi Pont{-}Tuset, Federico Perazzi, Alberto Montes,
  Kevis{-}Kokitsi Maninis, and Luc~Van Gool.
\newblock The 2019 {DAVIS} challenge on {VOS:} unsupervised multi-object
  segmentation.
\newblock {\em arXiv arXiv:1905.00737}, 2019.

\bibitem{Caesar20CVPR}
Holger Caesar, Varun Bankiti, Alex~H. Lang, Sourabh Vora, Venice~Erin Liong,
  Qiang Xu, Anush Krishnan, Yu Pan, Giancarlo Baldan, and Oscar Beijbom.
\newblock {nuScenes}: A multimodal dataset for autonomous driving.
\newblock In {\em CVPR}, 2020.

\bibitem{Choi15ICCV}
Wongun Choi.
\newblock Near-online multi-target tracking with aggregated local flow
  descriptor.
\newblock In {\em ICCV}, 2015.

\bibitem{dave20ECCV}
Achal Dave, Tarasha Khurana, Pavel Tokmakov, Cordelia Schmid, and Deva Ramanan.
\newblock {TAO}: A large-scale benchmark for tracking any object.
\newblock In {\em ECCV}, 2020.

\bibitem{Dave19ICCVW}
Achal Dave, Pavel Tokmakov, and Deva Ramanan.
\newblock Towards segmenting anything that moves.
\newblock In {\em ICCV Workshops}, 2019.

\bibitem{dendorfer20ijcv}
Patrick Dendorfer, Aljo\v{s}a O\v{s}ep, Anton Milan, Konrad Schindler, Daniel
  Cremers, Ian Reid, and Stefan Roth~Laura Leal-Taix{\'e}.
\newblock Motchallenge: A benchmark for single-camera multiple target tracking.
\newblock {\em IJCV}, 2020.

\bibitem{dhamija20WACV}
Akshay Dhamija, Manuel Gunther, Jonathan Ventura, and Terrance Boult.
\newblock The overlooked elephant of object detection: Open set.
\newblock In {\em WACV}, 2020.

\bibitem{dhamija18boult}
Akshay~Raj Dhamija, Manuel G{\"u}nther, and Terrance~E Boult.
\newblock Reducing network agnostophobia.
\newblock In {\em NeurIPS}, 2018.

\bibitem{Du_2020_TAO}
Fei Du, Bo Xu, Jiasheng Tang, Yuqi Zhang, Fan Wang, and Hao Li.
\newblock 1st place solution to {ECCV-TAO-2020:} detect and represent any
  object for tracking.
\newblock {\em arXiv preprint arXiv: 2101.08040}, 2021.

\bibitem{endres2010category}
Ian Endres and Derek Hoiem.
\newblock Category independent object proposals.
\newblock In {\em ECCV}, 2010.

\bibitem{Geiger14TPAMI}
Andreas Geiger, Martin Lauer, Christian Wojek, Christoph Stiller, and Raquel
  Urtasun.
\newblock 3d traffic scene understanding from movable platforms.
\newblock {\em PAMI}, 1012--1025(36):5, 2014.

\bibitem{Geiger12CVPR}
Andreas Geiger, Philip Lenz, and Raquel Urtasun.
\newblock Are we ready for autonomous driving? the {KITTI} vision benchmark
  suite.
\newblock In {\em CVPR}, 2012.

\bibitem{gowda2020alba}
Shreyank~N Gowda, Panagiotis Eustratiadis, Timothy Hospedales, and Laura
  Sevilla-Lara.
\newblock Alba: Reinforcement learning for video object segmentation.
\newblock {\em arXiv preprint arXiv:2005.13039}, 2020.

\bibitem{Gupta19CVPR}
Agrim Gupta, Piotr Dollar, and Ross Girshick.
\newblock {LVIS}: A dataset for large vocabulary instance segmentation.
\newblock In {\em CVPR}, 2019.

\bibitem{Haritaoglu00TPAMI}
Ismail Haritaoglu, David Harwood, and Larry~S. Davis.
\newblock W4: Real-time surveillance of people and their activities.
\newblock {\em PAMI}, 22:809--830, 2000.

\bibitem{He17ICCV}
K. He, G. Gkioxari, P. Doll{\'a}r, and R. Girshick.
\newblock Mask {R-CNN}.
\newblock In {\em ICCV}, 2017.

\bibitem{hwang21CVPR}
Jaedong Hwang, Seoung~Wug Oh, Joon-Young Lee, and Bohyung Han.
\newblock Exemplar-based open-set panoptic segmentation network.
\newblock In {\em CVPR}, 2021.

\bibitem{irani1998unified}
Michal Irani and P Anandan.
\newblock A unified approach to moving object detection in 2d and 3d scenes.
\newblock {\em TPAMI}, 20(6):577--589, 1998.

\bibitem{jain14ECCV}
Lalit~P Jain, Walter~J Scheirer, and Terrance~E Boult.
\newblock Multi-class open set recognition using probability of inclusion.
\newblock In {\em ECCV}, 2014.

\bibitem{JainNagel79TPAMI}
Ramesh Jain and H.-H Nagel.
\newblock On the analysis of accumulative difference pictures from image
  sequences of real world scenes.
\newblock {\em PAMI}, 1:206 -- 214, 1979.

\bibitem{joseph21CVPR}
K~J Joseph, Salman Khan, Fahad~Shahbaz Khan, and Vineeth~N Balasubramanian.
\newblock Towards open world object detection.
\newblock In {\em CVPR}, 2021.

\bibitem{kay2017kinetics}
Will Kay, Joao Carreira, Karen Simonyan, Brian Zhang, Chloe Hillier, Sudheendra
  Vijayanarasimhan, Fabio Viola, Tim Green, Trevor Back, Paul Natsev, et~al.
\newblock The kinetics human action video dataset.
\newblock {\em arXiv preprint arXiv:1705.06950}, 2017.

\bibitem{Kim15ICCV}
Chanho Kim, Fuxin Li, Arridhana Ciptadi, and James~M. Rehg.
\newblock Multiple hypothesis tracking revisited.
\newblock In {\em ICCV}, 2015.

\bibitem{kim20cvpr}
Dahun Kim, Sanghyun Woo, Joon-Young Lee, and In~So Kweon.
\newblock Video panoptic segmentation.
\newblock In {\em CVPR}, 2020.

\bibitem{LealTaixe16CVPRW}
Laura Leal-Taix{\'e}, Cristian Canton-Ferrer, and Konrad Schindler.
\newblock Learning by tracking: Siamese cnn for robust target association.
\newblock {\em CVPR Workshops}, 2016.

\bibitem{leal14cvpr}
Laura Leal-Taix{\'e}, Michele Fenzi, Alina Kuznetsova, Bodo Rosenhahn, and
  Silvio Savarese.
\newblock Learning an image-based motion context for multiple people tracking.
\newblock In {\em CVPR}, 2014.

\bibitem{Leal11ICCVW}
Laura Leal-Taix{\'e}, Gerard Pons-Moll, and Bodo Rosenhahn.
\newblock Everybody needs somebody: Modeling social and grouping behavior on a
  linear programming multiple people tracker.
\newblock In {\em ICCV Workshops}, 2011.

\bibitem{Leibe08IJCV}
Bastian Leibe, Ale{\v{s}} Leonardis, and Bernt Schiele.
\newblock Robust object detection with interleaved categorization and
  segmentation.
\newblock {\em IJCV}, 77(1-3):259--289, 2008.

\bibitem{Leibe08TPAMI}
Bastian Leibe, Konrad Schindler, Nico Cornelis, and Luc~Van Gool.
\newblock Coupled object detection and tracking from static cameras and moving
  vehicles.
\newblock {\em PAMI}, 30(10):1683--1698, 2008.

\bibitem{Lin14ECCV}
Tsung-Yi Lin, Michael Maire, Serge Belongie, James Hays, Pietro Perona, Deva
  Ramanan, Piotr Doll{\'a}r, and C.~Lawrence Zitnick.
\newblock Microsoft {COCO}: Common objects in context.
\newblock In {\em ECCV}, 2014.

\bibitem{liu19CVPR}
Ziwei Liu, Zhongqi Miao, Xiaohang Zhan, Jiayun Wang, Boqing Gong, and Stella~X
  Yu.
\newblock Large-scale long-tailed recognition in an open world.
\newblock In {\em CVPR}, 2019.

\bibitem{luiten20RAL}
Jonathon Luiten, Tobias Fischer, and Bastian Leibe.
\newblock Track to reconstruct and reconstruct to track.
\newblock {\em RAL}, 5(2):1803--1810, 2020.

\bibitem{luiten20ijcv}
Jonathon Luiten, Aljo\v{s}a O\v{s}ep, Patrick Dendorfer, Philip Torr, Andreas
  Geiger, Laura Leal-Taix{\'e}, and Bastian Leibe.
\newblock Hota: A higher order metric for evaluating multi-object tracking.
\newblock {\em IJCV}, 2020.

\bibitem{Luiten18ACCV}
Jonathon Luiten, Paul Voigtlaender, and Bastian Leibe.
\newblock Premvos: Proposal-generation, refinement and merging for video object
  segmentation.
\newblock In {\em ACCV}, 2018.

\bibitem{luiten20WACV}
Jonathon Luiten, Idil~Esen Zulfikar, and Bastian Leibe.
\newblock {UnOVOST}: Unsupervised offline video object segmentation and
  tracking.
\newblock In {\em WACV}, 2020.

\bibitem{Milan15CVPR}
Anton Milan, Laura Leal-Taix\'{e}, Konrad Schindler, and Ian Reid.
\newblock Joint tracking and segmentation of multiple targets.
\newblock In {\em CVPR}, 2015.

\bibitem{Milan14TPAMI}
Anton Milan, Stefan Roth, and Konrad Schindler.
\newblock Continuous energy minimization for multitarget tracking.
\newblock {\em PAMI}, 36(1):58--72, 2014.

\bibitem{Milan16TPAMI}
Anton Milan, Konrad Schindler, and Stefan Roth.
\newblock Multi-target tracking by discrete-continuous energy minimization.
\newblock {\em PAMI}, 38(10):2054--2068, 2016.

\bibitem{miller18ICRA}
Dimity Miller, Lachlan Nicholson, Feras Dayoub, and Niko S{\"u}nderhauf.
\newblock Dropout sampling for robust object detection in open-set conditions.
\newblock In {\em ICRA}, 2018.

\bibitem{Mitzel10ECCV}
Dennis Mitzel, Esther Horbert, Andreas Ess, and Bastian Leibe.
\newblock Multi-person tracking with sparse detection and continuous
  segmentation.
\newblock In {\em ECCV}, 2010.

\bibitem{Moosmann09IVS}
Frank Moosmann, Oliver Pink, and Christoph Stiller.
\newblock Segmentation of 3d lidar data in non-flat urban environments using a
  local convexity criterion.
\newblock In {\em Intel. Vehicles Symp.}, 2009.

\bibitem{Moosmann13ICRA}
Frank Moosmann and Christoph Stiller.
\newblock Joint self-localization and tracking of generic objects in 3d range
  data.
\newblock In {\em ICRA}, 2013.

\bibitem{osep18ECCVW}
Aljosa Osep, Paul Voigtlaender, Jonathon Luiten, Stefan Breuers, and Bastian
  Leibe.
\newblock Towards large-scale video video object mining.
\newblock In {\em ECCV Workshop on Interactive and Adaptive Learning in an Open
  World}, 2018.

\bibitem{Osep18ICRA}
Aljo\v{s}a O\v{s}ep, Wolfgang Mehner, Paul Voigtlaender, and Bastian Leibe.
\newblock Track, then decide: Category-agnostic vision-based multi-object
  tracking.
\newblock In {\em ICRA}, 2018.

\bibitem{Osep19ICRA}
Aljo\v{s}a O\v{s}ep, Paul Voigtlaender, Jonathon Luiten, Stefan Breuers, and
  Bastian Leibe.
\newblock Large-scale object mining for object discovery from unlabeled video.
\newblock In {\em ICRA}, 2019.

\bibitem{Osep20ICRA}
Aljo\v{s}a O\v{s}ep, Paul Voigtlaender, Mark Weber, Jonathon Luiten, and
  Bastian Leibe.
\newblock 4d generic video object proposals.
\newblock In {\em ICRA}, 2020.

\bibitem{Paragios00TPAMI}
Nikos Paragios and Rachid Deriche.
\newblock Geodesic active contours and level sets for the detection and
  tracking of moving objects.
\newblock {\em PAMI}, 22:266--280, 2000.

\bibitem{petrovic2020traffic}
{\DJ}or{\dj}e Petrovi{\'c}, Radomir Mijailovi{\'c}, and Dalibor
  Pe{\v{s}}i{\'c}.
\newblock Traffic accidents with autonomous vehicles: type of collisions,
  manoeuvres and errors of conventional vehicles’ drivers.
\newblock {\em Transportation research procedia}, 45:161--168, 2020.

\bibitem{Pinheiro16NIPS}
P.O. Pinheiro, R. Collobert, and P. Dollár.
\newblock Learning to segment object candidates.
\newblock In {\em NeurIPS}, 2015.

\bibitem{Pinheiro16ECCV}
P.H.O. Pinheiro, T.{-}Y. Lin, R. Collobert, and P. Doll{\'{a}}r.
\newblock Learning to refine object segments.
\newblock In {\em ECCV}, 2016.

\bibitem{Pirsiavash11CVPR}
Hamed Pirsiavash, Deva Ramanan, and Charles C.Fowlkes.
\newblock Globally-optimal greedy algorithms for tracking a variable number of
  objects.
\newblock In {\em CVPR}, 2011.

\bibitem{qi2021occluded}
Jiyang Qi, Yan Gao, Yao Hu, Xinggang Wang, Xiaoyu Liu, Xiang Bai, Serge
  Belongie, Alan Yuille, Philip Torr, and Song Bai.
\newblock Occluded video instance segmentation.
\newblock {\em arXiv preprint arXiv:2102.01558}, 2021.

\bibitem{Reid79TAC}
Donald~B Reid.
\newblock An algorithm for tracking multiple targets.
\newblock {\em IEEE Trans. Automatic Control}, 24(6):843--854, 1979.

\bibitem{Ren15NIPS}
Shaoqing Ren, Kaiming He, Ross Girshick, and Jian Sun.
\newblock Faster {R-CNN}: Towards real-time object detection with region
  proposal networks.
\newblock In {\em NeurIPS}, 2015.

\bibitem{rezatofighi19CVPR}
Hamid Rezatofighi, Nathan Tsoi, JunYoung Gwak, Amir Sadeghian, Ian Reid, and
  Silvio Savarese.
\newblock Generalized intersection over union: A metric and a loss for bounding
  box regression.
\newblock In {\em CVPR}, 2019.

\bibitem{ristani2016performance}
Ergys Ristani, Francesco Solera, Roger Zou, Rita Cucchiara, and Carlo Tomasi.
\newblock Performance measures and a data set for multi-target, multi-camera
  tracking.
\newblock In {\em ECCV}, 2016.

\bibitem{scheirer12PAMI}
Walter~J Scheirer, Anderson de Rezende~Rocha, Archana Sapkota, and Terrance~E
  Boult.
\newblock Toward open set recognition.
\newblock {\em PAMI}, 35(7):1757--1772, 2012.

\bibitem{scheirer14PAMI}
Walter~J Scheirer, Lalit~P Jain, and Terrance~E Boult.
\newblock Probability models for open set recognition.
\newblock {\em PAMI}, 36(11):2317--2324, 2014.

\bibitem{shi1998motion}
Jianbo Shi and Jitendra Malik.
\newblock Motion segmentation and tracking using normalized cuts.
\newblock In {\em ICCV}, 1998.

\bibitem{song2018pyramid}
Hongmei Song, Wenguan Wang, Sanyuan Zhao, Jianbing Shen, and Kin-Man Lam.
\newblock Pyramid dilated deeper convlstm for video salient object detection.
\newblock In {\em ECCV}, 2018.

\bibitem{Stauffer00PAMI}
Chris Stauffer and W.~Eric~L. Grimson.
\newblock Learning patterns of activity using real-time tracking.
\newblock {\em PAMI}, 22:747--757, 2000.

\bibitem{Sun2018PWC-Net}
Deqing Sun, Xiaodong Yang, Ming-Yu Liu, and Jan Kautz.
\newblock {PWC-Net}: {CNNs} for optical flow using pyramid, warping, and cost
  volume.
\newblock In {\em CVPR}, 2018.

\bibitem{sun20CVPR}
Pei Sun, Henrik Kretzschmar, Xerxes Dotiwalla, Aurelien Chouard, Vijaysai
  Patnaik, Paul Tsui, James Guo, Yin Zhou, Yuning Chai, Benjamin Caine, et~al.
\newblock Scalability in perception for autonomous driving: Waymo open dataset.
\newblock In {\em CVPR}, 2020.

\bibitem{Teichman11ICRA}
Alex Teichman, Jesse Levinson, and Sebastian Thrun.
\newblock Towards {3D} object recognition via classification of arbitrary
  object tracks.
\newblock In {\em ICRA}, 2011.

\bibitem{Ventura19CVPR}
Carles Ventura, Miriam Bellver, Andreu Girbau, Amaia Salvador, Ferran
  Marqu{\'{e}}s, and Xavier {Gir{'{o}} i Nieto}.
\newblock {RVOS:} end-to-end recurrent network for video object segmentation.
\newblock In {\em CVPR}, 2019.

\bibitem{Voigtlaender19CVPR}
Paul Voigtlaender, Michael Krause, Aljosa Osep, Jonathon Luiten, B.B.G Sekar,
  Andreas Geiger, and Bastian Leibe.
\newblock {MOTS}: Multi-object tracking and segmentation.
\newblock In {\em CVPR}, 2019.

\bibitem{Wang12ICRA}
Dominic~Zeng Wang, Ingmar Posner, and Paul Newman.
\newblock {W}hat could move? {F}inding cars, pedestrians and bicyclists in {3D}
  laser data.
\newblock In {\em ICRA}, 2012.

\bibitem{wang2021unidentified}
Weiyao Wang, Matt Feiszli, Heng Wang, and Du Tran.
\newblock Unidentified video objects: A benchmark for dense, open-world
  segmentation.
\newblock {\em arXiv preprint arXiv:2104.04691}, 2021.

\bibitem{wang2019learning}
Wenguan Wang, Hongmei Song, Shuyang Zhao, Jianbing Shen, Sanyuan Zhao,
  Steven~CH Hoi, and Haibin Ling.
\newblock Learning unsupervised video object segmentation through visual
  attention.
\newblock In {\em CVPR}, 2019.

\bibitem{weber21arxiv}
Mark Weber, Jun Xie, Maxwell Collins, Yukun Zhu, Paul Voigtlaender, Hartwig
  Adam, Bradley Green, Andreas Geiger, Bastian Leibe, Daniel Cremers,
  Aljo\u{s}a O\u{s}ep, Laura Leal-Taix\'e, and Chen Liang-Chieh.
\newblock {STEP:} segmenting and tracking every pixel.
\newblock In {\em NeurIPS Track on Datasets and Benchmarks}, 2021.

\bibitem{Wen15arxiv}
Longyin Wen, Dawei Du, Zhaowei Cai, Zhen Lei, Ming-Ching Chang, Honggang Qi,
  Jongwoo Lim, Ming-Hsuan Yang, and Siwei Lyu.
\newblock Ua-detrac: A new benchmark and protocol for multi-object detection
  and tracking.
\newblock {\em arXiv preprint arXiv:1511.04136}, 2015.

\bibitem{wong20corl}
Kelvin Wong, Shenlong Wang, Mengye Ren, Ming Liang, and Raquel Urtasun.
\newblock Identifying unknown instances for autonomous driving.
\newblock In {\em CoRL}, 2020.

\bibitem{Wren97TPAMI}
Christopher~Richard Wren, Ali Azarbayejani, Trevor Darrell, and Alex Pentland.
\newblock Pfinder: Real-time tracking of the human body.
\newblock {\em PAMI}, 19:780--785, 1997.

\bibitem{wu2019detectron2}
Yuxin Wu, Alexander Kirillov, Francisco Massa, Wan-Yen Lo, and Ross Girshick.
\newblock Detectron2.
\newblock \url{https://github.com/facebookresearch/detectron2}, 2019.

\bibitem{xie19cvpr}
Christopher Xie, Yu Xiang, Zaid Harchaoui, and Dieter Fox.
\newblock Object discovery in videos as foreground motion clustering.
\newblock In {\em CVPR}, 2019.

\bibitem{Xu18ECCV}
Ning Xu, Linjie Yang, Yuchen Fan, Jianchao Yang, Dingcheng Yue, Yuchen Liang,
  Brian Price, Scott Cohen, and Thomas Huang.
\newblock {YouTube-VOS}: Sequence-to-sequence video object segmentation.
\newblock In {\em ECCV}, 2018.

\bibitem{xu20ECCV}
Zhenbo Xu, Wei Zhang, Xiao Tan, Wei Yang, Huan Huang, Shilei Wen, Errui Ding,
  and Liusheng Huang.
\newblock Segment as points for efficient online multi-object tracking and
  segmentation.
\newblock In {\em ECCV}, 2020.

\bibitem{Yang19ICCV}
Linjie Yang, Yuchen Fan, and Ning Xu.
\newblock Video instance segmentation.
\newblock In {\em ICCV}, 2019.

\bibitem{bdd100k}
Fisher Yu, Haofeng Chen, Xin Wang, Wenqi Xian, Yingying Chen, Fangchen Liu,
  Vashisht Madhavan, and Trevor Darrell.
\newblock Bdd100k: A diverse driving dataset for heterogeneous multitask
  learning.
\newblock In {\em CVPR}, 2020.

\bibitem{zhou2020matnet}
Tianfei Zhou, Jianwu Li, Shunzhou Wang, Ran Tao, and Jianbing Shen.
\newblock Matnet: Motion-attentive transition network for zero-shot video
  object segmentation.
\newblock {\em IEEE Transactions on Image Processing}, 29:8326--8338, 2020.

\end{thebibliography}
}

\clearpage

\setcounter{page}{1}

\twocolumn[
\centering
\Large
\textbf{Opening up Open World Tracking} \\
\vspace{0.5em}Supplementary Material \\
\vspace{1.0em}
] %
\appendix

\section{Defining Distractor Classes}
\label{sec:distractor}
We observe that some of the objects in TAO are similar to COCO classes, \ie, visually similar to the set of \textit{known} classes, but have different labels. Thus, they cannot be easily separated into \textit{known} and \textit{unknown}. We treat these categories as \textit{distractor} classes and ignore them during the evaluation (similarly as in existing closed-world multi-object tracking datasets, \eg, \cite{Geiger12CVPR,dendorfer20ijcv}). 

\begin{figure}[t]
\begin{center}
\end{center}
\includegraphics[width=1.0\linewidth]{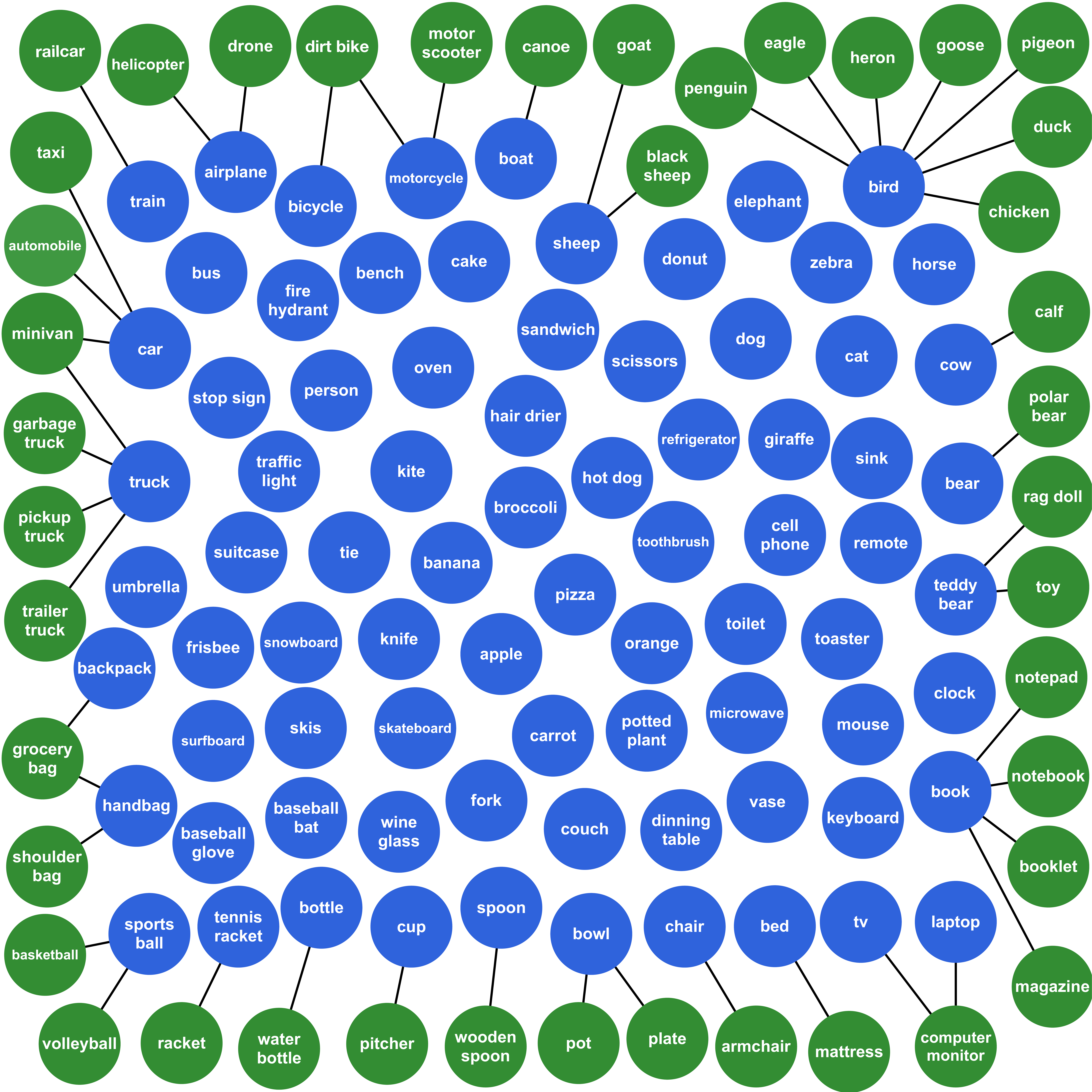}
\caption{\textbf{Known and distractor object classes} The link between the \textcolor{blue}{\textit{known}} (blue) and the \textcolor{green}{\textit{distractor}} (green) indicates  visual similarity.}
\label{fig:known-distractor}
\end{figure}

Different from prior work, finding \textit{distractor} classes from over 800 TAO classes is not an easy task. 
Therefore, we develop a semi-automatic way to determine the \textit{distractor} classes by looking at the COCO-class prediction frequency of a COCO pre-trained Mask R-CNN. 
We run the detector on all frames in the TAO validation set and predict proposals with COCO class labels. 
These COCO class prediction frequencies automatically highlight TAO object classes that are very visually similar to COCO classes. As an example, if a class `minivan' (in TAO vocabulary) is frequently detected as a `car' class (in COCO vocabulary), we tag this class for manual verification. 
This way we mark $42$ classes as distractors, see Fig.~\ref{fig:known-distractor}. To ensure distractor classes do not ``leak`` into the set of \textit{unknown} classes, we assign classes to distractors whenever there is any ambiguity among annotators of whether they are visually or semantically similar to their associated COCO classes.

\section{Implementation Details}
\label{sec:details}

\PAR{Proposal generation} We use a Cascade Mask R-CNN model directly from Detectron 2~\cite{wu2019detectron2}. During the inference, we disable the non-maxima-suppression (NMS) of Mask-RCNN and obtain 1000 proposals per frame. 

\PAR{Association similarity} In the TAO dataset, the videos are annotated at one frame per second (every 30th frame annotated), and we gather pairs of contiguous annotated frames (\textit{current frame} and \textit{next frame}, 1 second apart). We extract matching ground truth objects in each frame-pair and form the set of paired ground truth objects as our evaluation set. As described in Sec. \textcolor{red}{5}, we apply different similarity measures to see if given a proposal that matches with ground truth in the \textit{current frame}, can be successfully associated with a proposal matching with the paired ground truth in \textit{next frame}. To form a reasonable set of paired ground truths, two factors must be considered:  (i) the ground truths with the same ID (in the same track) must be present in both frames; (ii) for each ground truth in \textit{current frame}, there should be at least one proposal that overlaps with it with an IoU $> 0.5$, and the same for \textit{next frame}. By forcing these two constraints, we ensure that the evaluation score will not be affected by unpaired ground truth or missing proposals, and therefore the evaluation can sorely focus on the association ability of various similarity measures. To compare performance, we use top-1 accuracy:

$$
\text{accuracy} = \frac{\text{number of successful associations}}{\text{total number of paired GTs}}
$$

For methods using optical flow in Table \textcolor{red}{3}, we use PWC-Net~\cite{Sun2018PWC-Net} (with fine-tuned weights on MPI Sintel~\cite{MPI-Sintel}) to calculate optical flow vectors. 

\section{Additional Experimental Evaluation Results}
\label{sec:experiments}

In Tab.~\ref{tab:long-term} we outline the results of ablation studies on various long-term tracking and overlap removal strategies on TAO-OW val. These results are discussed in Sec. 5.2 and Sec. 5.3 in the main paper.

In Tab.~\ref{tab:closed-set-sota} we outline results we obtain with our open-world tracking baseline (OWTB). We discuss these results in the Sec.~6 of the main paper. 

\subsection{Long-term tracking} 
After obtaining object proposals and determining a method for calculating the similarity between proposals over time, we combine all the proposals together into long-term tracks.
We investigate various approaches from prior work, excluding certain expensive or complex approaches, such as QBPO optimization~\cite{Osep18ICRA}. 
\cite{Dave19ICCVW} uses Hungarian matching with a keep-alive mechanism to keep tracks alive through occlusions or missing detections.
\cite{luiten20WACV} first builds tracklets using Hungarian matching and then merges these tracklets in a second offline step. 

\begin{table}
\scriptsize{
\center
\tabcolsep=0.11cm
\begin{tabular}{c l c c c c c c c c c c }
\toprule
 & & & \multicolumn{4}{c}{Known} & &  \multicolumn{4}{c}{Unknown}  \\
 & & & OWTA & D.Re & A.Re & A.Pr & & OWTA & D.Re & A.Re & A.Pr\\
\cmidrule{1-2}
\cmidrule{4-7}
\cmidrule{9-12}
    \multirow{3}{*}{\rotatebox[origin=c]{90}{NO$\to$T}}
    & Hung.   & & 60.6 & 78.8 & 56.6 & 61.3 & &  39.8 & 49.7 & 39.7 & 53.4 \\   
    & Hung.+KA & & 60.0  & 78.8 & 57.4 & 57.2 & &  39.7 & 49.7 & 40.7 & 48.9 \\   
    & Hung.+OffTM  & & 60.5 & 78.8 & 58.4 & 57.4 & &  40.2 & 49.7 & 42.0 & 48.6\\   
\cmidrule{1-2}
\cmidrule{4-7}
\cmidrule{9-12}
    \multirow{3}{*}{\rotatebox[origin=c]{90}{T$\to$NO}}
    & Hung.  &  & 59.8 & 78.5 & 55.5 & 60.7 & &  39.8 & 49.6 & 39.5 & 54.1 \\   
    & Hung.+KA & & 59.3 & 78.4 & 56.4 & 56.9 & & 40.0 &  49.6 & 41.0 & 50.1 \\   
    & Hung.+OffTM  & & 59.7 & 78.4 & 57.1 & 57.0 & &  40.1 & 49.6 & 41.5 & 49.6 \\   
\bottomrule
\end{tabular}
\caption{\textbf{Long-term tracking and Overlap removal.} Ablation of various long-term tracking and overlap removal strategies on TAO-OW val. Hung.: Online Hungarian algorithm; KA: Online multi-step keep-alive strategy, OffTM: Offline tracklet merging. NO$\to$T: Non-overlap first, and then track. T$\to$NO: Track first, and then non-overlap.}
\label{tab:long-term}
}
\end{table}

In \Cref{tab:long-term} we compare both of these approaches to long-term tracking, along with just a simple online Hungarian approach which both build upon. In general, the keep-alive strategy performs slightly better than without it, but the offline tracklet merging approach works the best of all. Note, however, that there is only a small difference between all these approaches. Generally, what one approach gains in association-recall, it loses most of in association precision. Successful long-term tracking is still an open challenge in both open-world and closed-world tracking.

\subsection{Overlap removal} 
Finally, we investigate different approaches for removing overlaps between different tracks. This boils down to assigning a score per proposal such that we remove segments of proposals that overlap with any proposal with a higher score. We investigate two approaches for scoring proposals for overlap removal. The first, inspired by \cite{luiten20WACV} and \cite{Dave19ICCVW}, simply takes the per frame proposal score (which we investigated earlier) and uses this for determining which proposals should be given priority to occlude other proposals, such that occluded proposals are made smaller to not overlap. Since this is done at the proposal level, it can be done before long-term tracking takes place. We call this `Non-overlap and then track'. The second approach follows~\cite{Osep18ICRA} and performs tracking first on the set of non-overlapping proposals. Each track as a whole is then scored using the mean score of each proposal in a track. Then overlap removal occurs using these track scores. 
\Cref{tab:long-term} shows that these two approaches generally perform very similarly, though the simpler `non-overlap then track' approach produces slightly better results.

\PAR{OWTB vs. previous closed-world trackers.}
Does a tracker designed for performing well in open-world tracking also work in the traditional closed-world scenario? To test this, we evaluate our OWTB on two previous tracking benchmarks, DAVIS unsupervised~\cite{Caelles19arXiv} and KITTI-MOTS~\cite{Voigtlaender19CVPR}.
We choose DAVIS for video object segmentation, and KITTI-MOTS as it is the most commonly used MOTS benchmark.
For DAVIS we use our own proposal-generation method.
For KITTI-MOTS we use detections supplied by the benchmarks.
\Cref{tab:closed-set-sota} compares our method with prior work.

Despite not being tuned for these datasets, OWTB is competitive on both closed-world benchmarks.
Note all other methods are specifically tuned for these benchmarks and the particular classes in the benchmark, reinforcing the strong generalization capability of our method.

\begin{table}
\footnotesize{
\center
\tabcolsep=0.09cm
\begin{tabular}[t]{l c c c}
\toprule
\\
 DAVIS Unsup. & $\mathcal{J}$ \& $\mathcal{F}$ & $\mathcal{J}$ & $\mathcal{F}$  \\
 \midrule
    RVOS~\cite{Ventura19CVPR} & 41.2 & 36.8	& 45.7 \\
    PDB~\cite{song2018pyramid} & 55.1	& 53.2 & 57.0 \\
    AGS~\cite{wang2019learning} & 57.5& 55.5&59.5\\
    ALBA~\cite{gowda2020alba} & 58.4 & 56.6 & 60.2	\\
MATNet~\cite{zhou2020matnet}  & 58.6 & 56.7 & 60.4\\
    STEm-Seg~\cite{Athar20eccv}  & 64.7  & 61.5 & 67.8 	 \\   
    UnOVOST~\cite{luiten20WACV} & 67.9& 66.4 & 69.3  \\   
    
    \textbf{OWTB (Ours)}  & 65.5 & 63.7 & 67.4 \\   
    \bottomrule
\end{tabular}~\begin{tabular}[t]{lcc}
    \toprule
        & \multicolumn{2}{c}{HOTA} \\
     KITTI-MOTS & car & ped.  \\
    \midrule
    TrackR-CNN~\cite{Voigtlaender19CVPR} & 56.5 & 41.9 \\
    PointTrack~\cite{xu20ECCV}    & 61.9 & 54.4  \\   
    \textbf{OWTB (Ours)}   & 64.0 & 52.7  \\   
\bottomrule
\end{tabular}
\caption{Results of our OWTB on closed-world benchmarks DAVIS Unsupervised (val) and KITTI-MOTS (test), compared to all previous published methods. \textcolor{red}{*}MOTSFusion additionally uses stereo-depth information.}
\label{tab:closed-set-sota}
}
\end{table}

\section{Limitations and Societal Impacts}
\label{sec:limitsandsociety}

\PAR{Limitations of presented baselines.}
With our benchmark, for the first time we are able to evaluate the difficulty of detecting and tracking unknown objects, using our proposed OWTB (Open World Tracking Baseline). From
this we extract the following insights on the limitations of this approach, and believe that this could motivate future research directions:
\setlist{nolistsep,leftmargin=*}
\begin{itemize}
\item \textit{Unknown object detection is significantly harder} (and thus less accurate) than known object detection:
Unknown object detections are often incorrectly grouped into a union of different unknown objects; often only parts of objects are detected instead of the whole; many objects are also simply never recalled at all. Unknown object detection could be improved by taking into account temporal context (multi-frame) for detection, effectively combining elements of tracking and detection.
\item \textit{Low quality detections make tracking much harder}. When detections can be trusted (as with common classes), tracking reduces to identity assignment. When they cannot, the trackers must be robust to missing or partial detections.
\item \textit{Fast moving objects, with large deformation, are very challenging}.
\item \textit{Unknown objects which completely disappear and reappear again are almost never correctly tracked.} Building robust long-term appearance models of previously unseen objects is a key future research direction.
\item \textit{Relying on labeled data alone limits tracking methods.} Our current approach only transfers knowledge from labeled known classes to unknowns. Using unlabeled training data could result in potentially large improvements..
\end{itemize}

\PAR{Societal Impacts.}
Open-world tracking allows operating in a world populated by never-before-seen possibly dynamic obstacles, and learning about semantic concepts with minimal supervision. Unfortunately, object tracking also has privacy and surveillance repercussions, as it can be used for person tracking.
Our work focuses more on the class-agnostic setting, rather than the well-established pedestrian tracking, but could be used for this purpose as well.

\section{Additional Qualitative Results}
\label{sec:qualitative}

\begin{figure}[t]
    \centering
    \setlength{\fboxsep}{1pt}%
    \begin{subfigure}[b]{1\linewidth}
        \includegraphics[width=\mysize\linewidth]{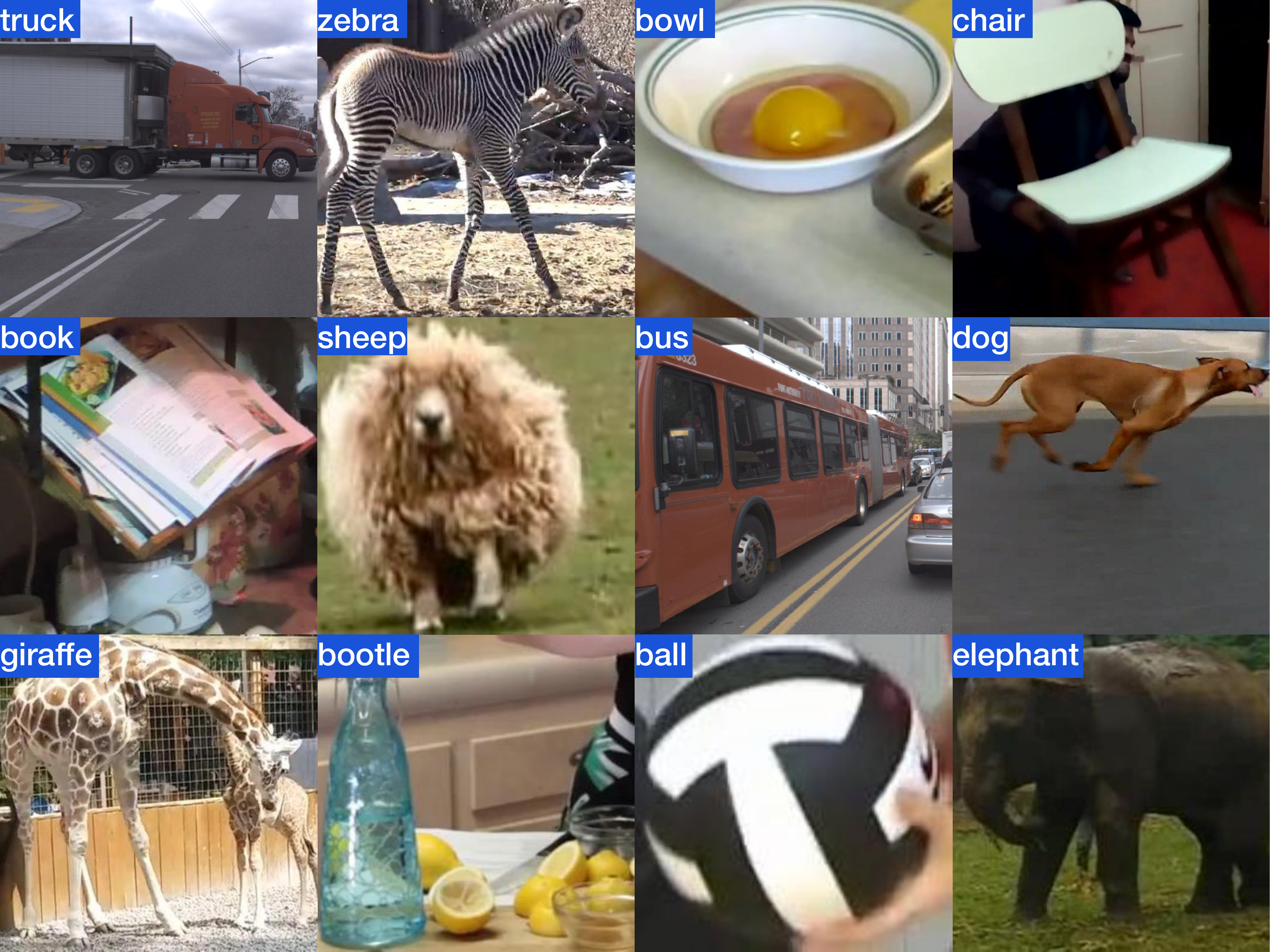}
    \end{subfigure}
    \caption{Additional examples of \known object categories.}
    \label{fig:qualitative-tao-known}
\end{figure}

We provide the additional examples of \textit{known}, \textit{distractor} and \textit{unknown} object categories in Figures \ref{fig:qualitative-tao-known},  \ref{fig:qualitative-tao-distractor} and \ref{fig:qualitative-tao-unknown}. We also show tracking results of our Open-World Tracking Baseline (OWTB) for \textit{knowns}, \textit{unknowns} and \textit{unknown unknowns} in Figures \ref{fig:tracking_known}, \ref{fig:tracking_unknown} and \ref{fig:additional_non_gt_tracking}.

\renewcommand{\mysize}{1.0}
\begin{figure}[t]
    \centering
    \setlength{\fboxsep}{1pt}%
    \begin{subfigure}[b]{1\linewidth}
        \includegraphics[width=\mysize\linewidth]{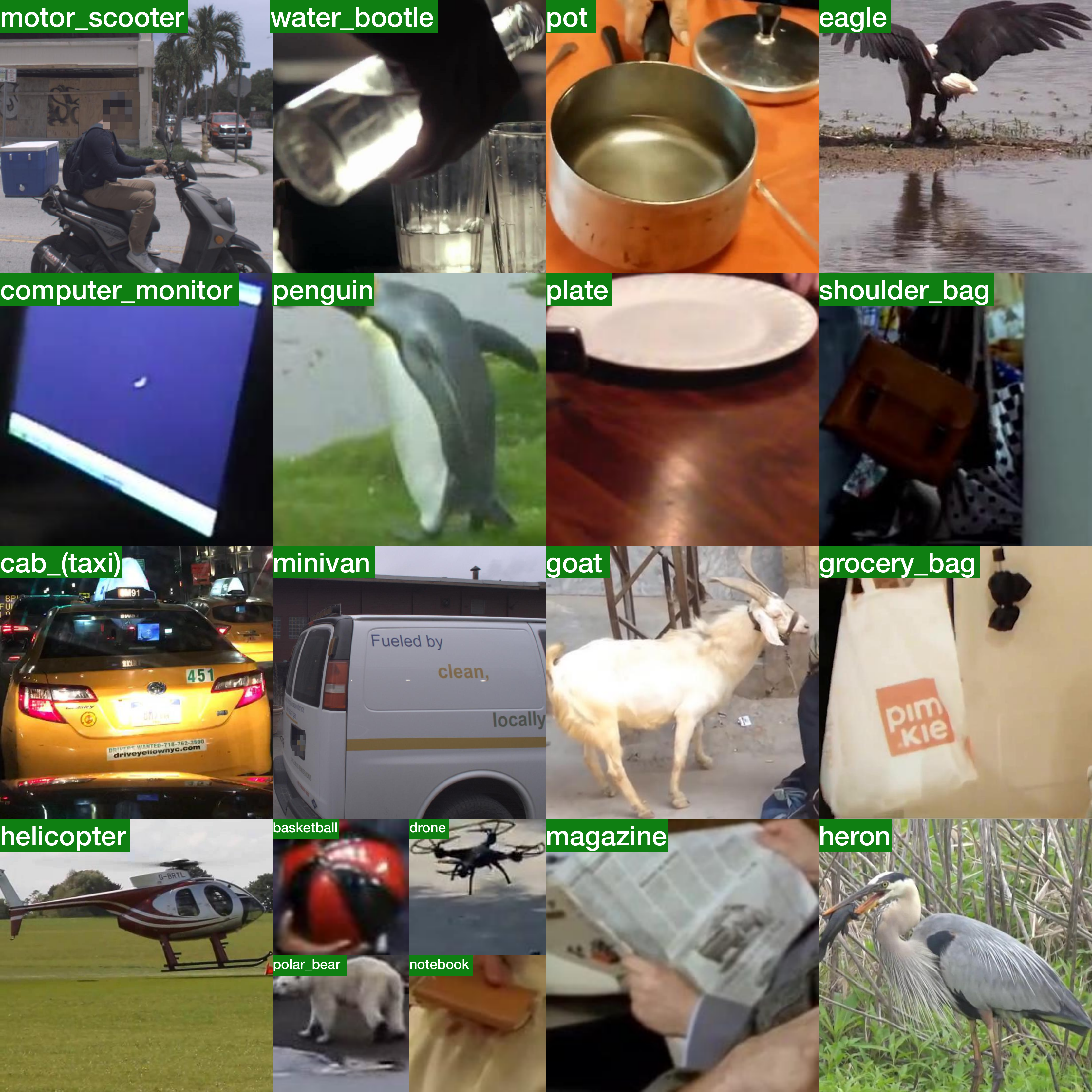}
    \end{subfigure}
    \caption{Examples of \textit{distractor} object categories.}
    \label{fig:qualitative-tao-distractor}
\end{figure}

\makeatletter
\setlength{\@fptop}{0pt}
\makeatother
\renewcommand{\mysize}{1.0}
\begin{figure}[t]
    \centering
    \includegraphics[width=\mysize\linewidth]{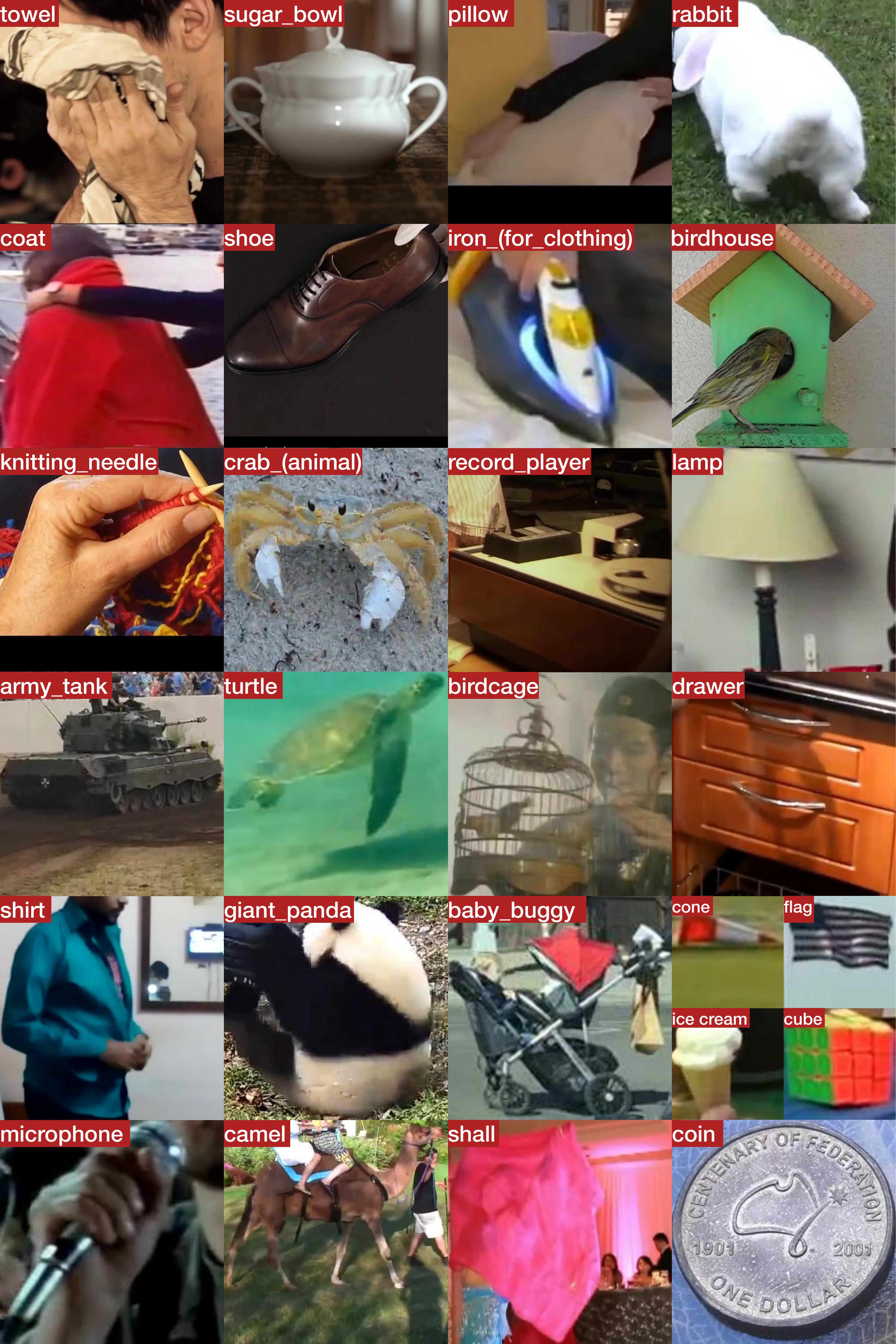}
    \caption{Additional examples of \unknown object categories.}
    \label{fig:qualitative-tao-unknown}
\end{figure}

\begin{figure*}[t]
    \centering
    \setlength{\fboxsep}{0.32pt}%
    \begin{subfigure}[b]{0.32\linewidth}
        \includegraphics[width=\mysize\linewidth]{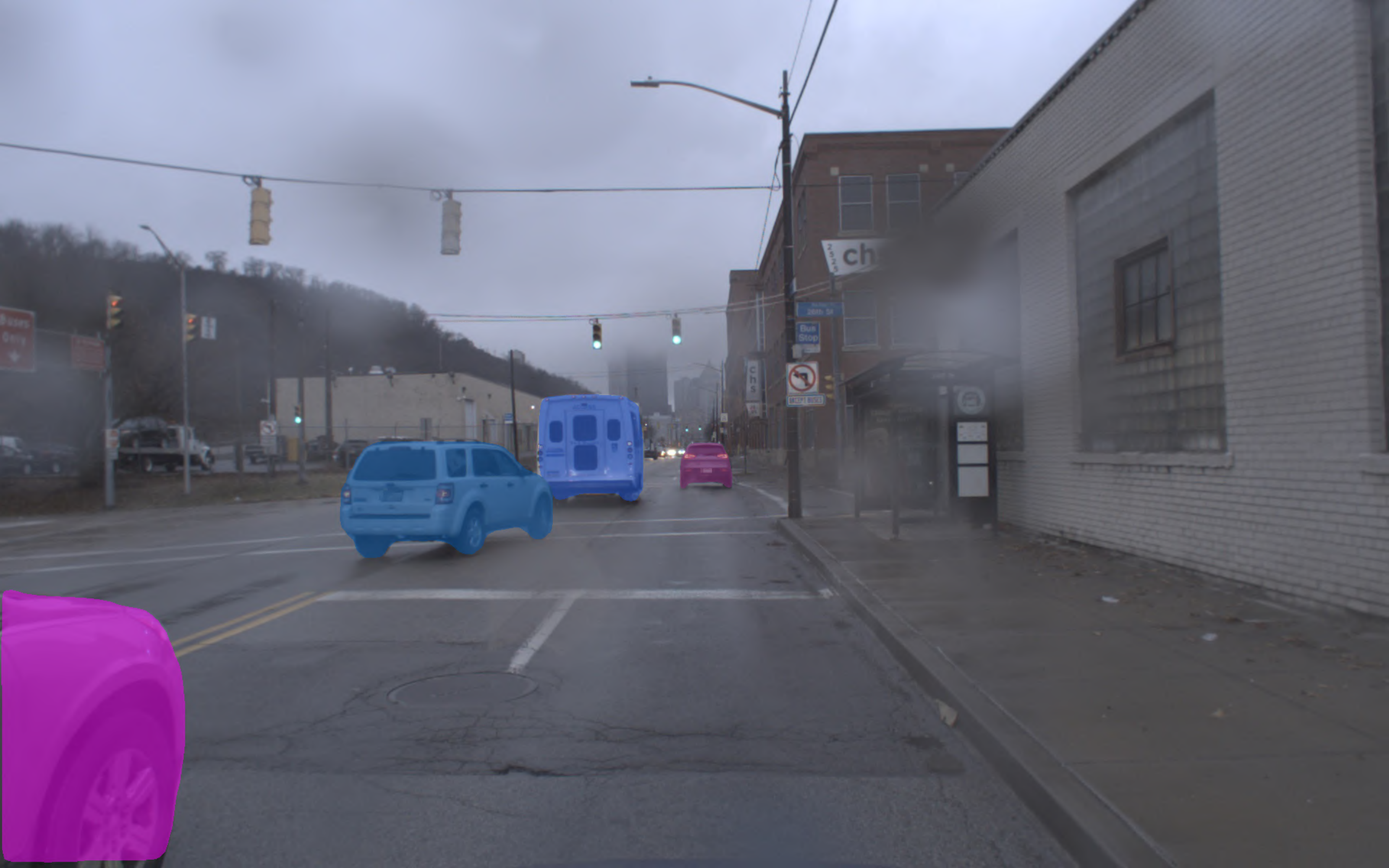}
        \vspace{-16pt}
    \end{subfigure}
    \begin{subfigure}[b]{0.32\linewidth}
        \includegraphics[width=\mysize\linewidth]{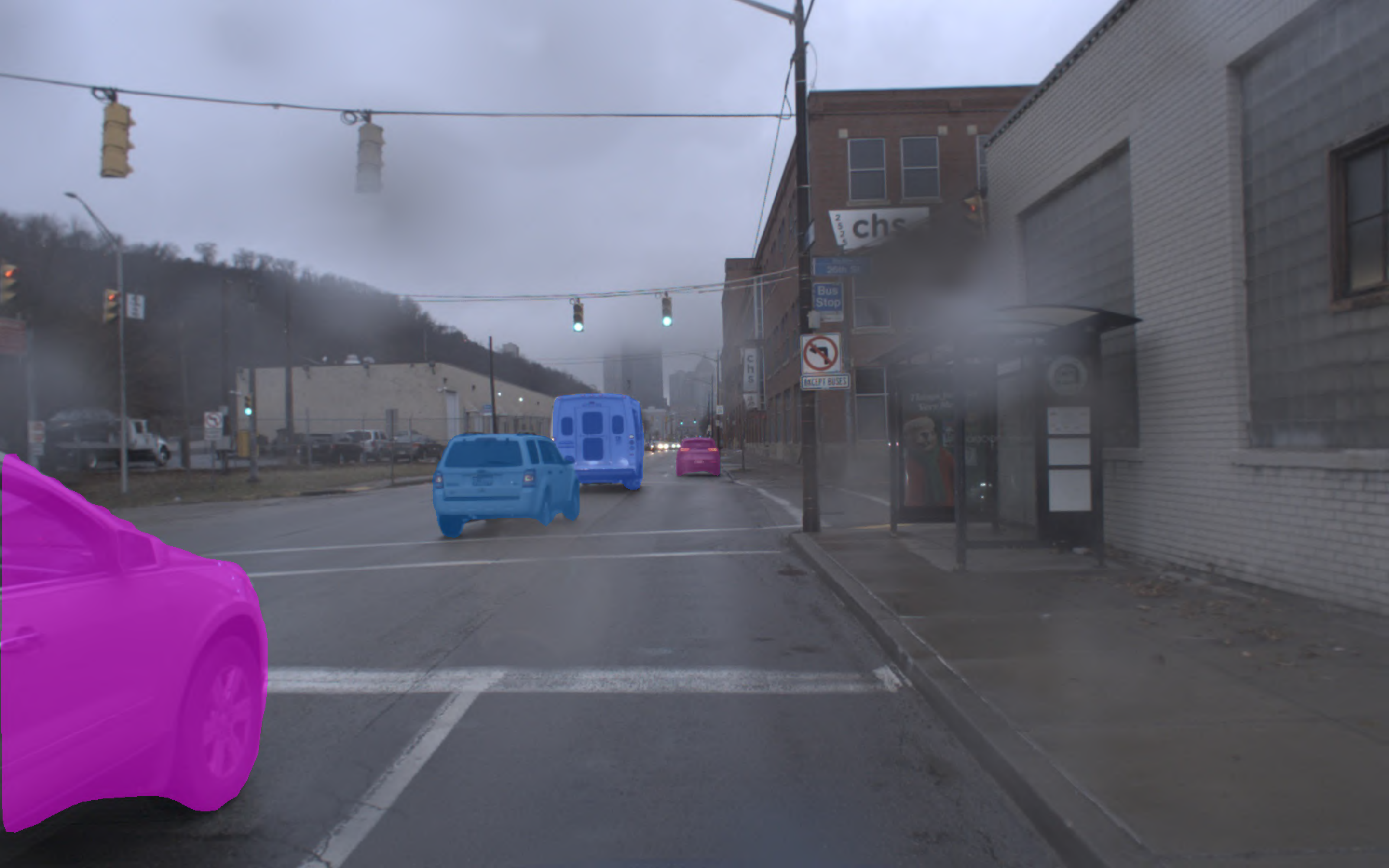}
        \vspace{-16pt}
    \end{subfigure}
    \begin{subfigure}[b]{0.32\linewidth}
     \includegraphics[width=\mysize\linewidth]{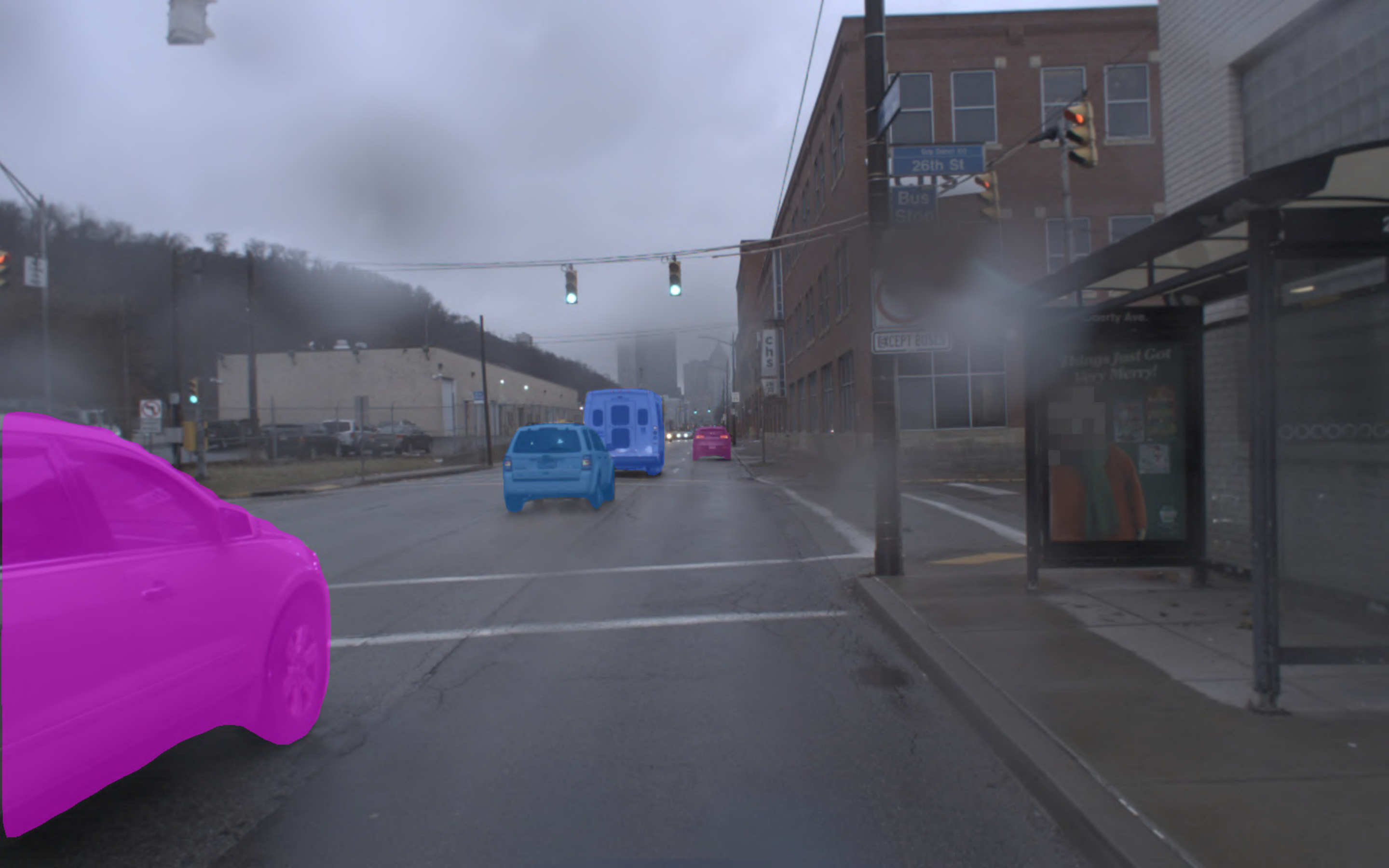}
        \vspace{-16pt}
    \end{subfigure}
    \vspace{5pt}
	
    \begin{subfigure}[b]{0.32\linewidth}
        \includegraphics[width=\mysize\linewidth]{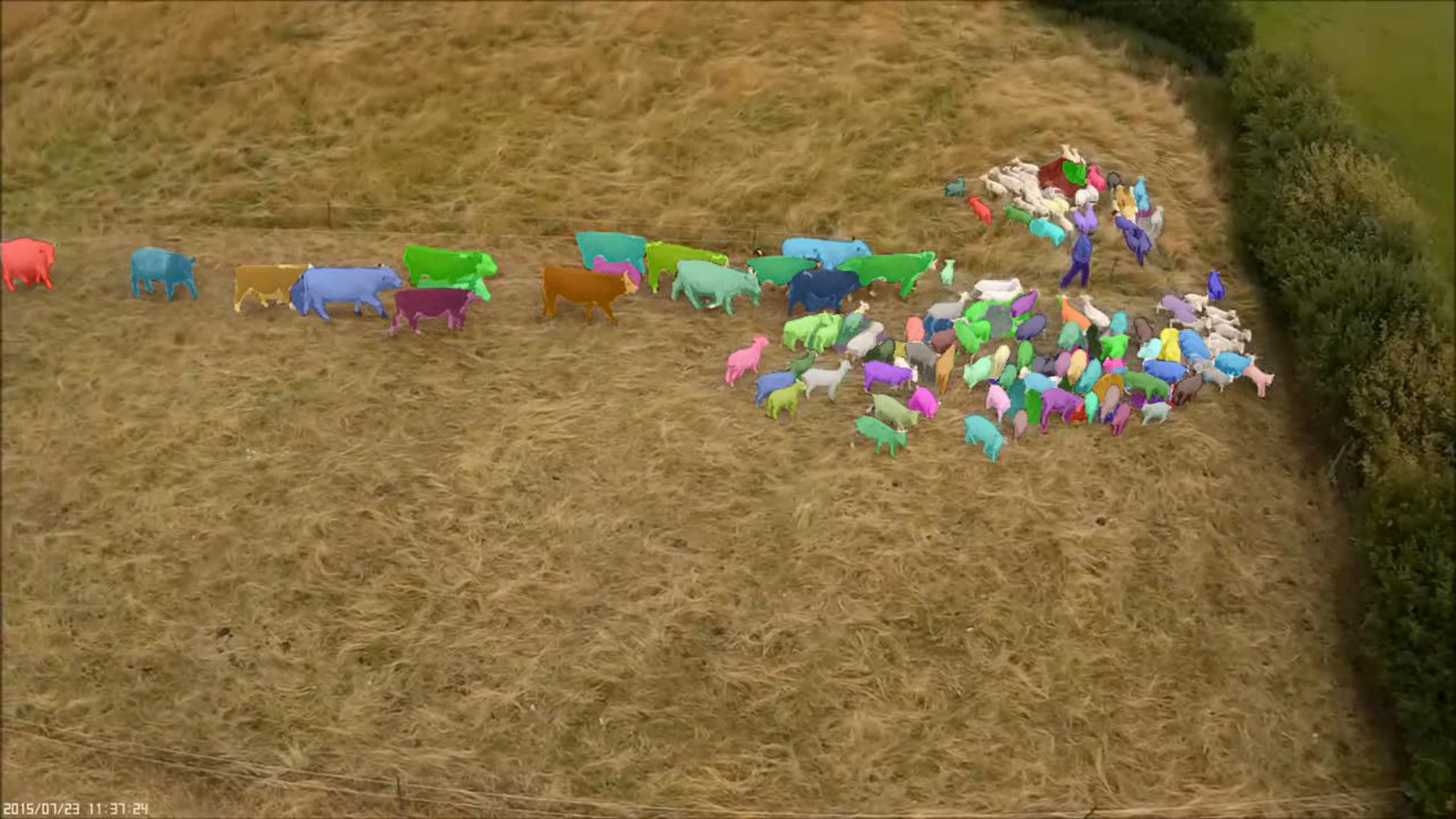}
        \vspace{-16pt}
    \end{subfigure}
    \begin{subfigure}[b]{0.32\linewidth}
        \includegraphics[width=\mysize\linewidth]{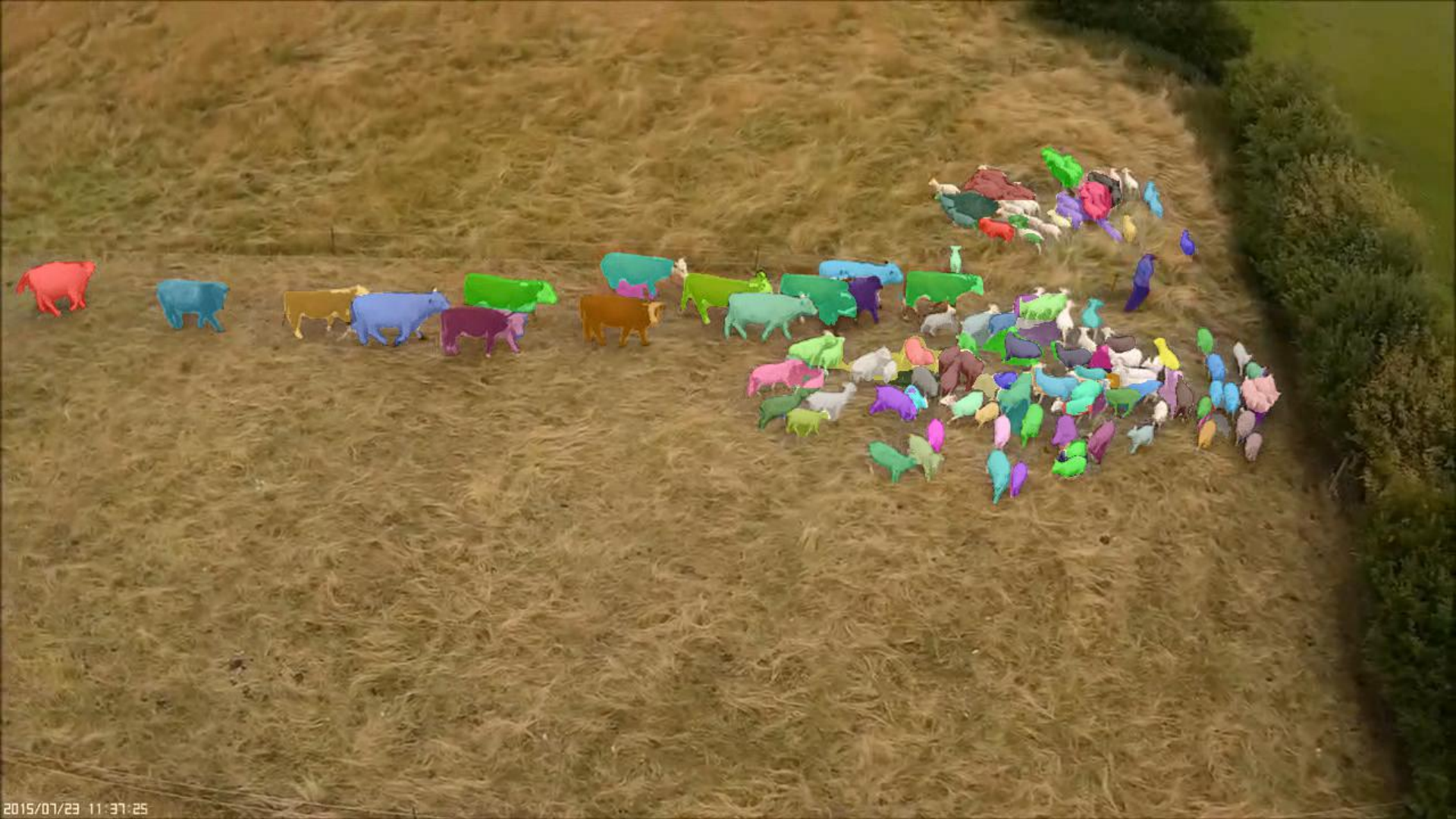}
        \vspace{-16pt}
    \end{subfigure}
        \begin{subfigure}[b]{0.32\linewidth}
        \includegraphics[width=\mysize\linewidth]{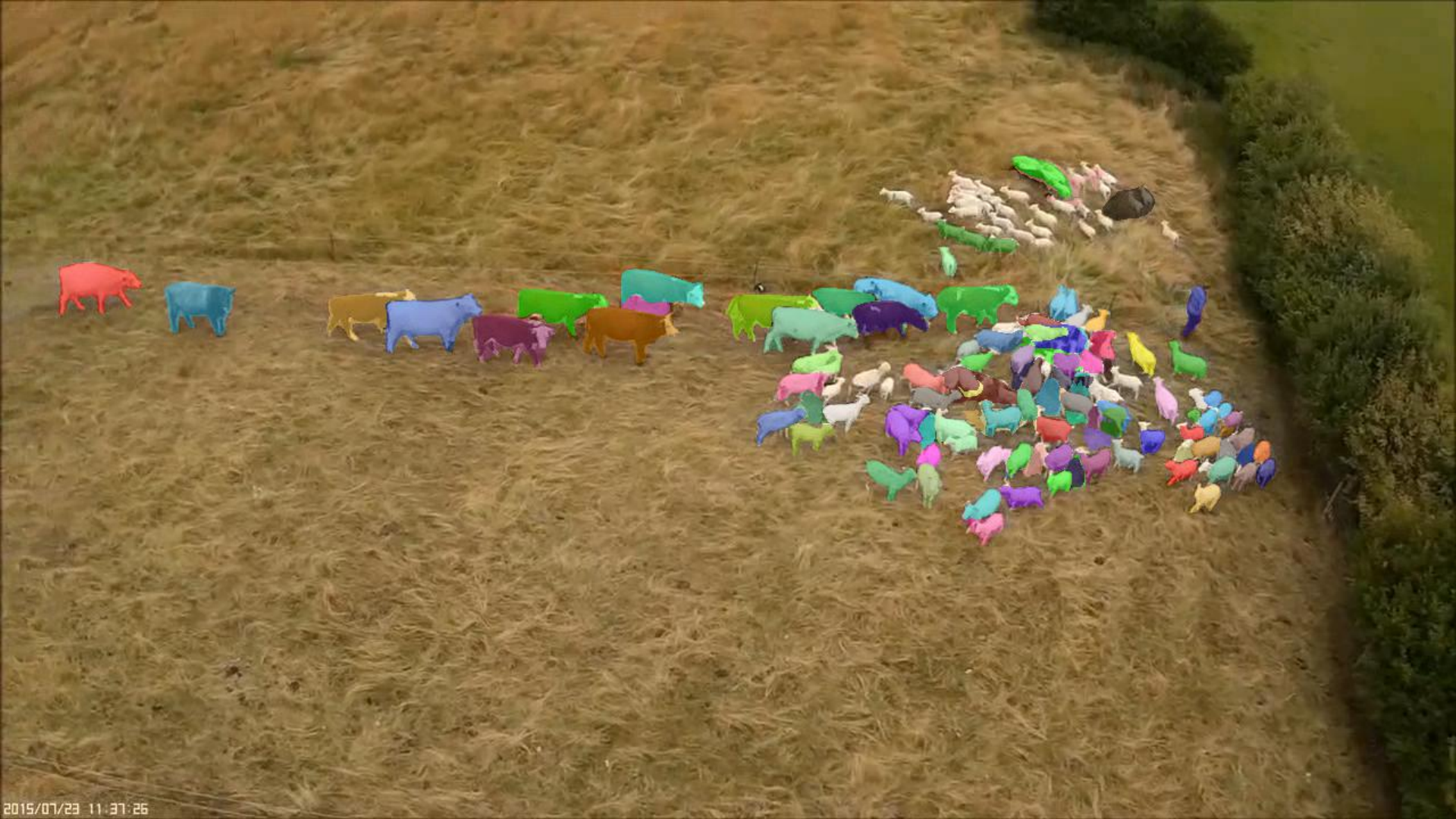}
        \vspace{-16pt}
    \end{subfigure}
    \vspace{5pt}
	
    \begin{subfigure}[b]{0.32\linewidth}
        \includegraphics[width=\mysize\linewidth]{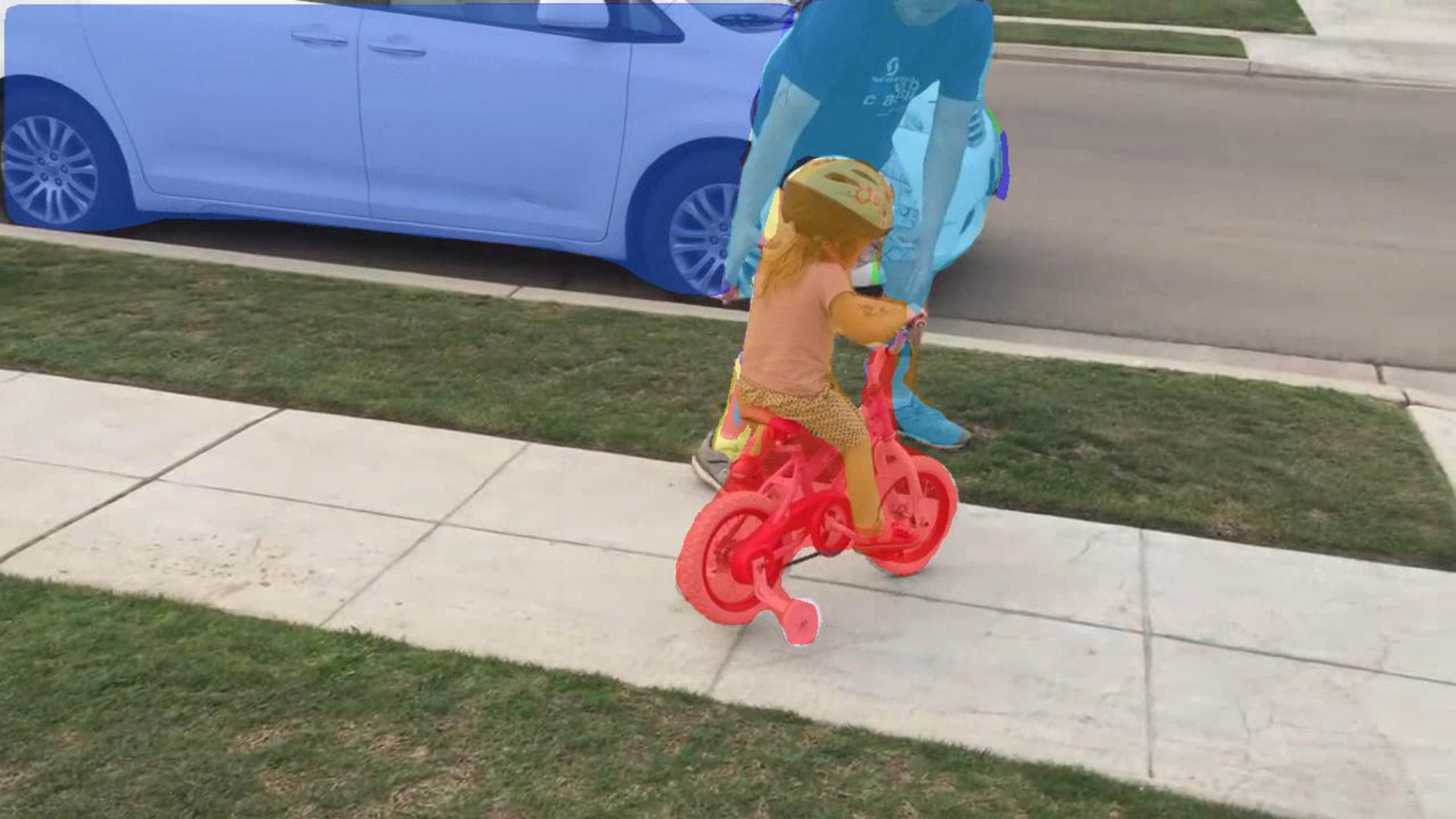}
        \vspace{-16pt}
    \end{subfigure}
    \begin{subfigure}[b]{0.32\linewidth}
        \includegraphics[width=\mysize\linewidth]{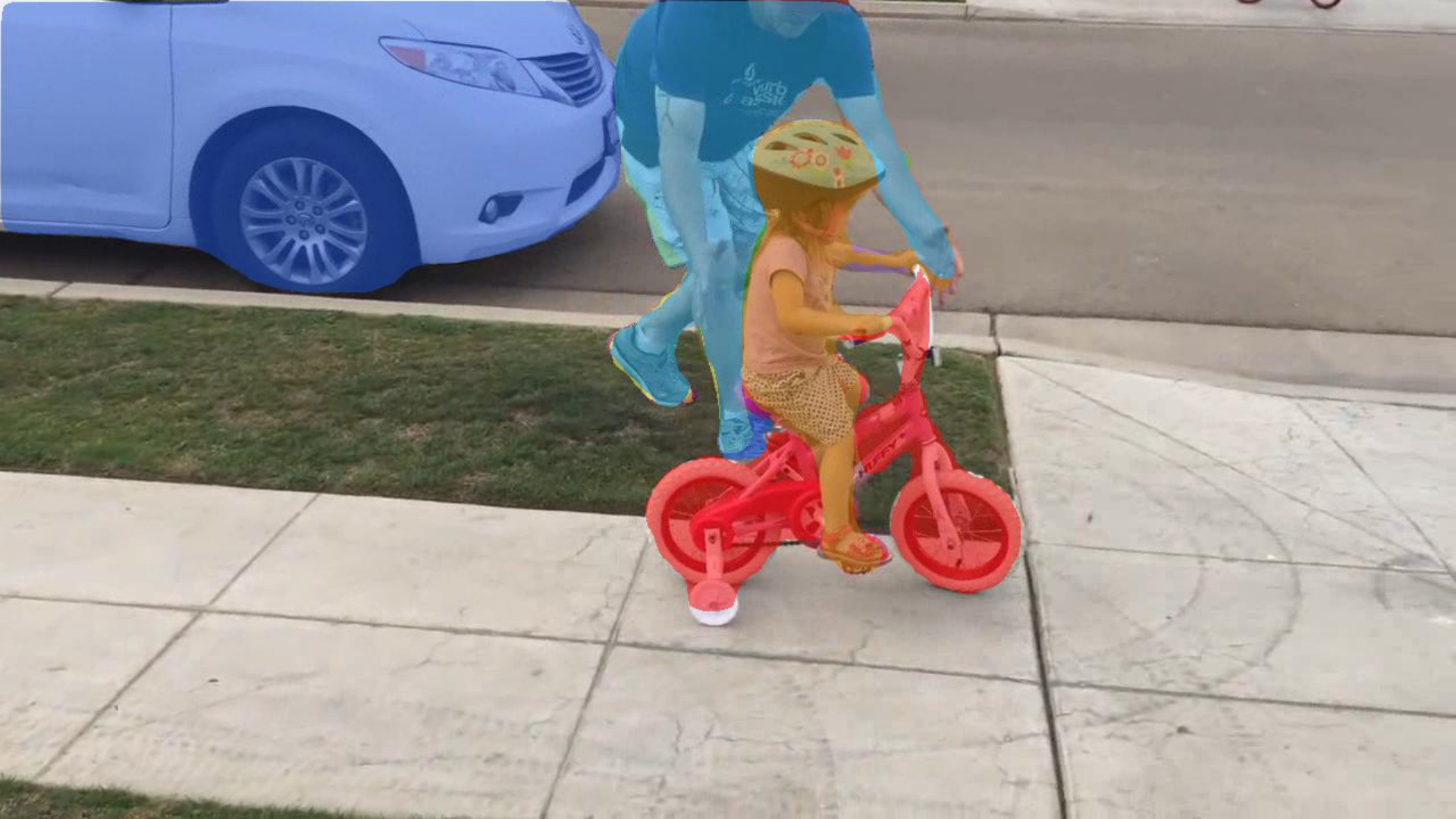}
        \vspace{-16pt}
    \end{subfigure}
        \begin{subfigure}[b]{0.32\linewidth}
        \includegraphics[width=\mysize\linewidth]{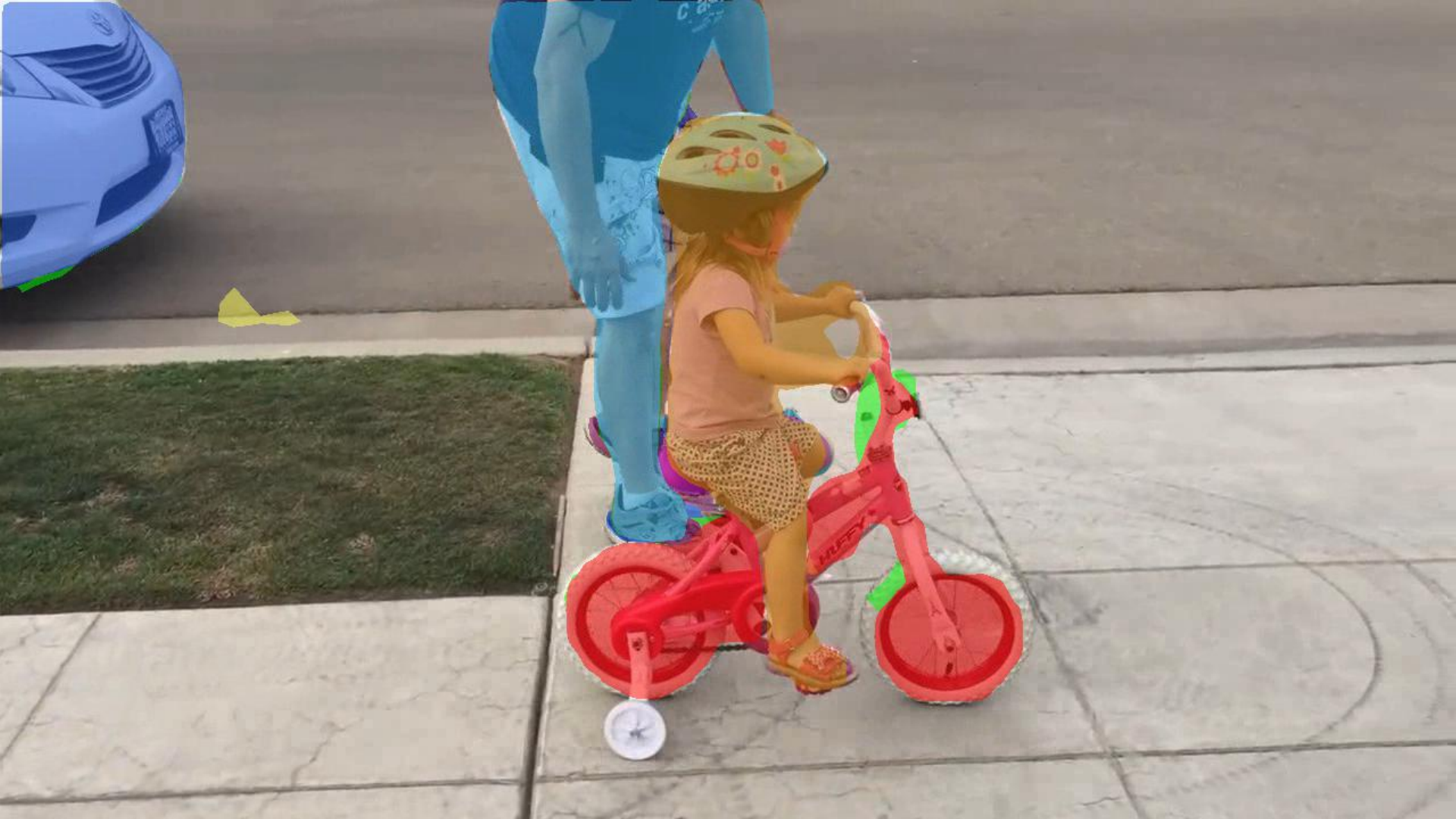}
        \vspace{-16pt}
    \end{subfigure}
    \vspace{5pt}
    
    \begin{subfigure}[b]{0.32\linewidth}
        \includegraphics[width=\mysize\linewidth]{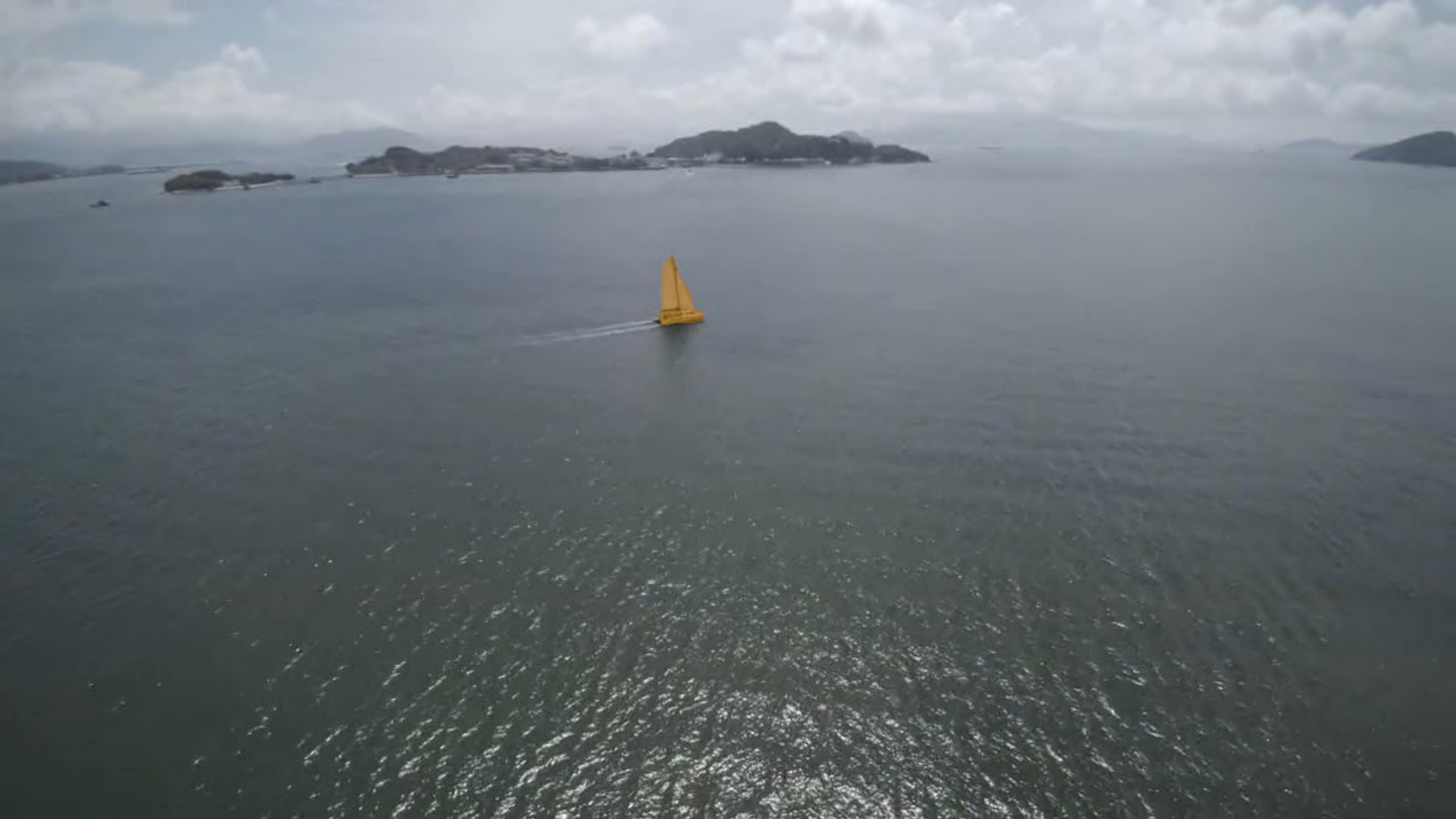}
        \vspace{-16pt}
    \end{subfigure}
    \begin{subfigure}[b]{0.32\linewidth}
        \includegraphics[width=\mysize\linewidth]{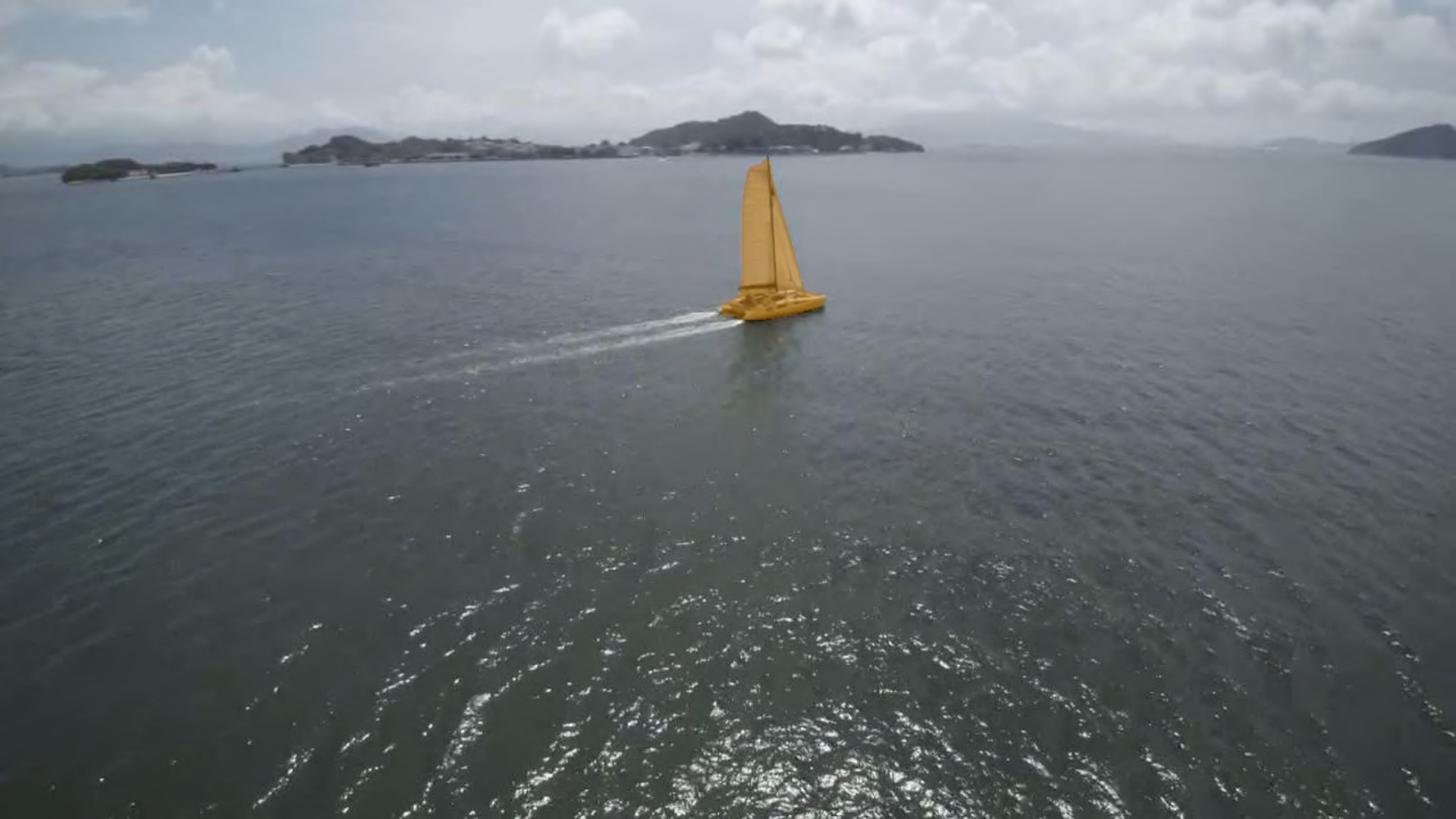}
        \vspace{-16pt}
    \end{subfigure}
    \begin{subfigure}[b]{0.32\linewidth}
     \includegraphics[width=\mysize\linewidth]{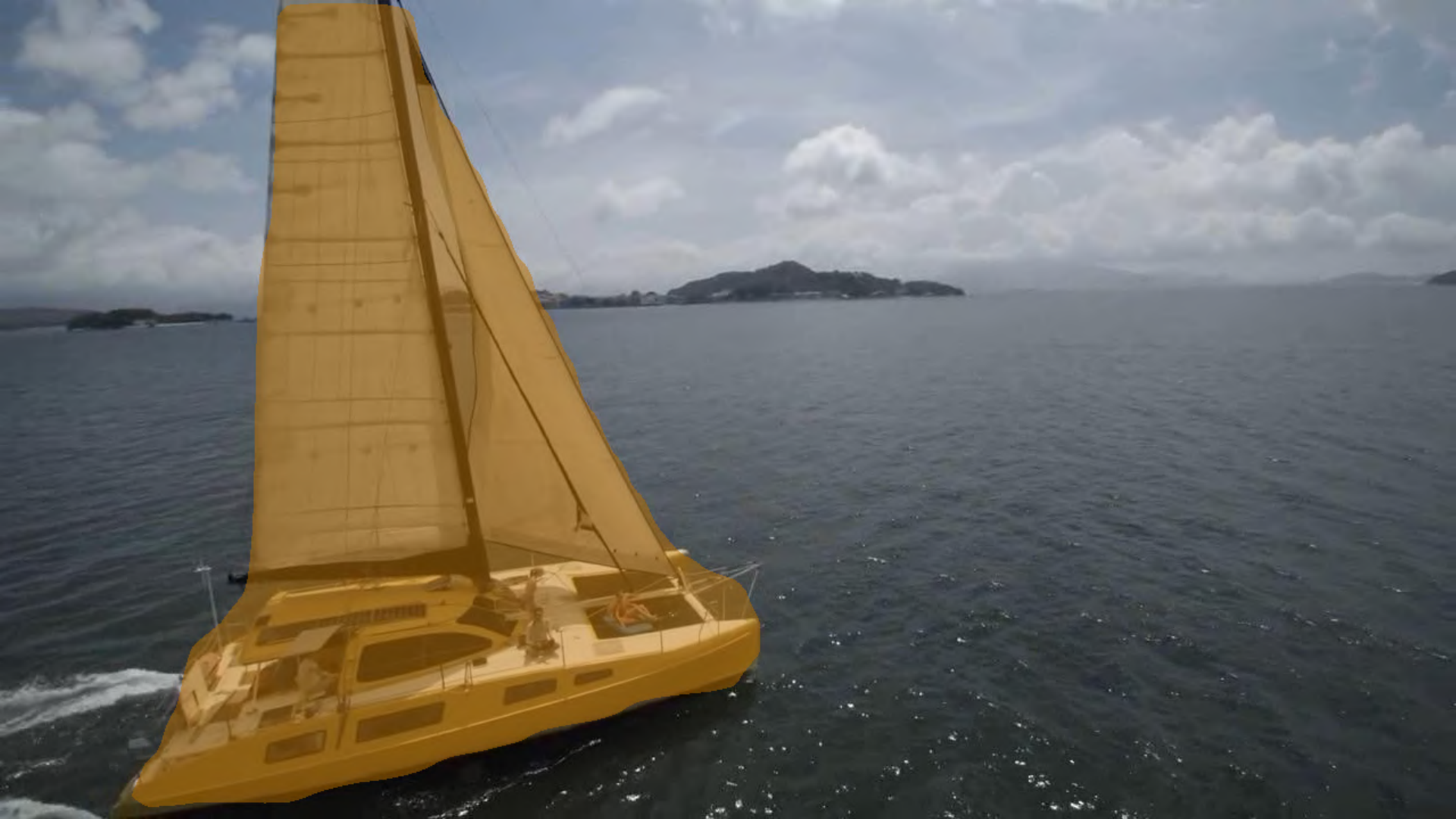}
        \vspace{-16pt}
    \end{subfigure}
    \vspace{5pt}
    
    \begin{subfigure}[b]{0.32\linewidth}
        \includegraphics[width=\mysize\linewidth]{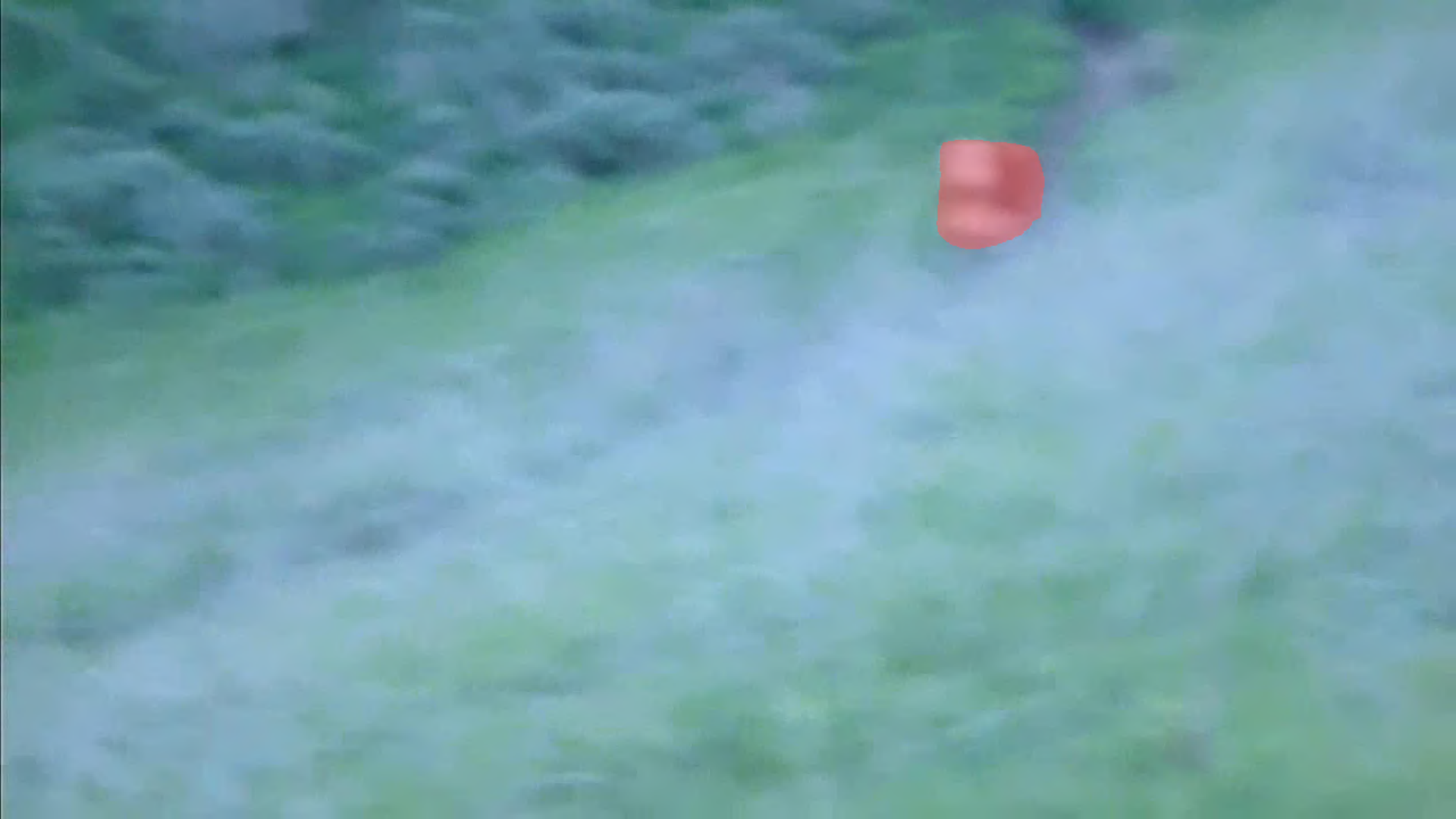}
        \vspace{-16pt}
    \end{subfigure}
    \begin{subfigure}[b]{0.32\linewidth}
        \includegraphics[width=\mysize\linewidth]{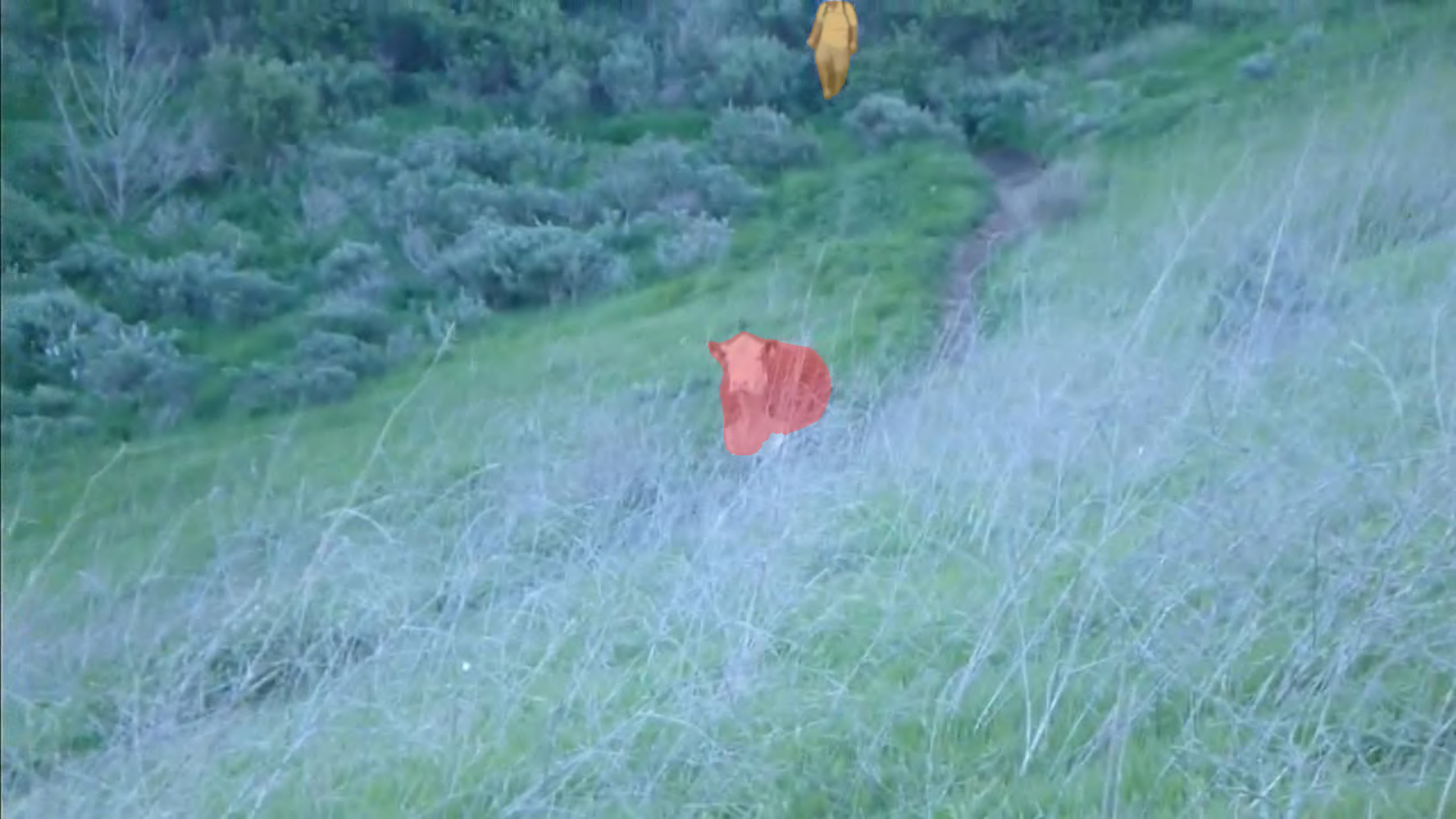}
        \vspace{-16pt}
    \end{subfigure}
        \begin{subfigure}[b]{0.32\linewidth}
        \includegraphics[width=\mysize\linewidth]{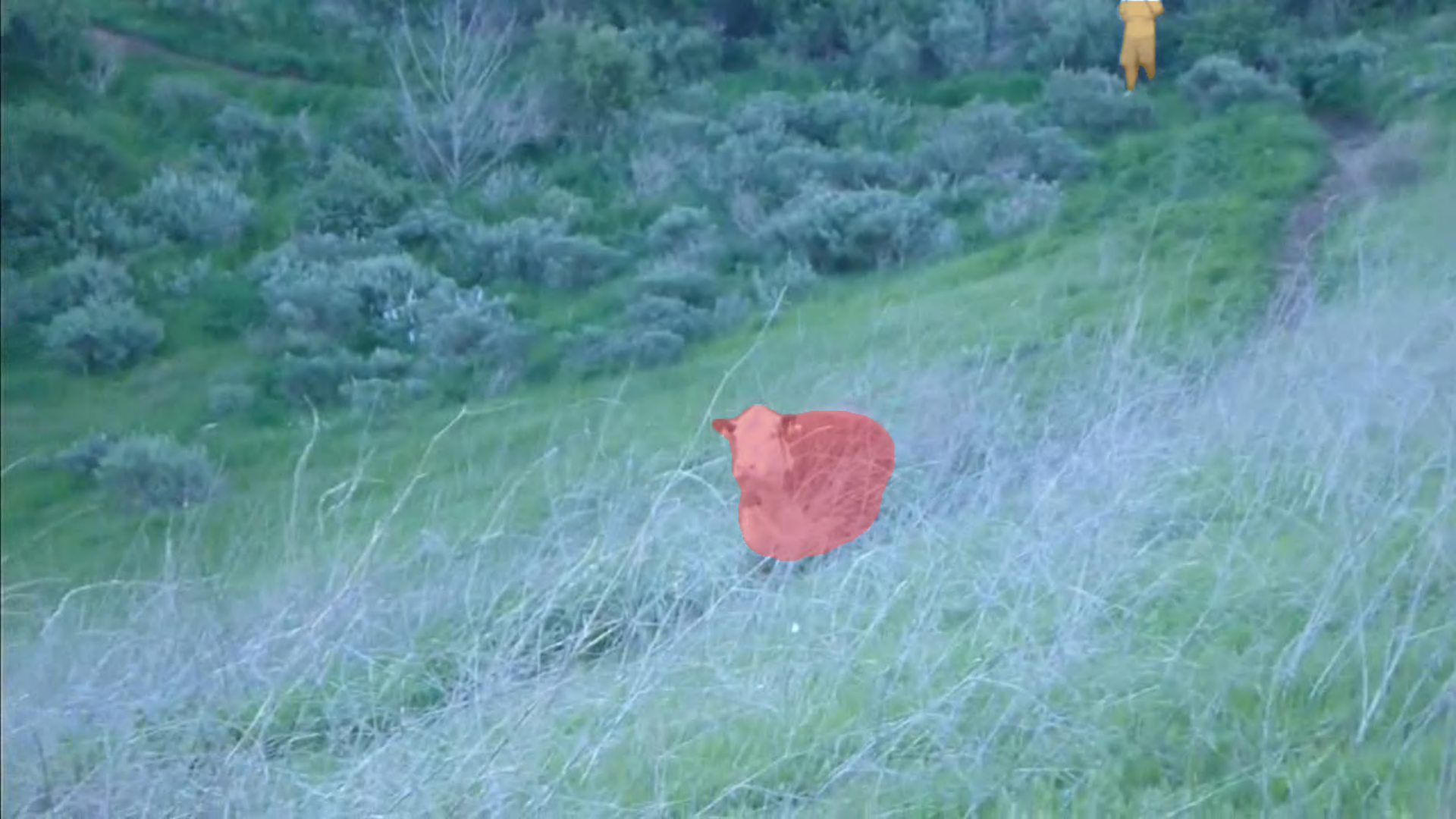}
        \vspace{-16pt}
    \end{subfigure}
    \vspace{5pt}
    
    \begin{subfigure}[b]{0.32\linewidth}
        \includegraphics[width=\mysize\linewidth]{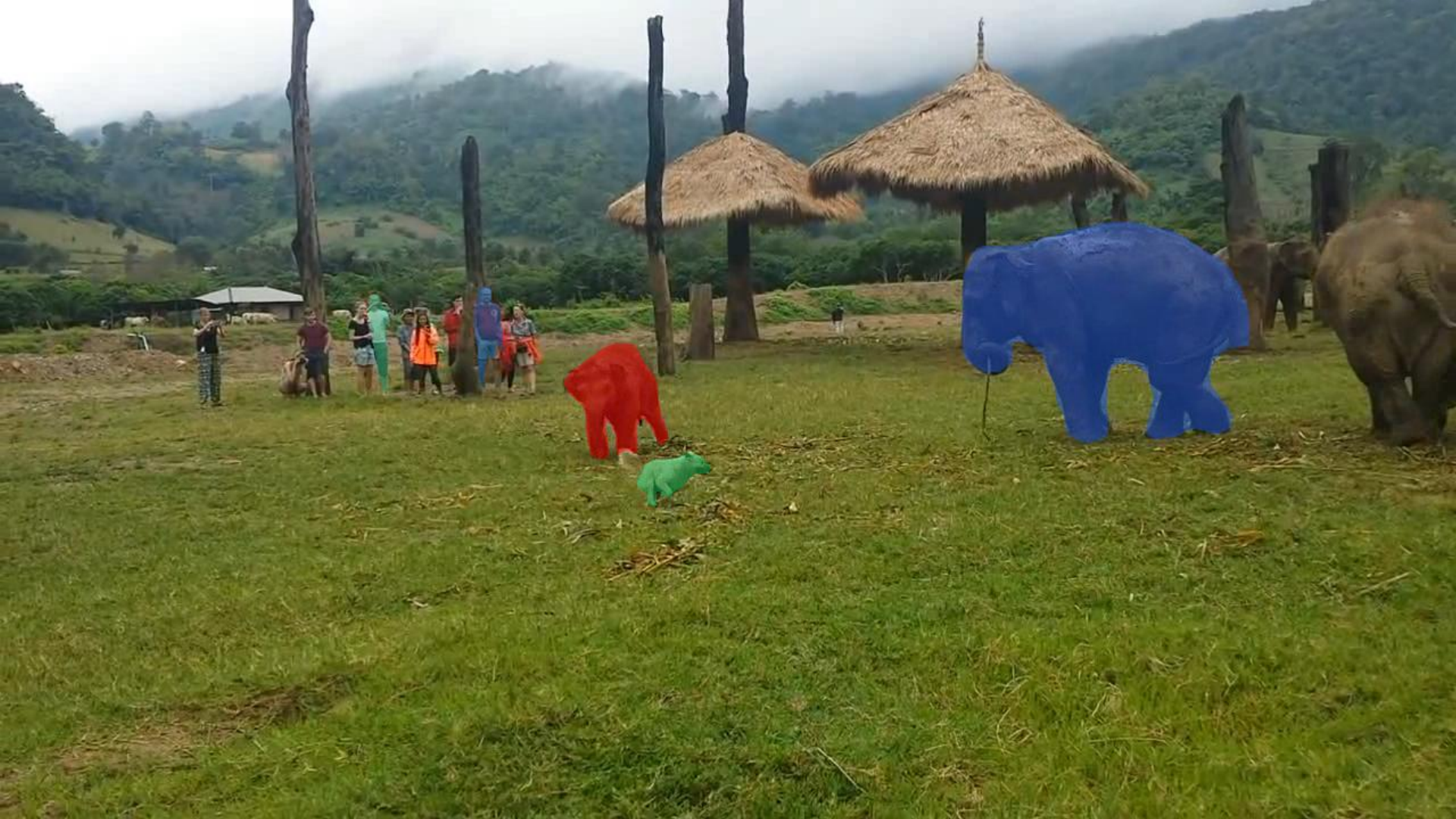}
        \vspace{-16pt}
    \end{subfigure}
    \begin{subfigure}[b]{0.32\linewidth}
        \includegraphics[width=\mysize\linewidth]{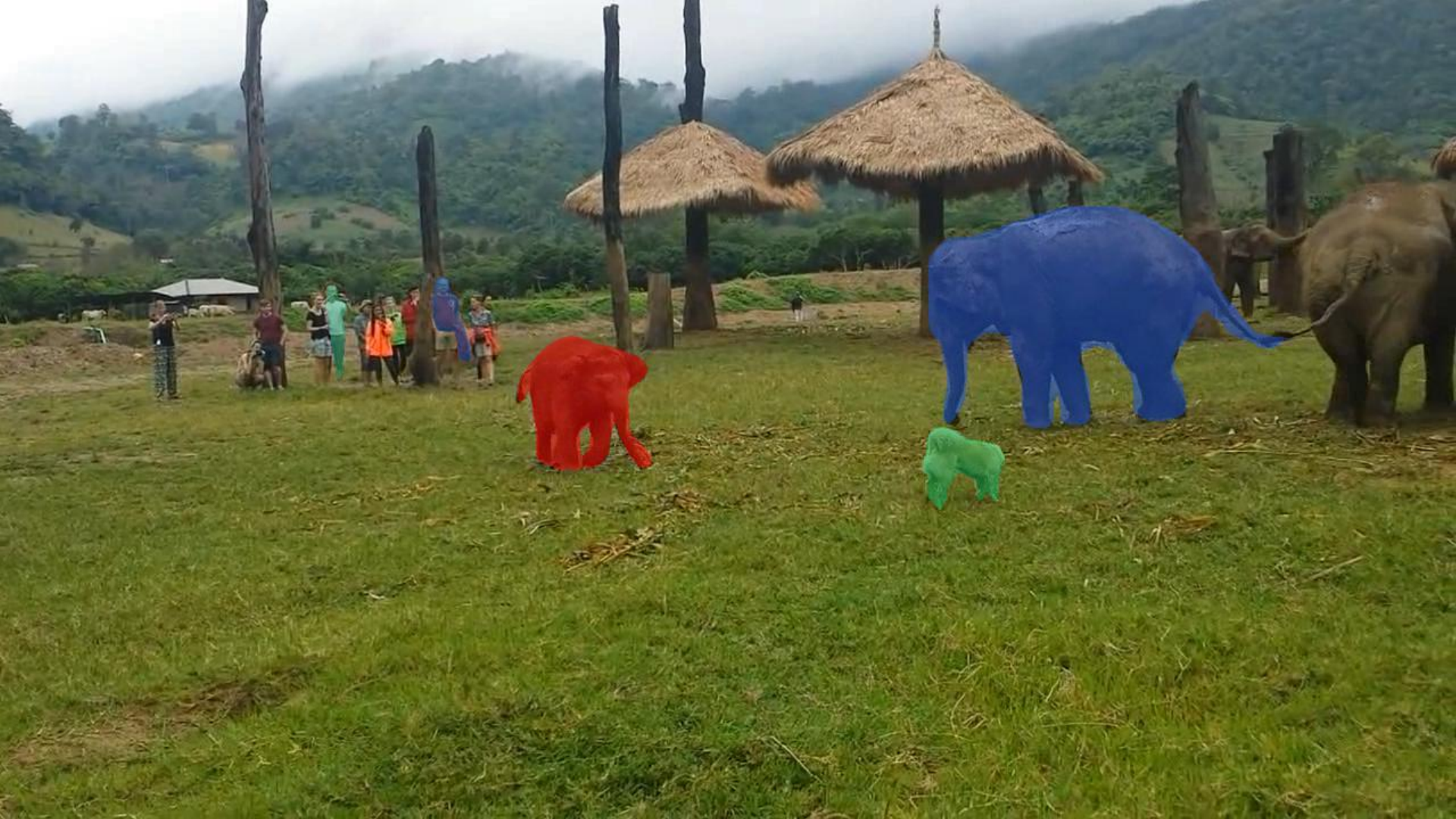}
        \vspace{-16pt}
    \end{subfigure}
        \begin{subfigure}[b]{0.32\linewidth}
        \includegraphics[width=\mysize\linewidth]{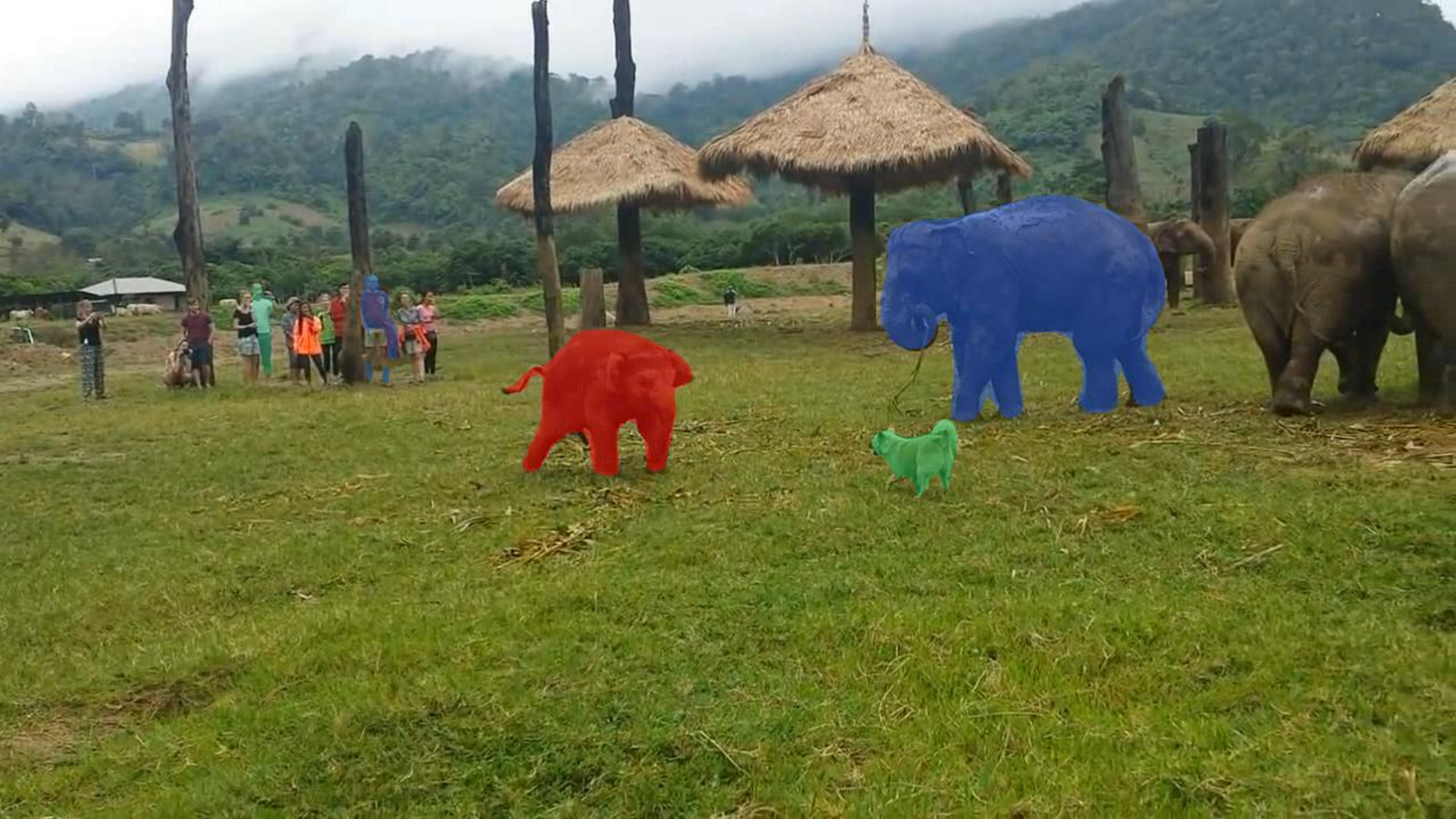}
        \vspace{-16pt}
    \end{subfigure}
    \vspace{5pt}

    \caption{\textbf{Tracking results for \known.} Examples of \known objects tracked by OWTB. OWTB is capable of tracking objects in cluttered scenes (\textit{second row}), and making robust associations despite of motion blur (\textit{fifth row}).}
    \label{fig:tracking_known}
\end{figure*}

\begin{figure*}[t]
    \centering
    \setlength{\fboxsep}{0.32pt}%
    \begin{subfigure}[b]{0.32\linewidth}
        \includegraphics[width=\mysize\linewidth]{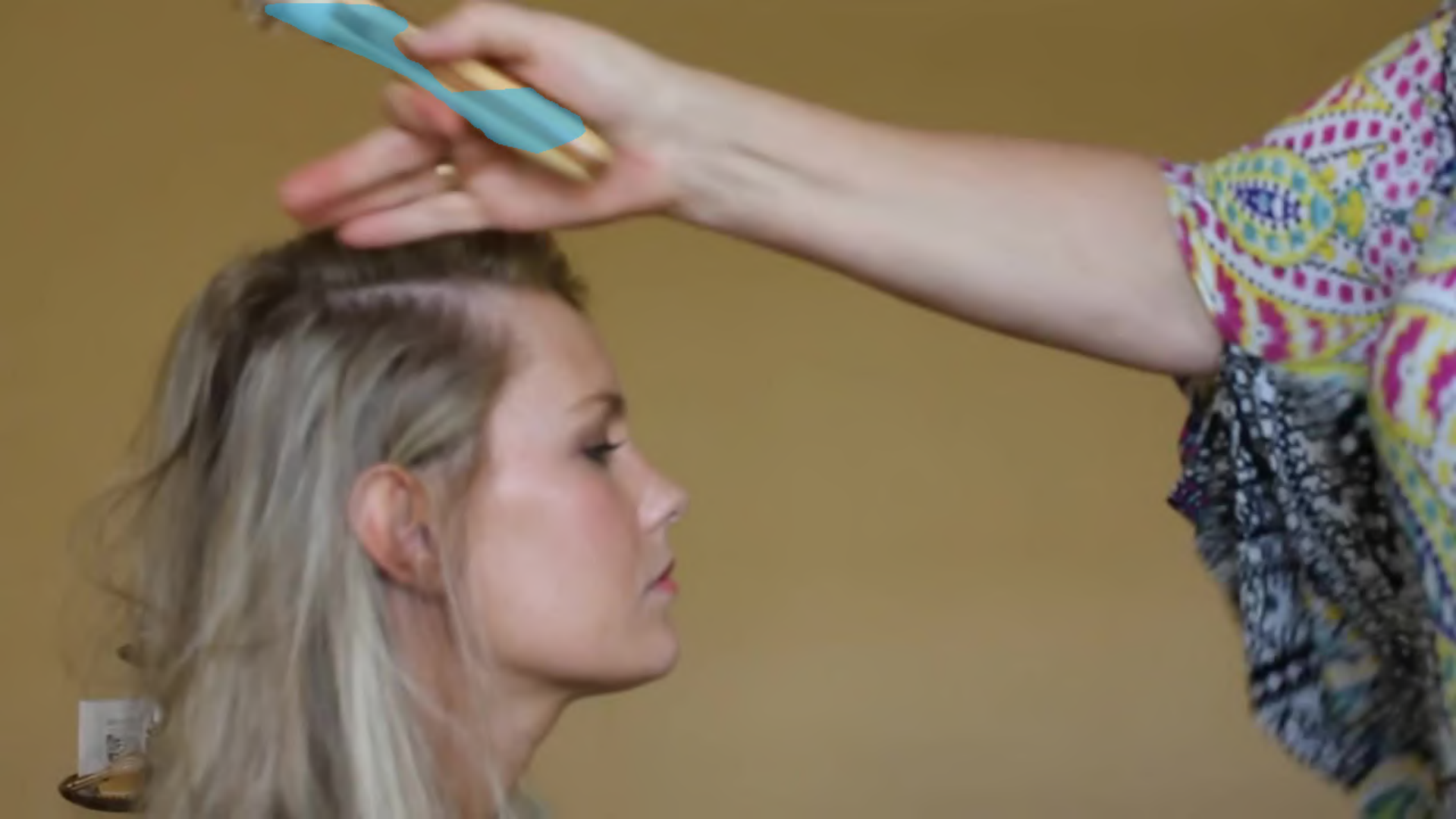}
        \vspace{-16pt}
    \end{subfigure}
    \begin{subfigure}[b]{0.32\linewidth}
        \includegraphics[width=\mysize\linewidth]{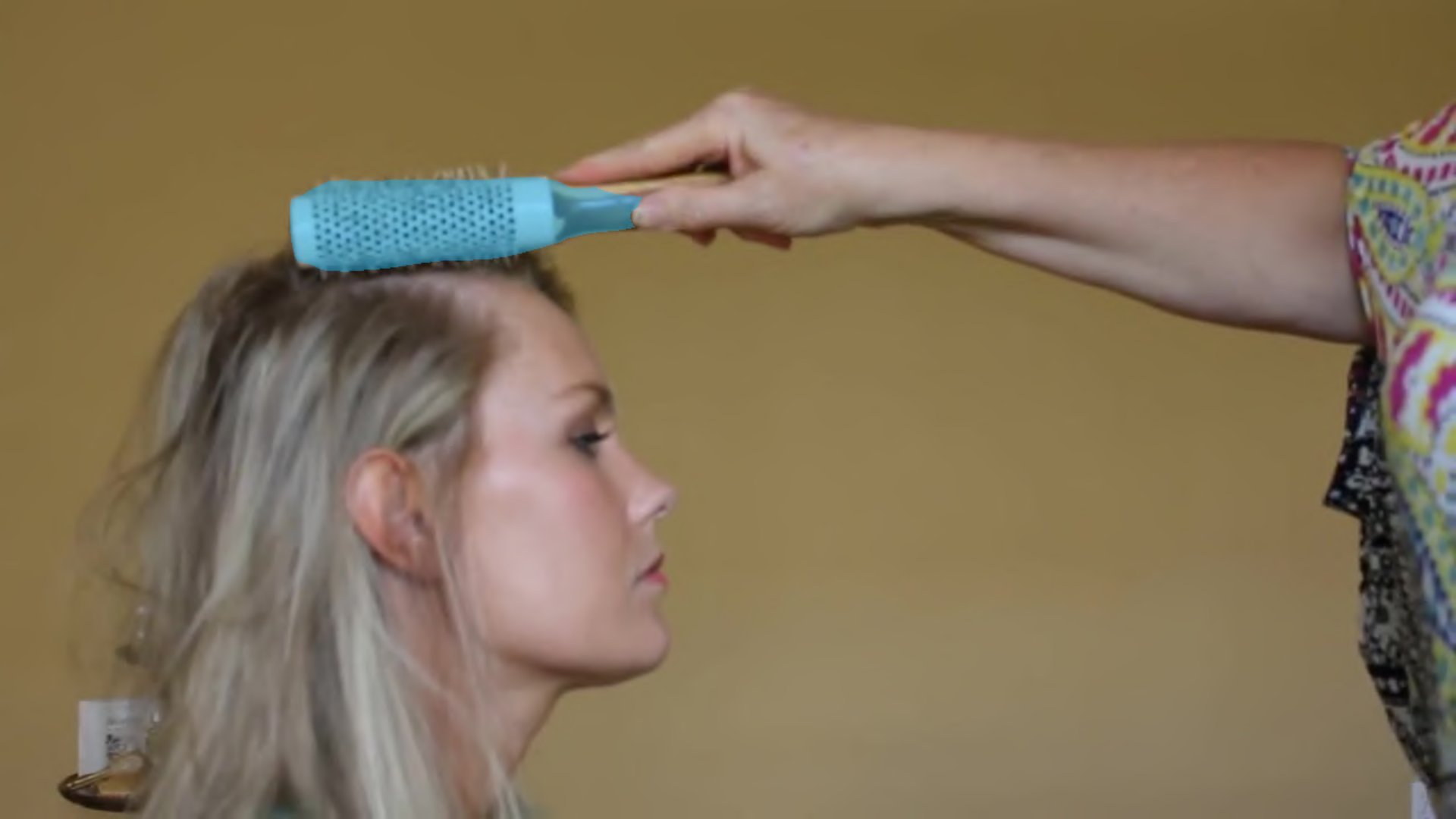}
        \vspace{-16pt}
    \end{subfigure}
    \begin{subfigure}[b]{0.32\linewidth}
     \includegraphics[width=\mysize\linewidth]{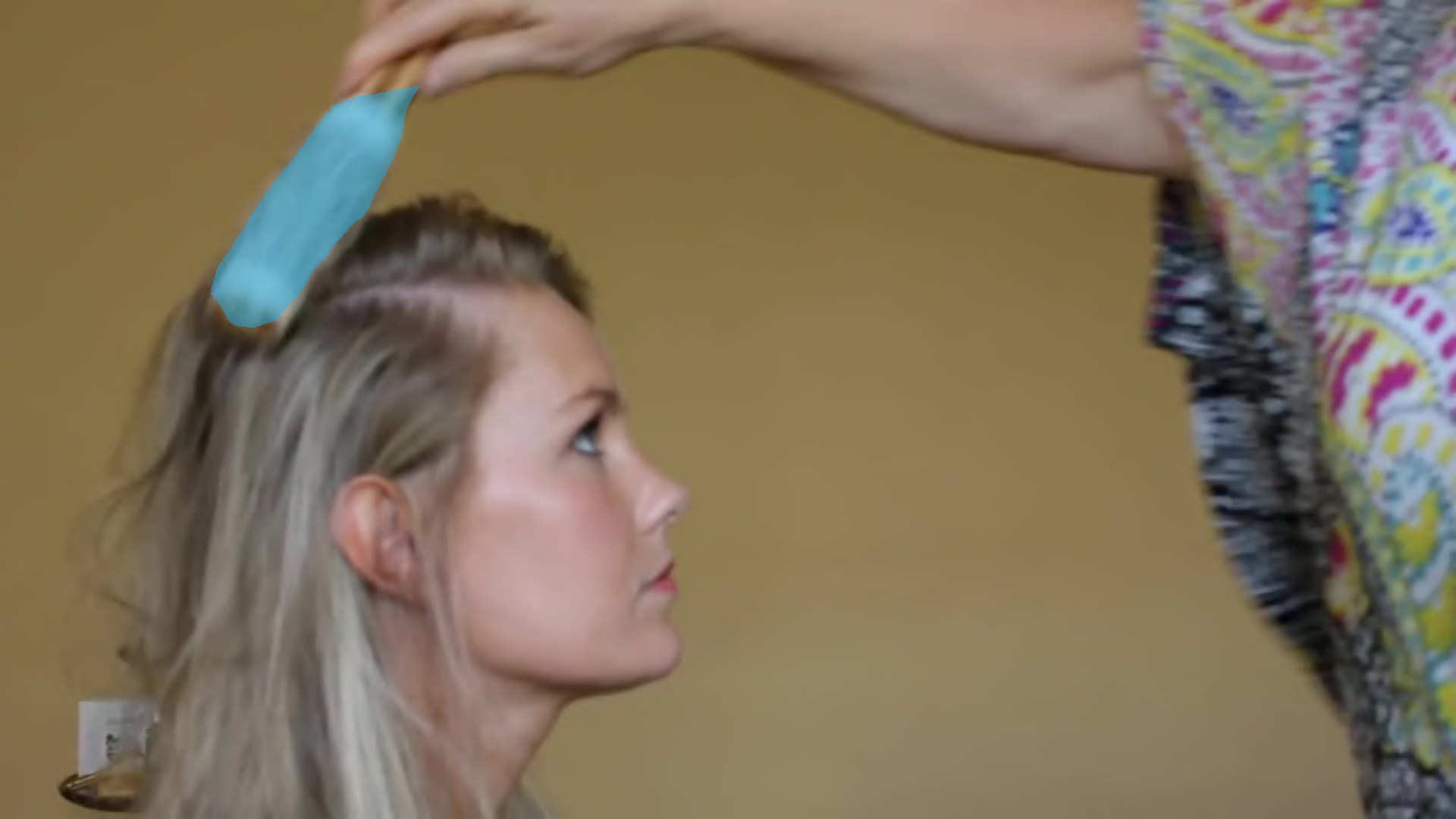}
        \vspace{-16pt}
    \end{subfigure} 
    \vspace{5pt}
	
    \begin{subfigure}[b]{0.32\linewidth}
        \includegraphics[width=\mysize\linewidth]{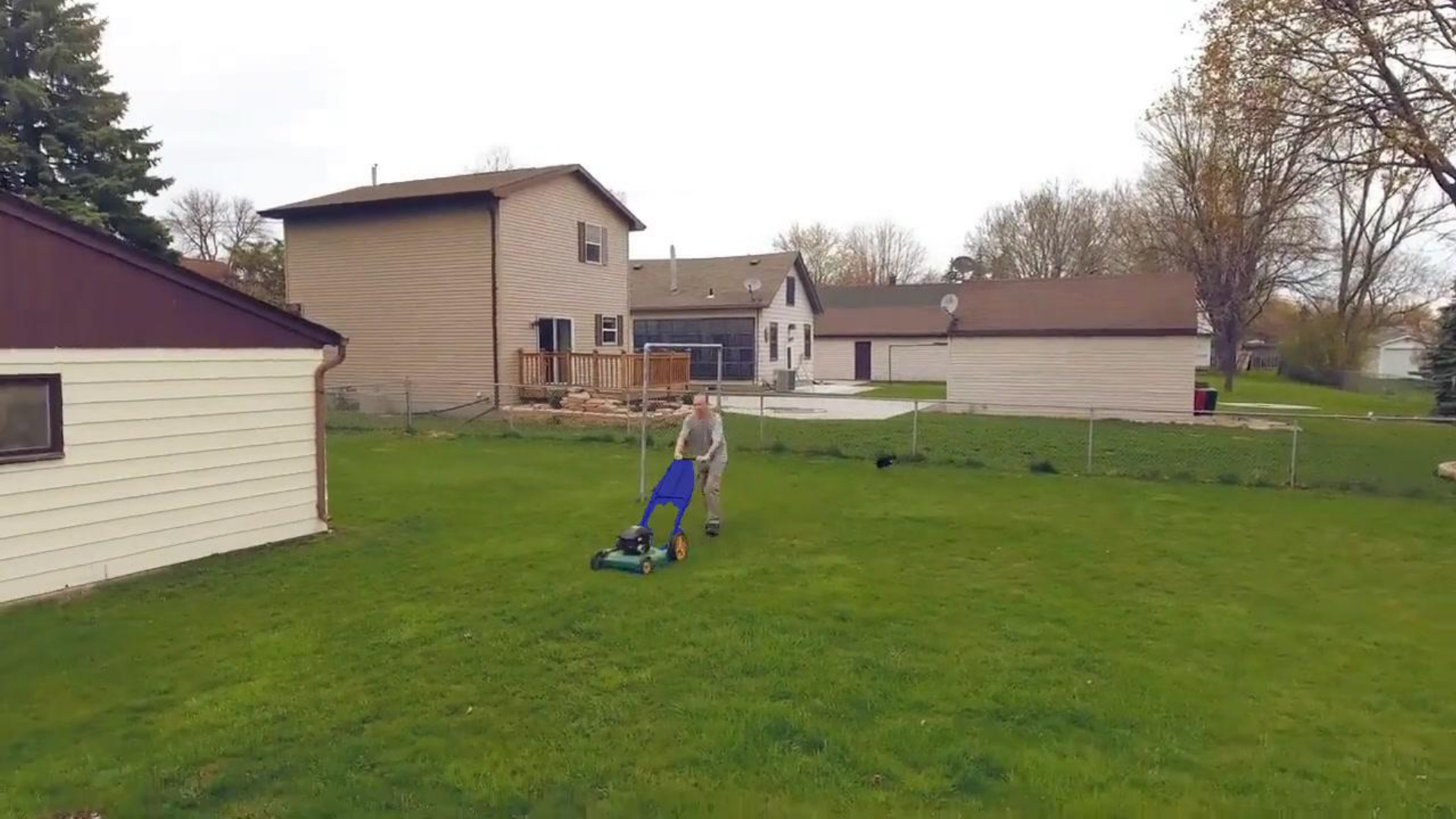}
        \vspace{-16pt}
    \end{subfigure}
    \begin{subfigure}[b]{0.32\linewidth}
        \includegraphics[width=\mysize\linewidth]{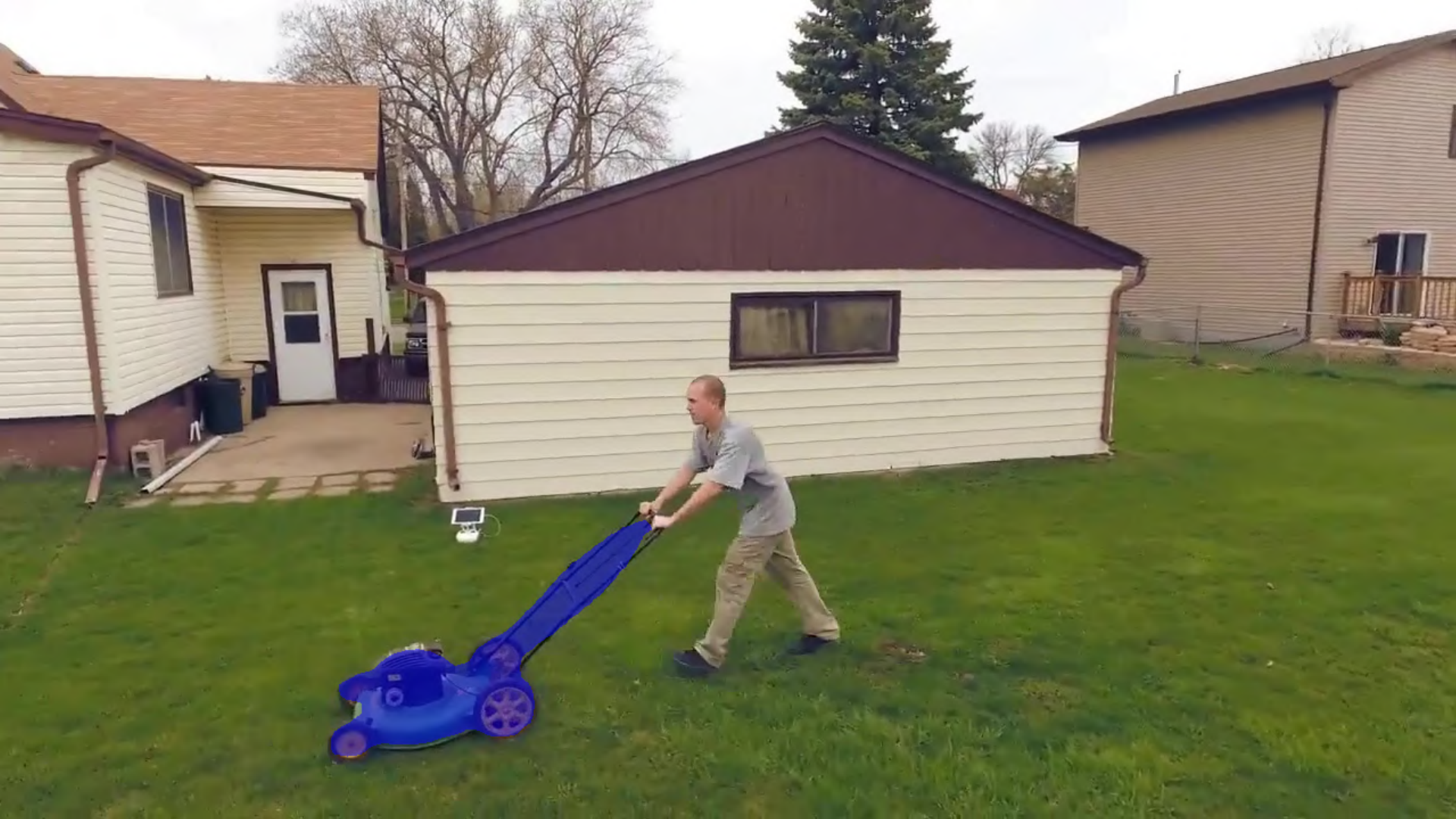}
        \vspace{-16pt}
    \end{subfigure}
        \begin{subfigure}[b]{0.32\linewidth}
        \includegraphics[width=\mysize\linewidth]{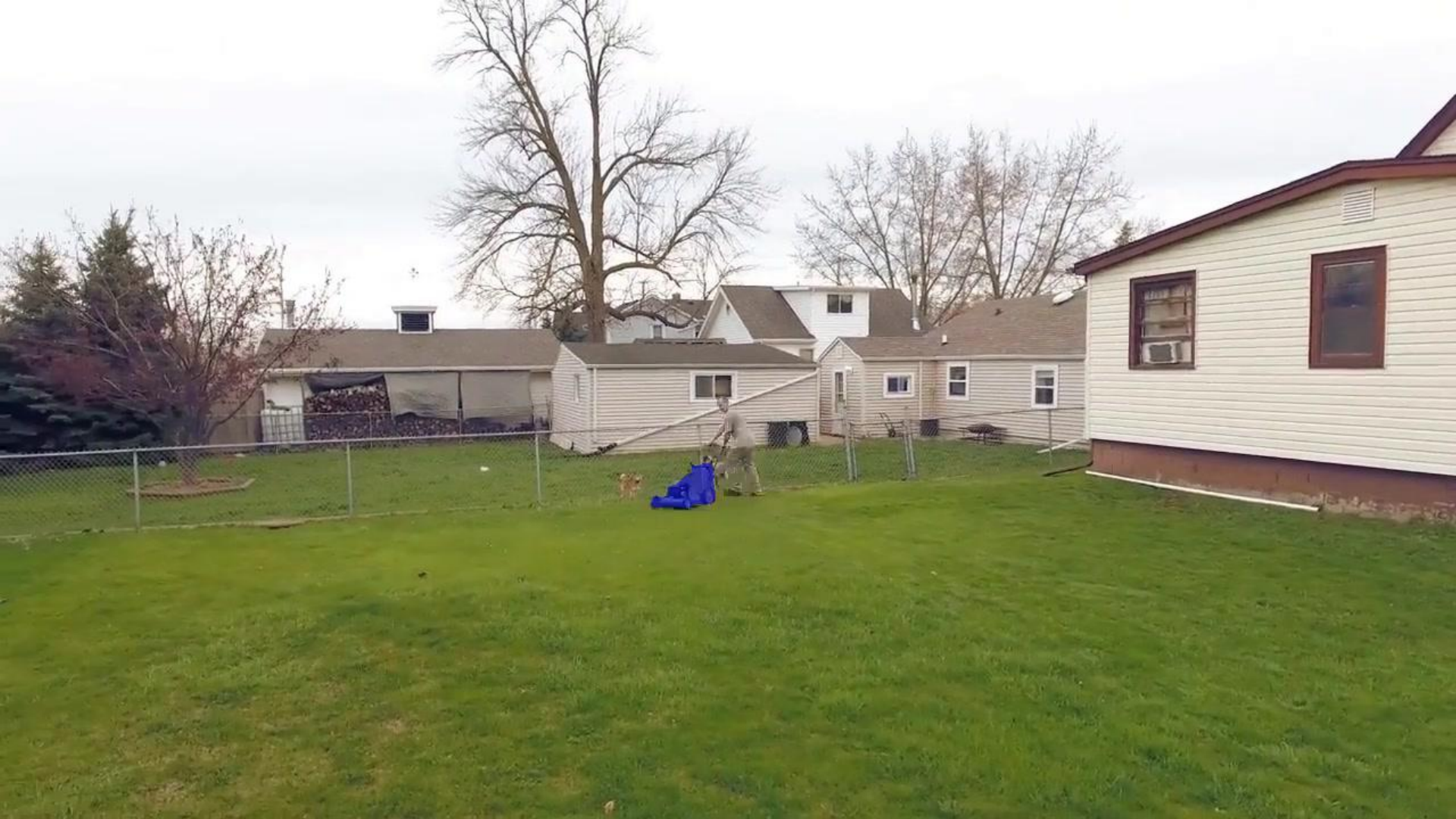}
        \vspace{-16pt}
    \end{subfigure}
    \vspace{5pt}
	
    \begin{subfigure}[b]{0.32\linewidth}
        \includegraphics[width=\mysize\linewidth]{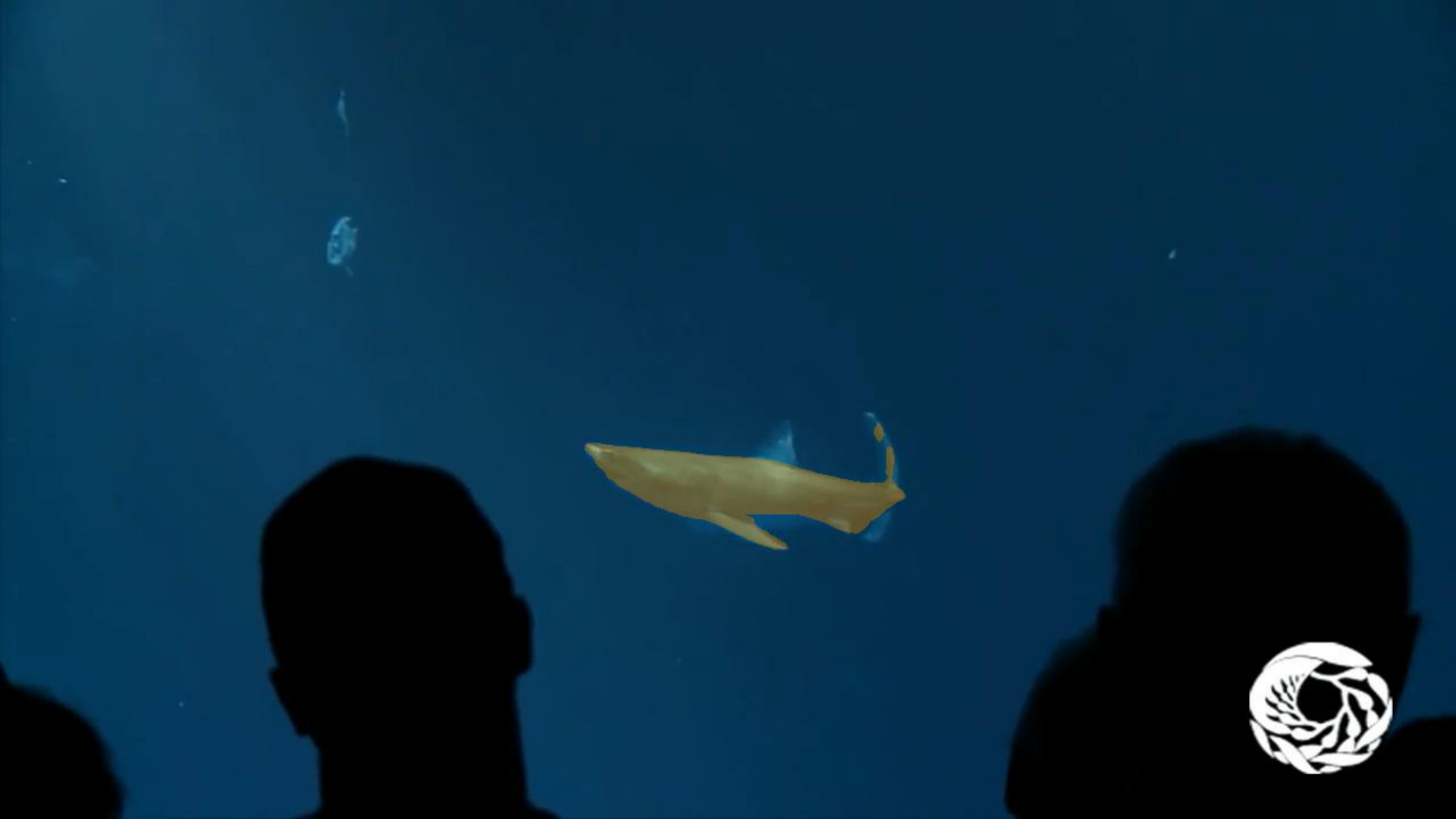}
        \vspace{-16pt}
    \end{subfigure}
    \begin{subfigure}[b]{0.32\linewidth}
        \includegraphics[width=\mysize\linewidth]{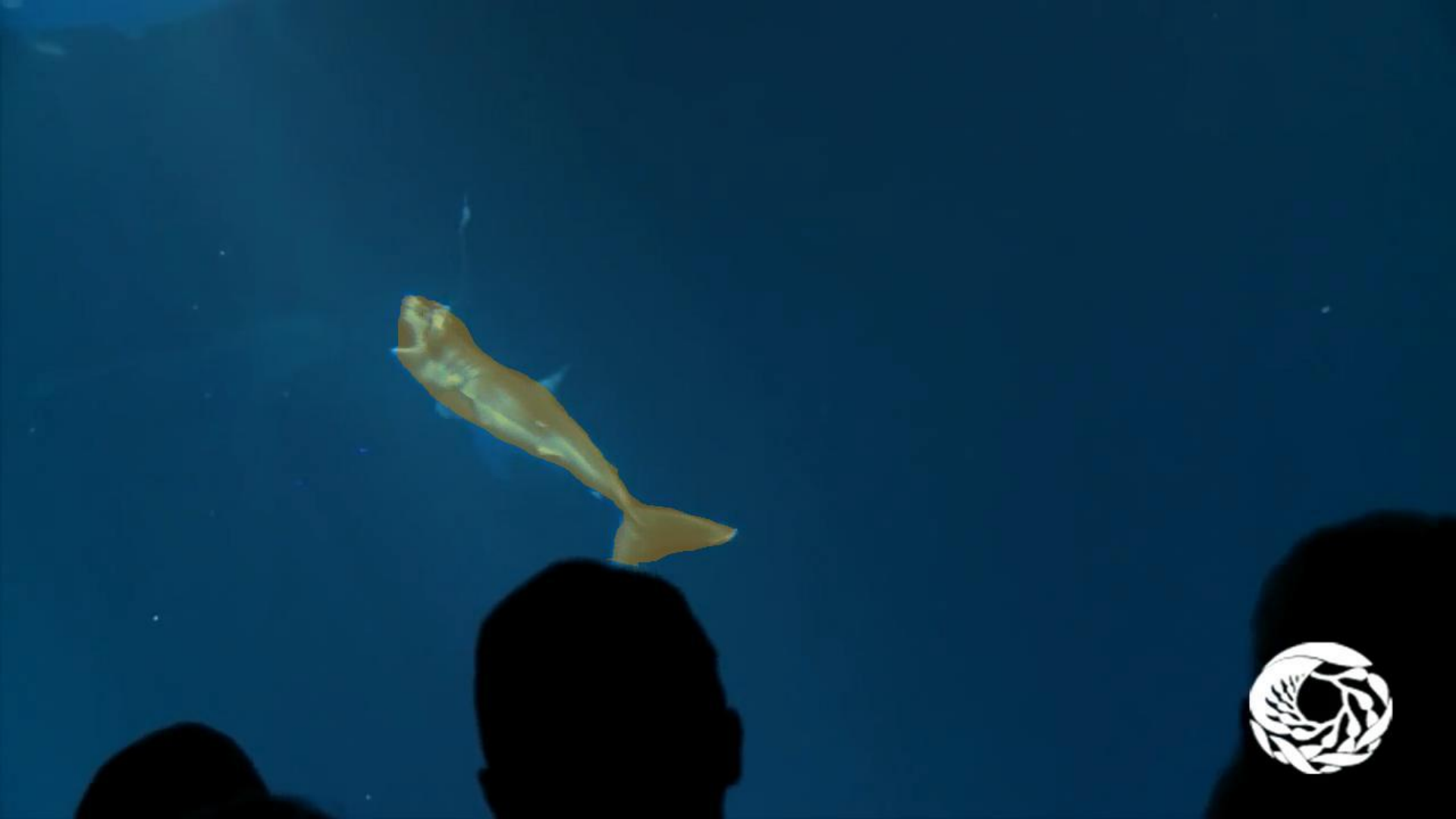}
        \vspace{-16pt}
    \end{subfigure}
        \begin{subfigure}[b]{0.32\linewidth}
        \includegraphics[width=\mysize\linewidth]{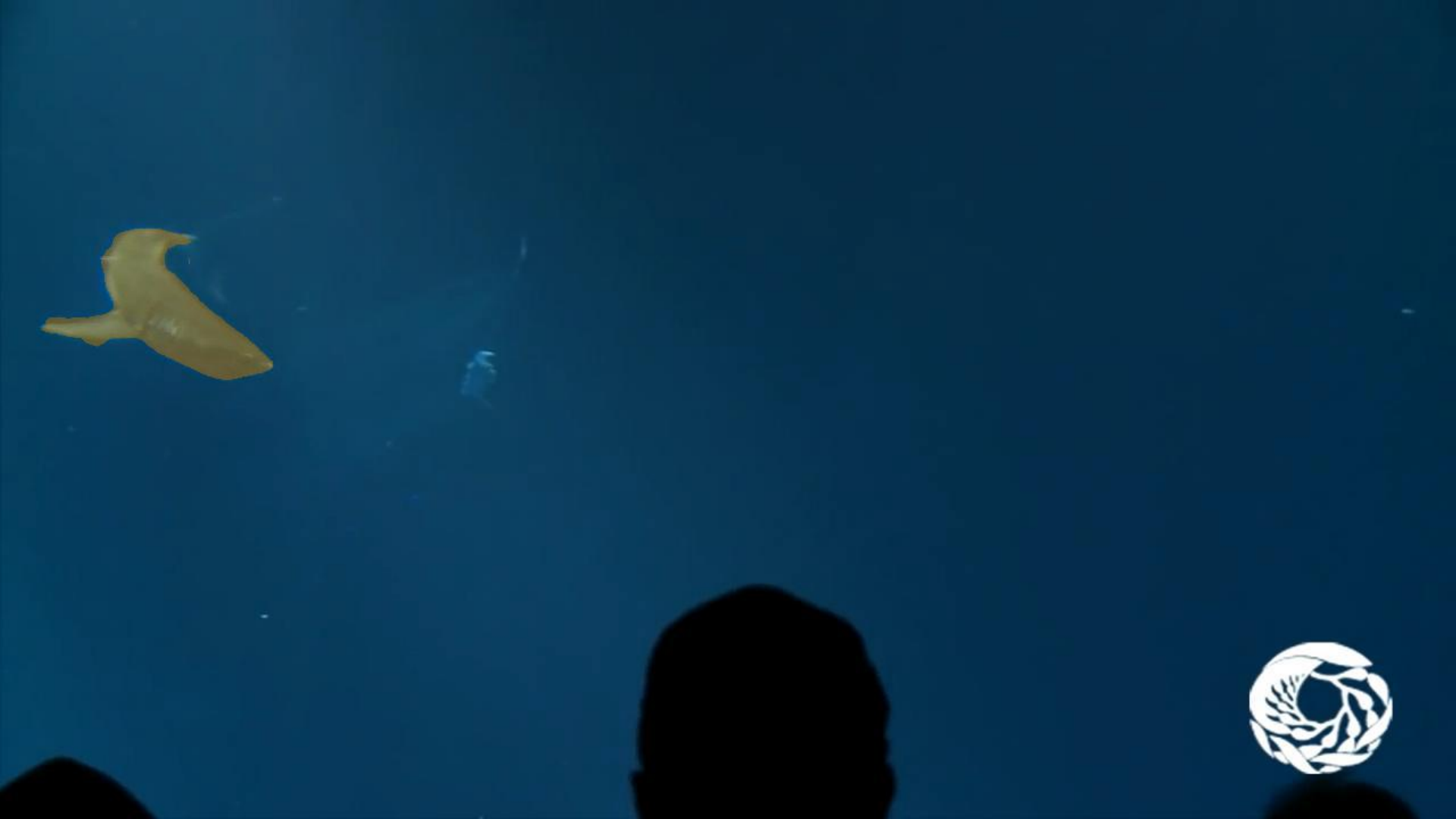}
        \vspace{-16pt}
    \end{subfigure}
    \vspace{5pt}
    
    \begin{subfigure}[b]{0.32\linewidth}
        \includegraphics[width=\mysize\linewidth]{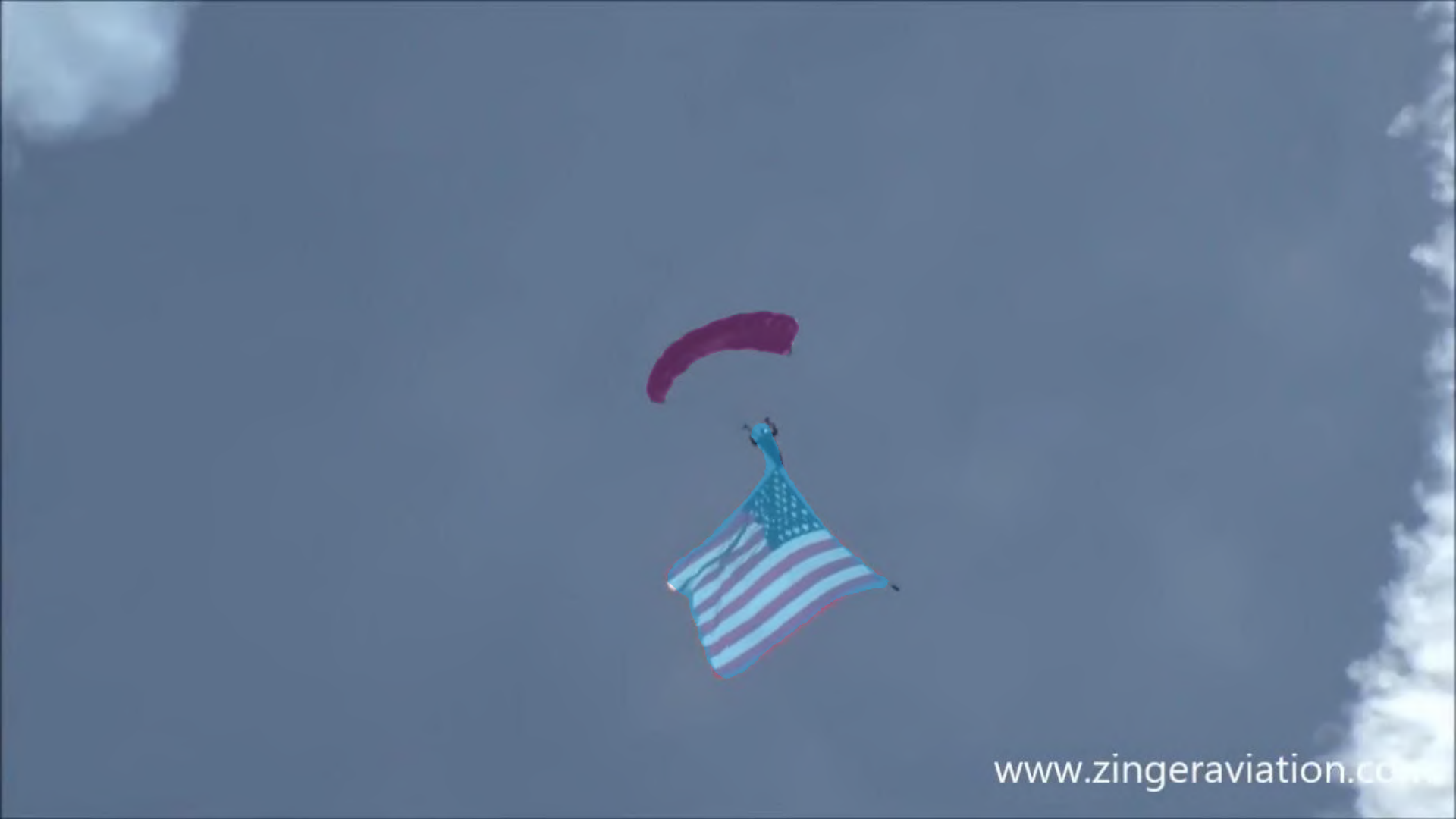}
        \vspace{-16pt}
    \end{subfigure}
    \begin{subfigure}[b]{0.32\linewidth}
        \includegraphics[width=\mysize\linewidth]{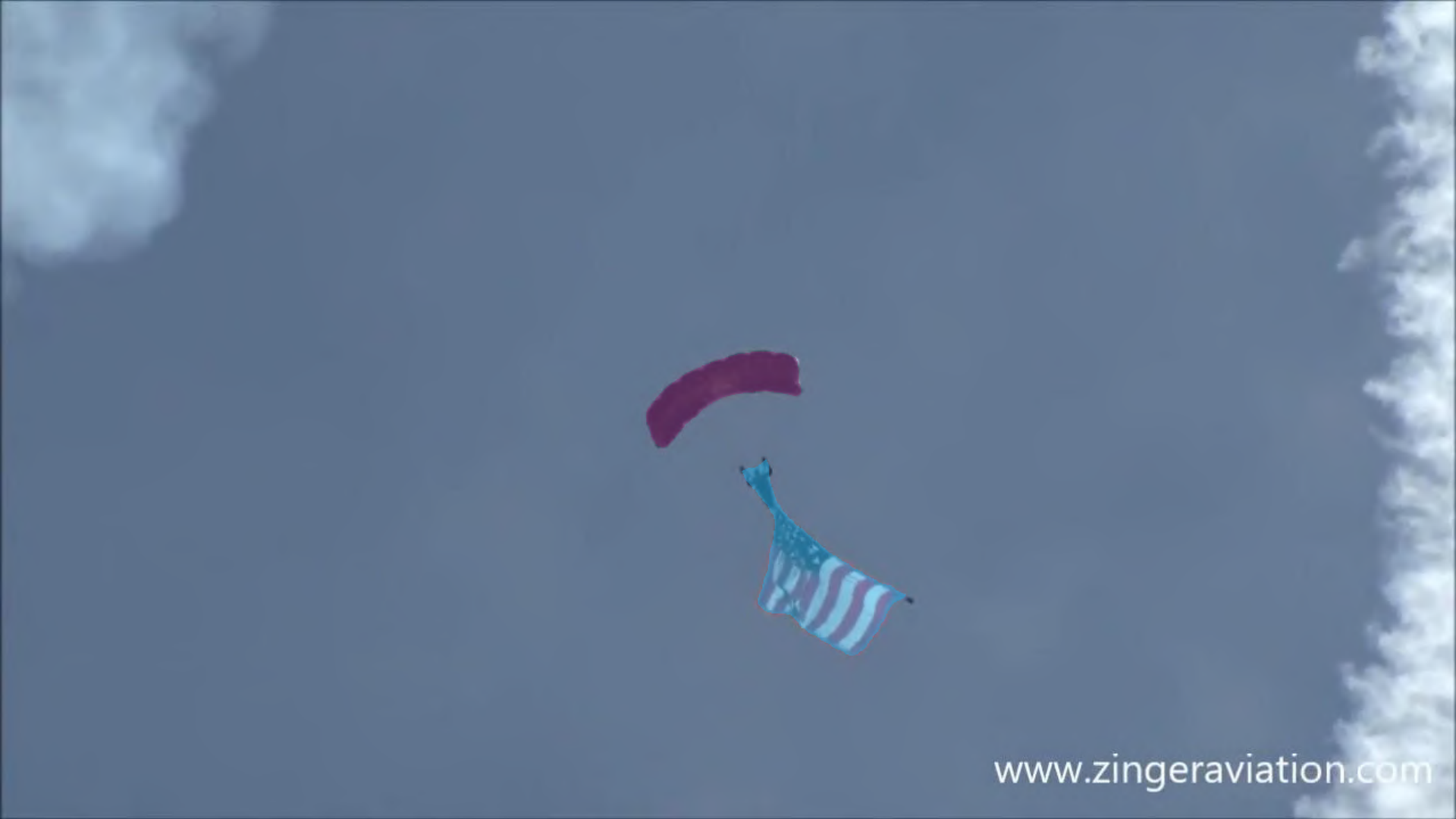}
        \vspace{-16pt}
    \end{subfigure}
    \begin{subfigure}[b]{0.32\linewidth}
     \includegraphics[width=\mysize\linewidth]{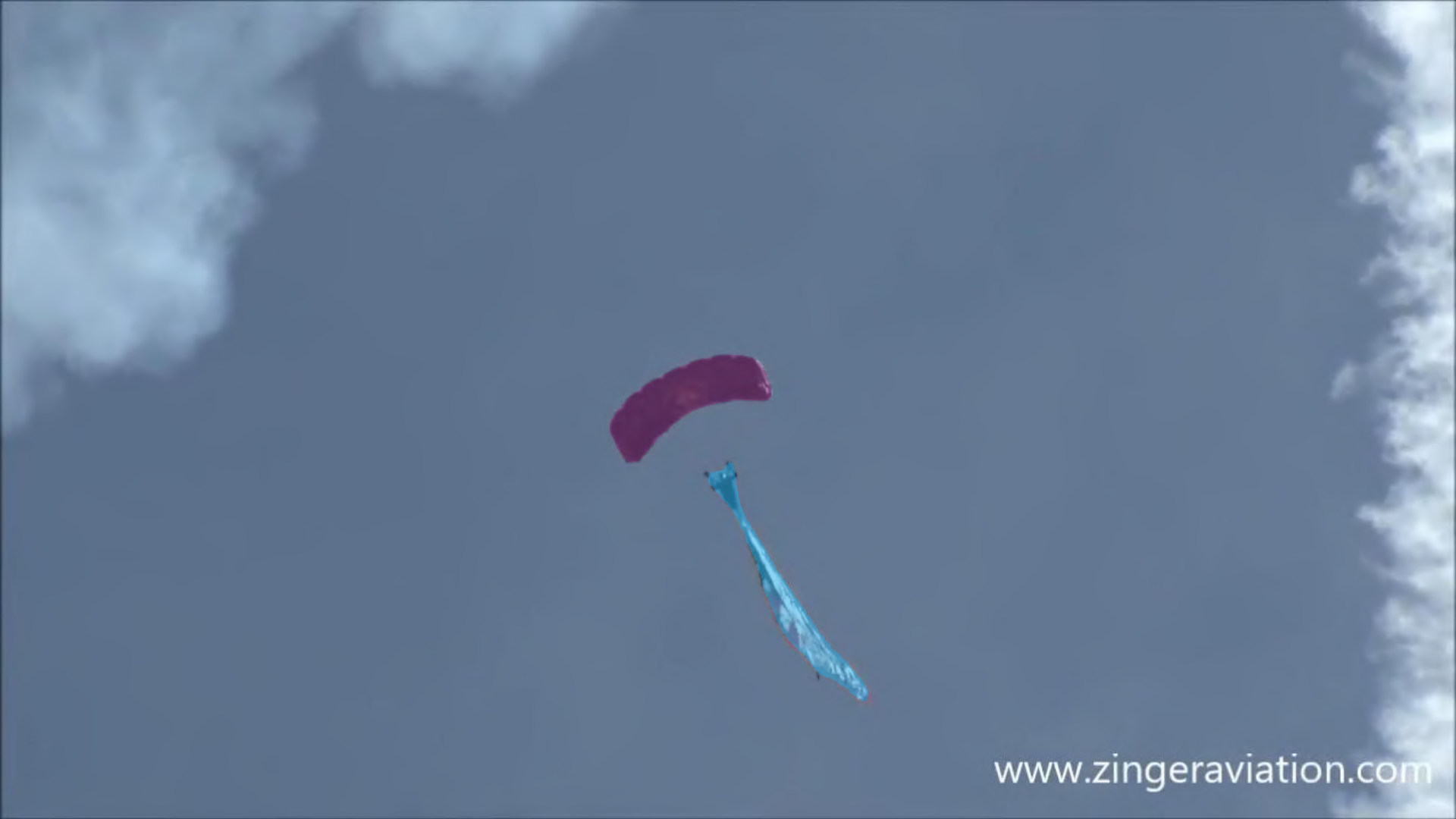}
        \vspace{-16pt}
    \end{subfigure} 
    \vspace{5pt}
    
    \begin{subfigure}[b]{0.32\linewidth}
        \includegraphics[width=\mysize\linewidth]{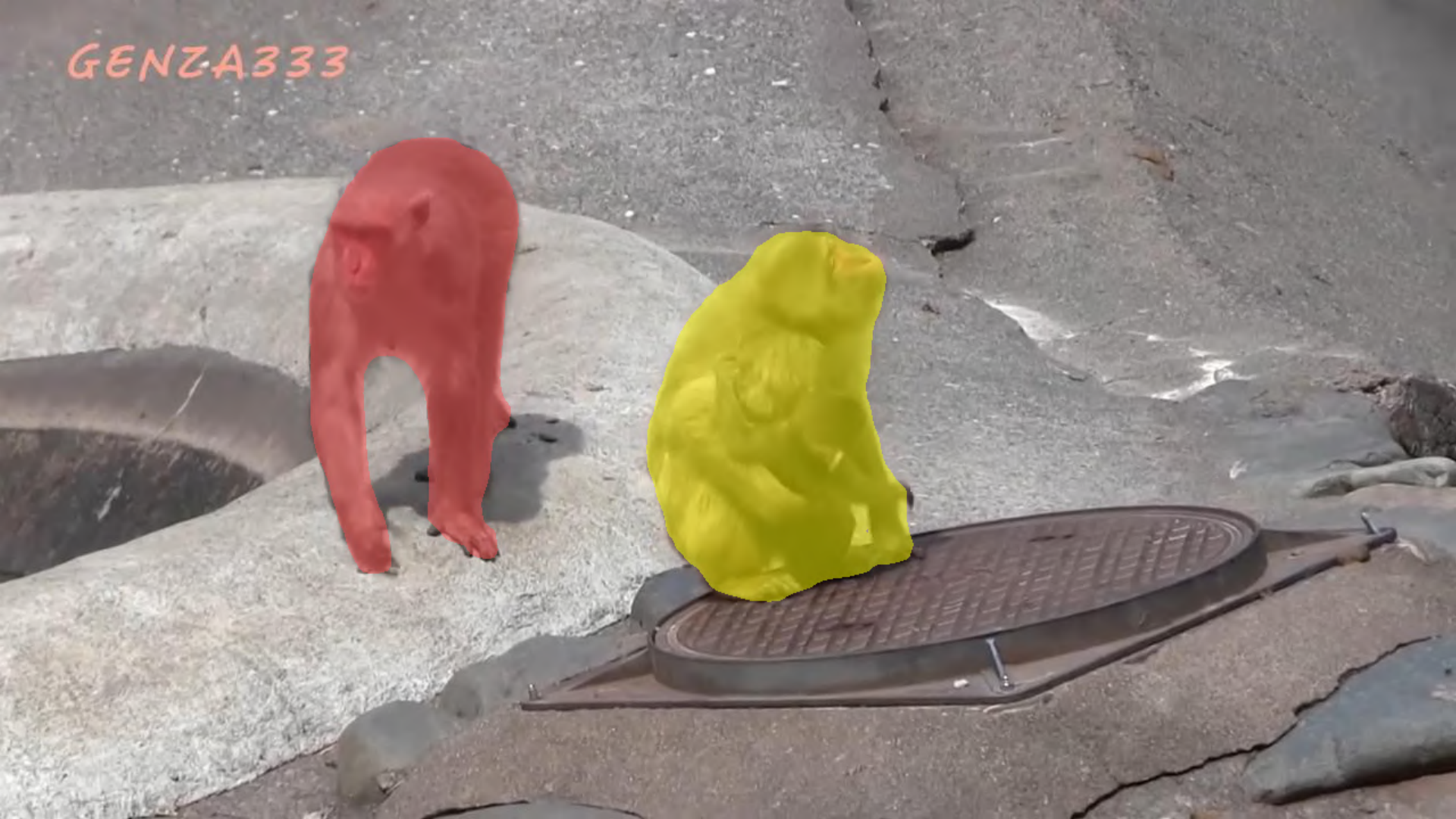}
        \vspace{-16pt}
    \end{subfigure}
    \begin{subfigure}[b]{0.32\linewidth}
        \includegraphics[width=\mysize\linewidth]{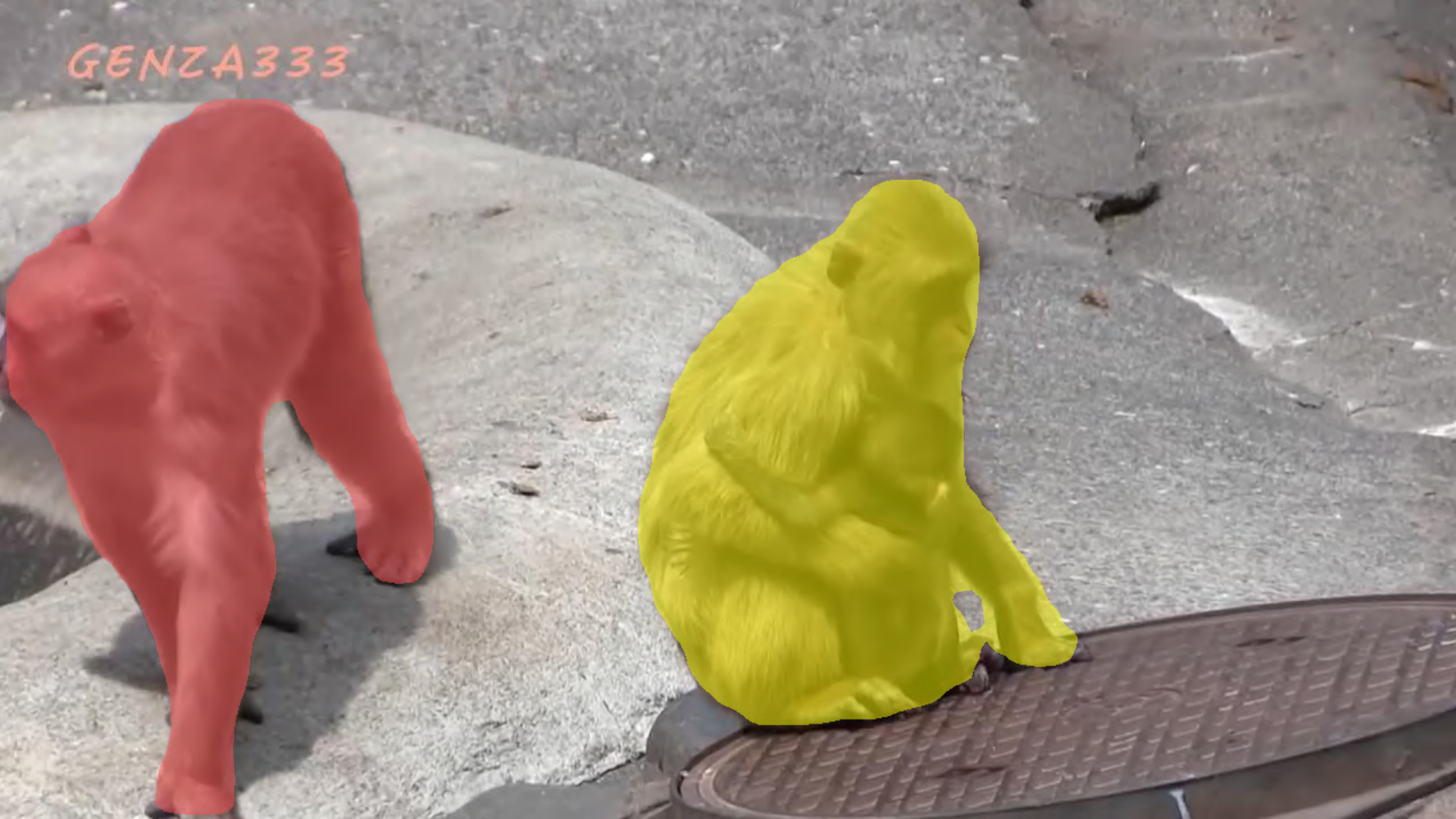}
        \vspace{-16pt}
    \end{subfigure}
        \begin{subfigure}[b]{0.32\linewidth}
        \includegraphics[width=\mysize\linewidth]{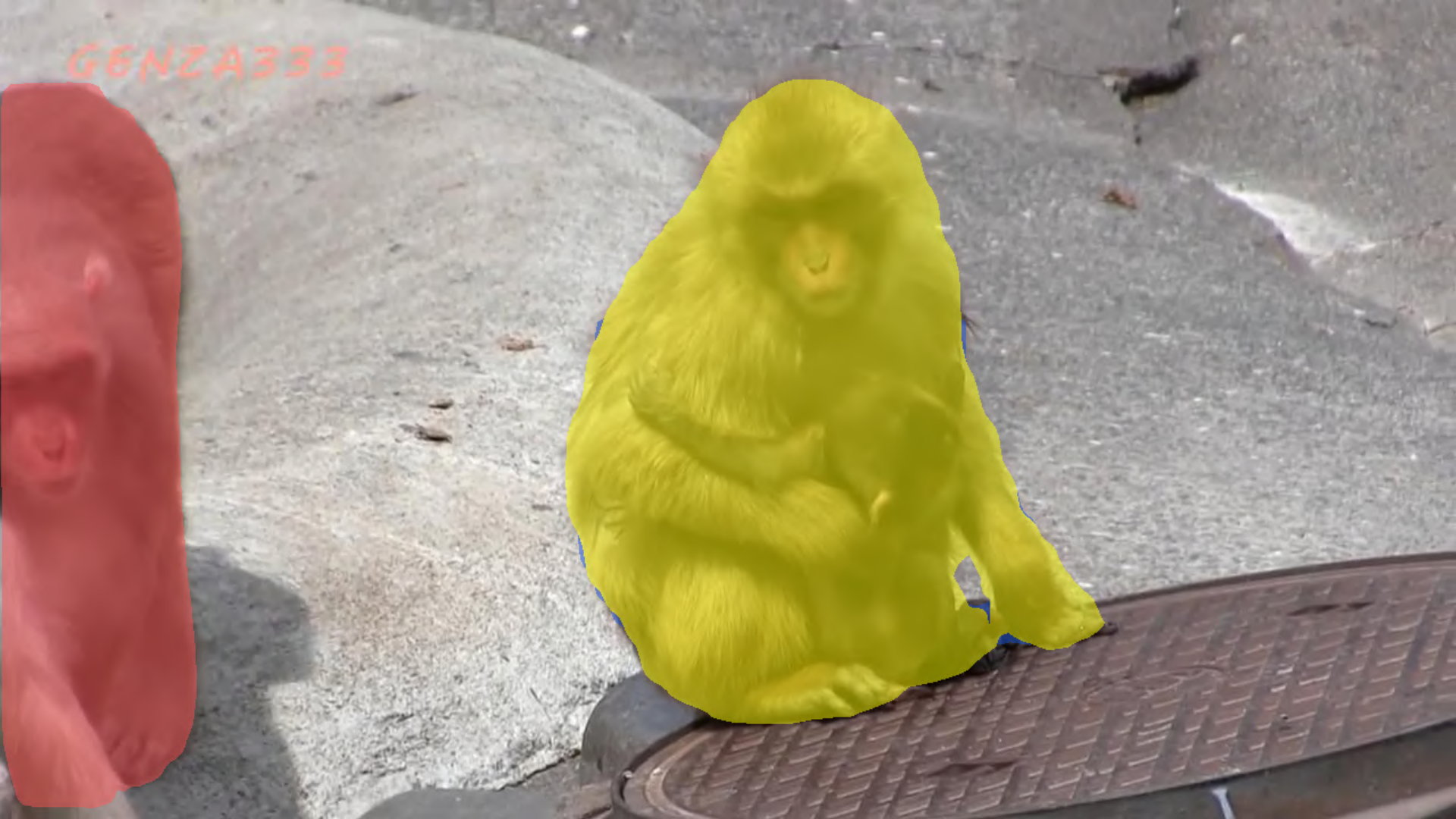}
        \vspace{-16pt}
    \end{subfigure}
    \vspace{5pt}
    
    \begin{subfigure}[b]{0.32\linewidth}
        \includegraphics[width=\mysize\linewidth]{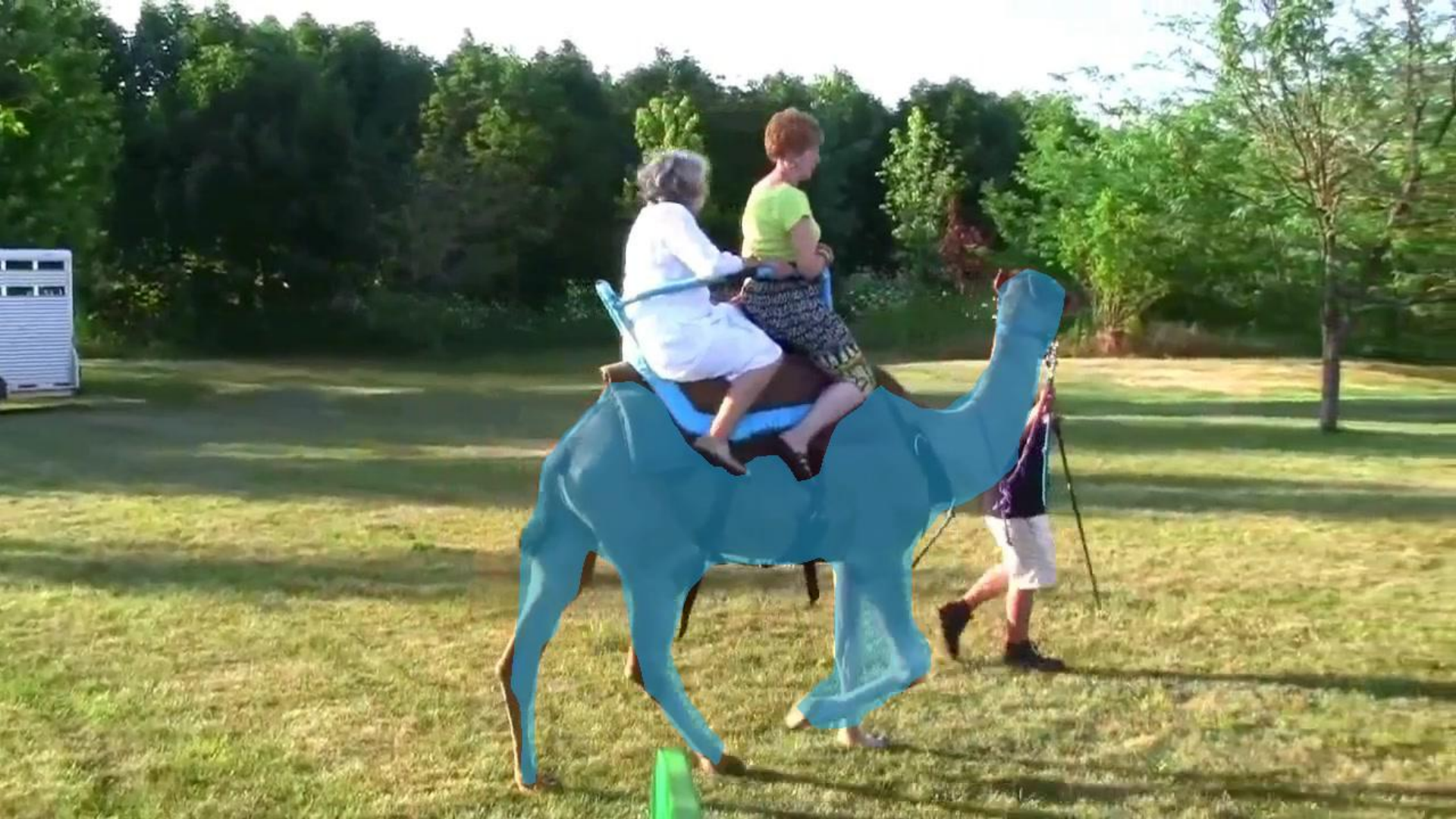}
        \vspace{-16pt}
    \end{subfigure}
    \begin{subfigure}[b]{0.32\linewidth}
        \includegraphics[width=\mysize\linewidth]{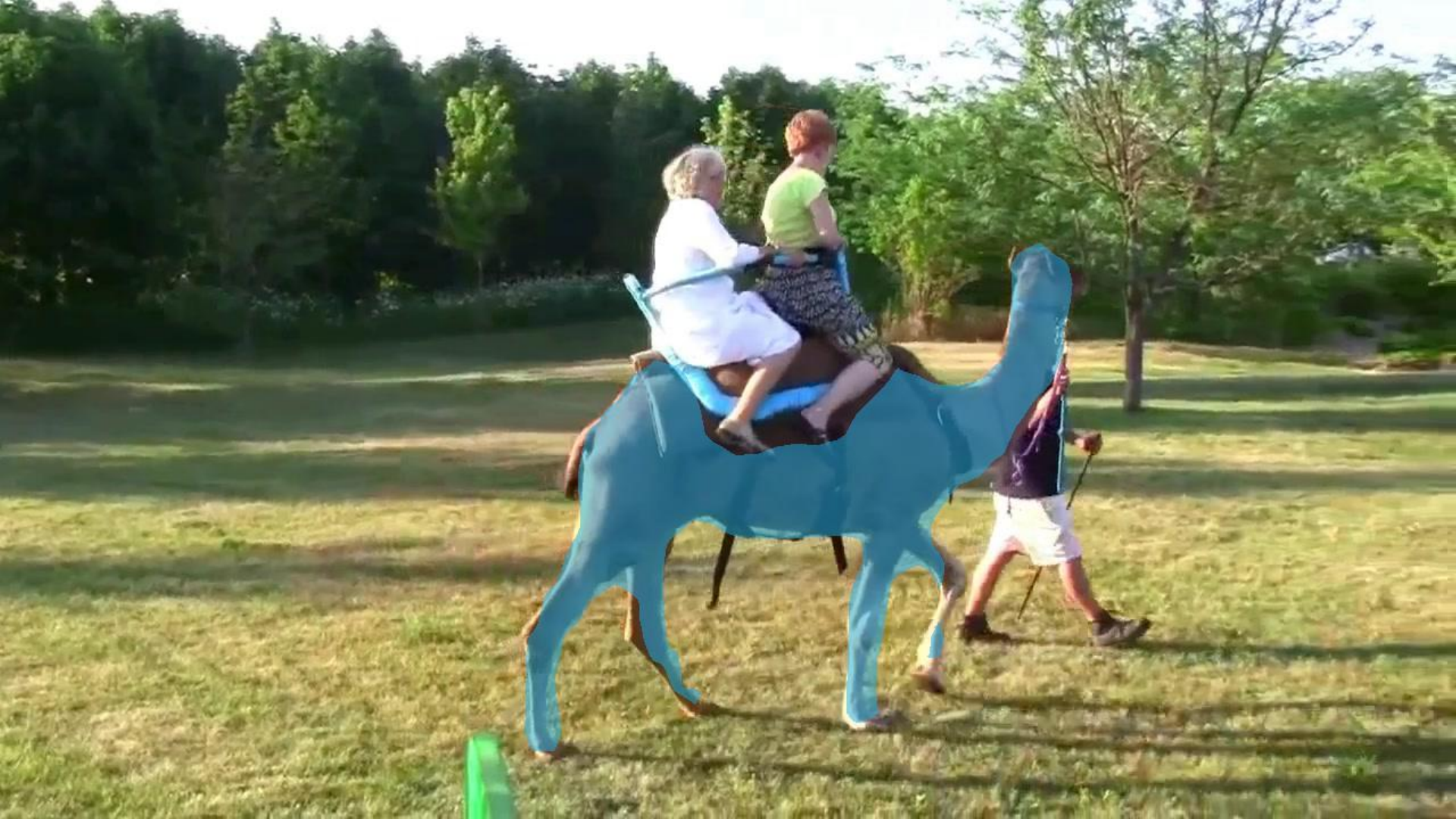}
        \vspace{-16pt}
    \end{subfigure}
        \begin{subfigure}[b]{0.32\linewidth}
        \includegraphics[width=\mysize\linewidth]{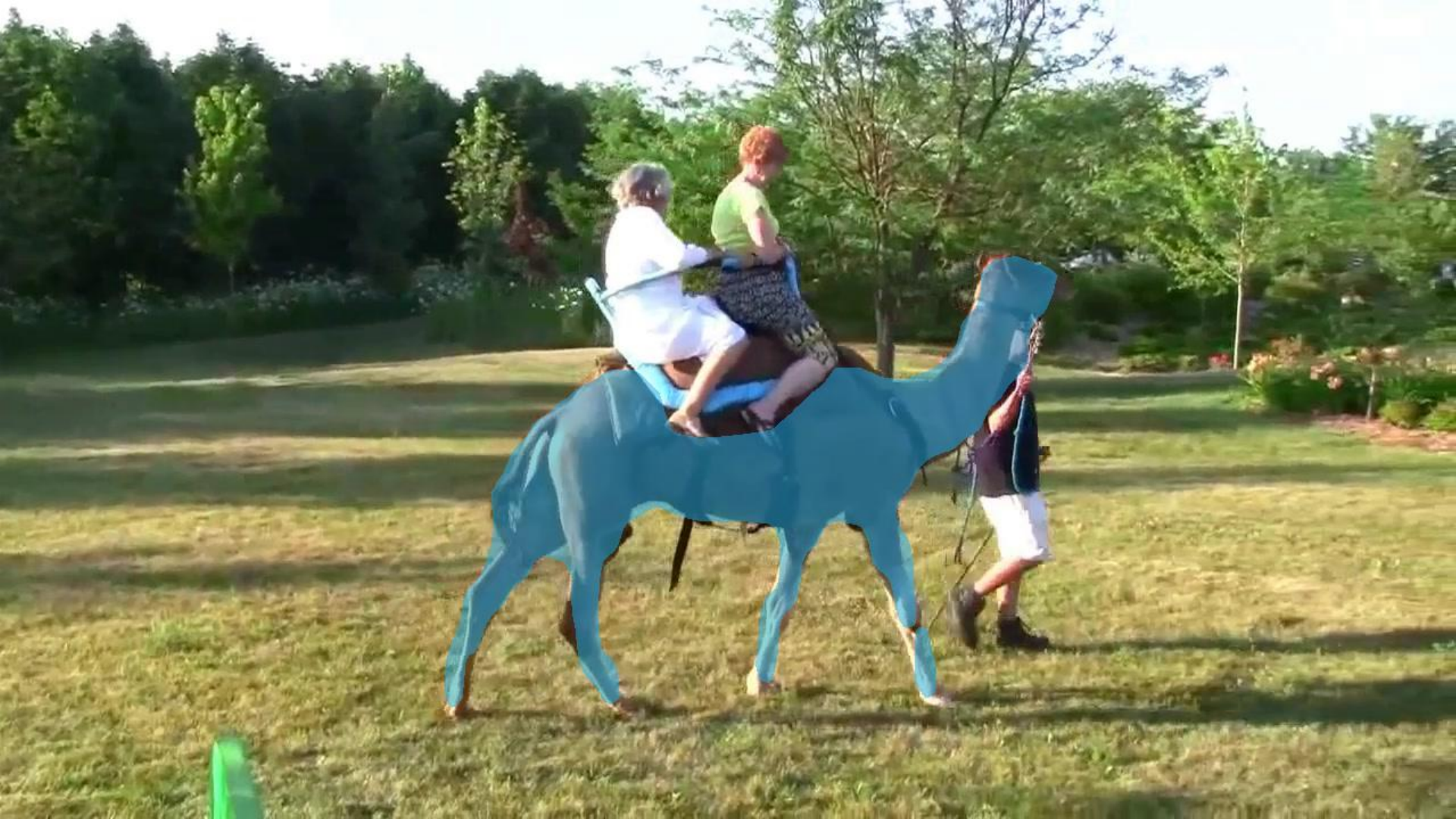}
        \vspace{-16pt}
    \end{subfigure}
    \vspace{5pt}
    
    \begin{subfigure}[b]{0.32\linewidth}
        \includegraphics[width=\mysize\linewidth]{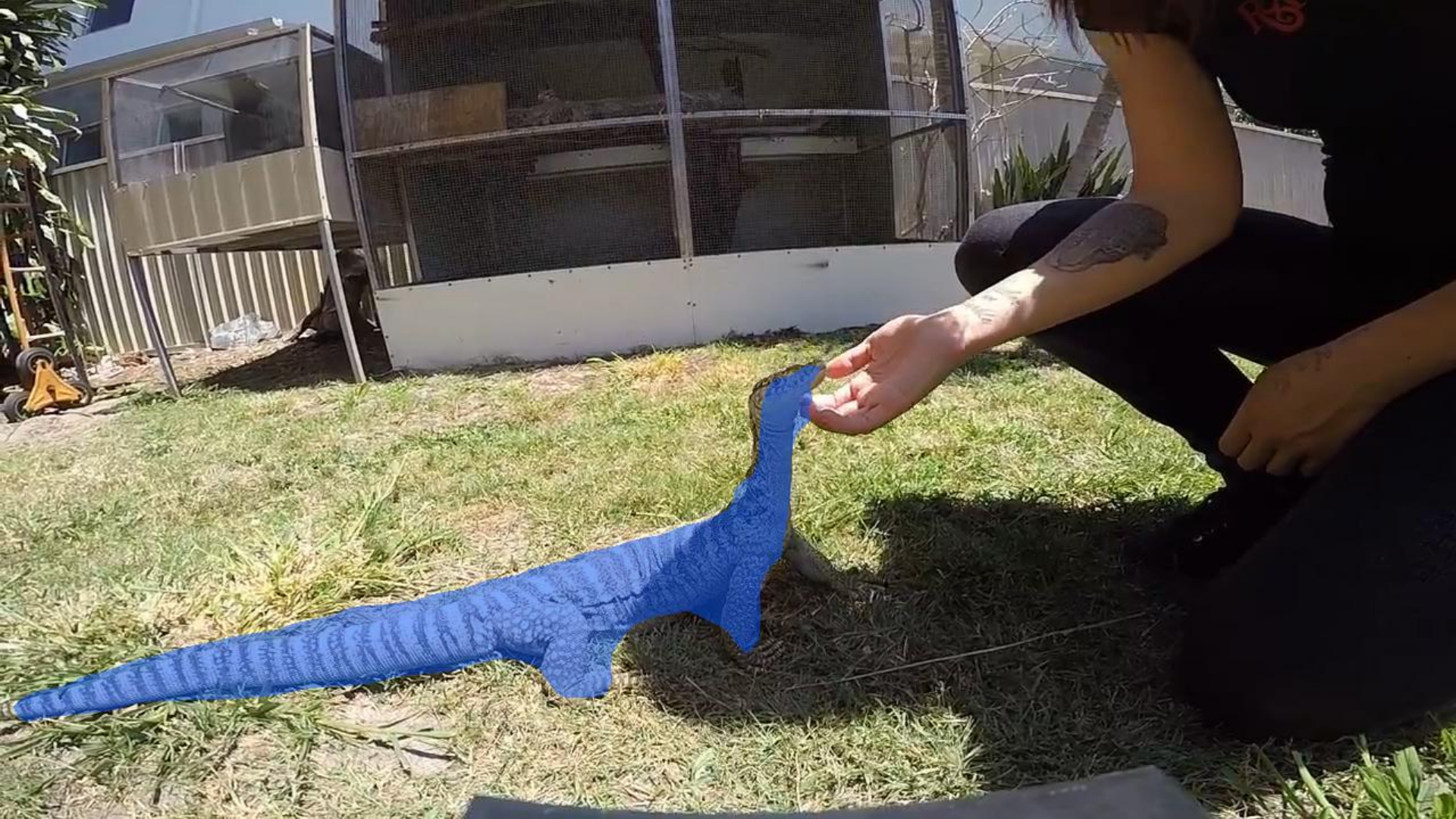}
        \vspace{-16pt}
    \end{subfigure}
    \begin{subfigure}[b]{0.32\linewidth}
        \includegraphics[width=\mysize\linewidth]{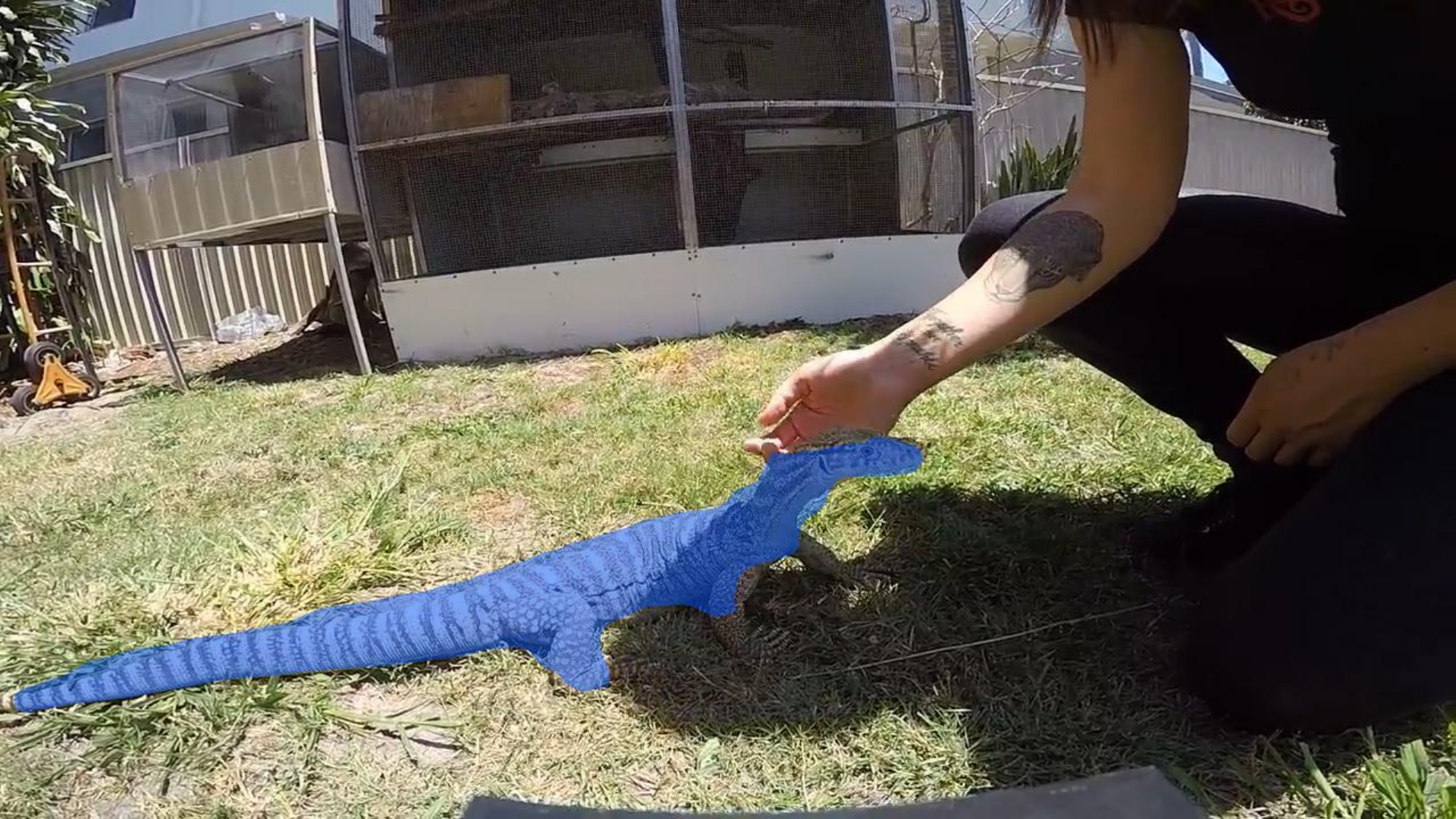}
        \vspace{-16pt}
    \end{subfigure}
        \begin{subfigure}[b]{0.32\linewidth}
        \includegraphics[width=\mysize\linewidth]{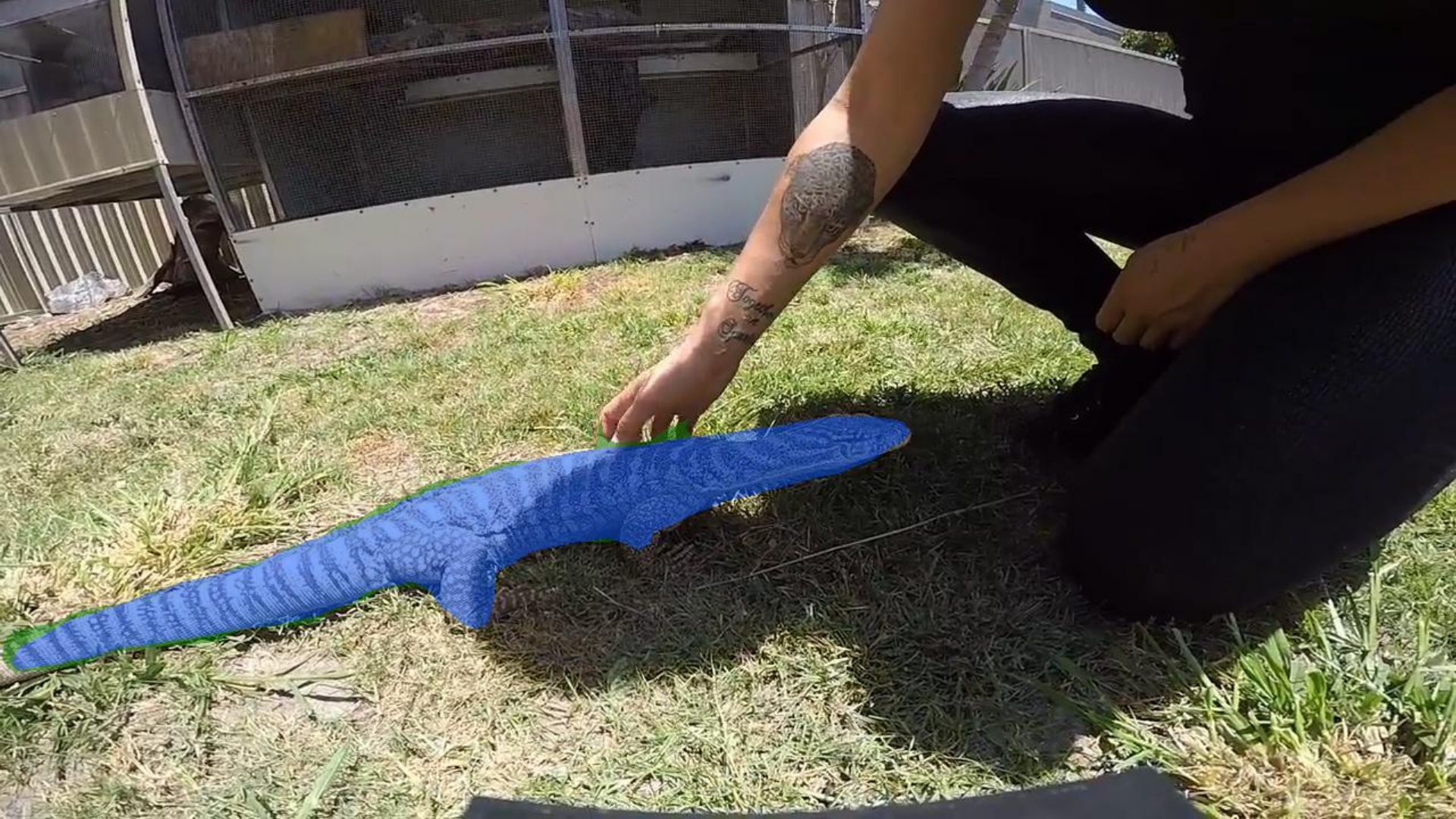}
        \vspace{-16pt}
    \end{subfigure}
    \vspace{5pt}
    
    \caption{\textbf{Tracking results for \unknown.} Examples of \unknown objects tracked by OWTB. OWTB is robust for motion blur (\textit{first row, third image}) and large motions (\textit{second row}).}
    \label{fig:tracking_unknown}
\end{figure*}

\begin{figure*}[t]
    \centering
    \setlength{\fboxsep}{0.32pt}%
    \begin{subfigure}[b]{0.32\linewidth}
        \includegraphics[width=\mysize\linewidth]{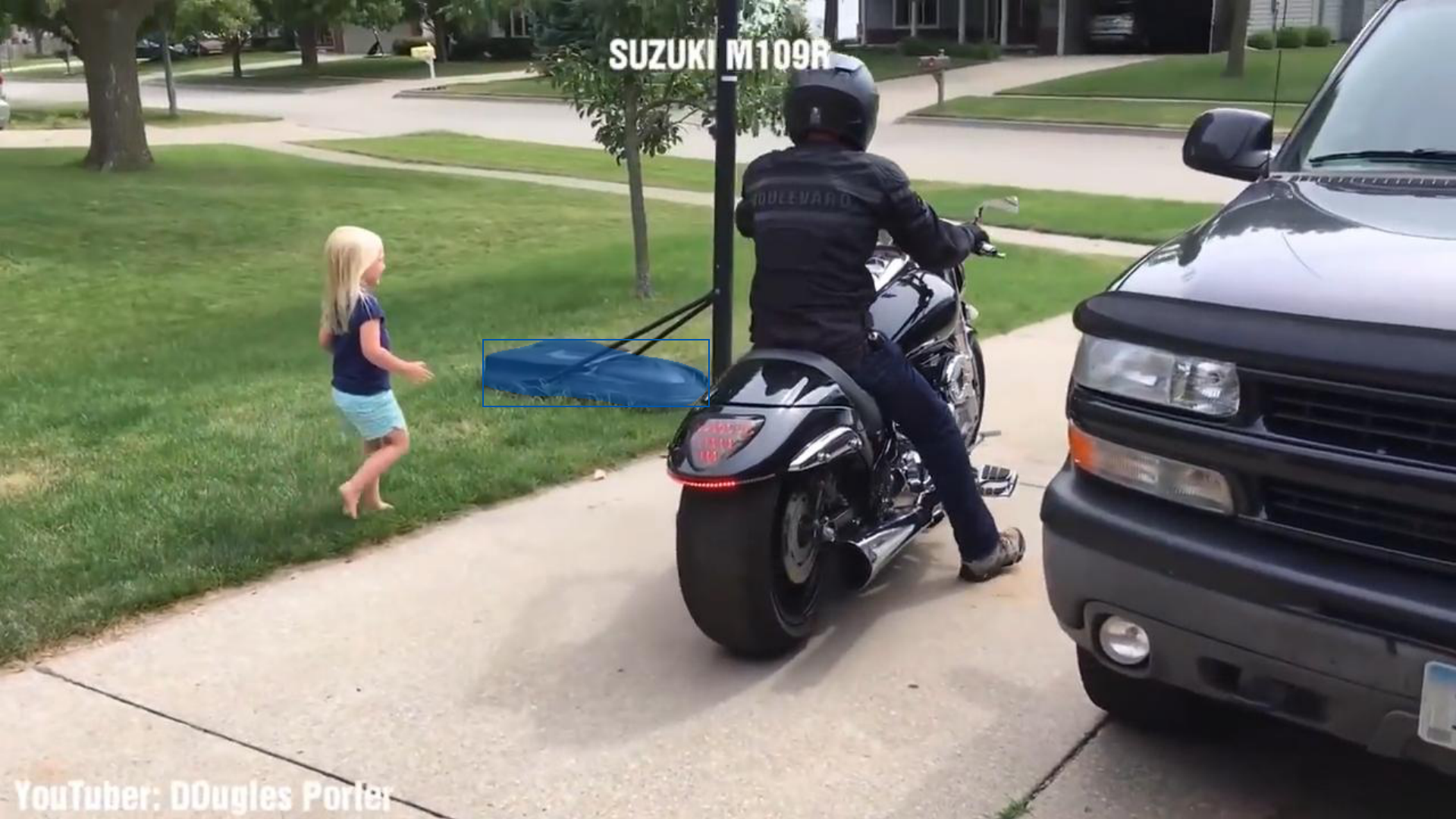}
        \vspace{-16pt}
    \end{subfigure}
    \begin{subfigure}[b]{0.32\linewidth}
        \includegraphics[width=\mysize\linewidth]{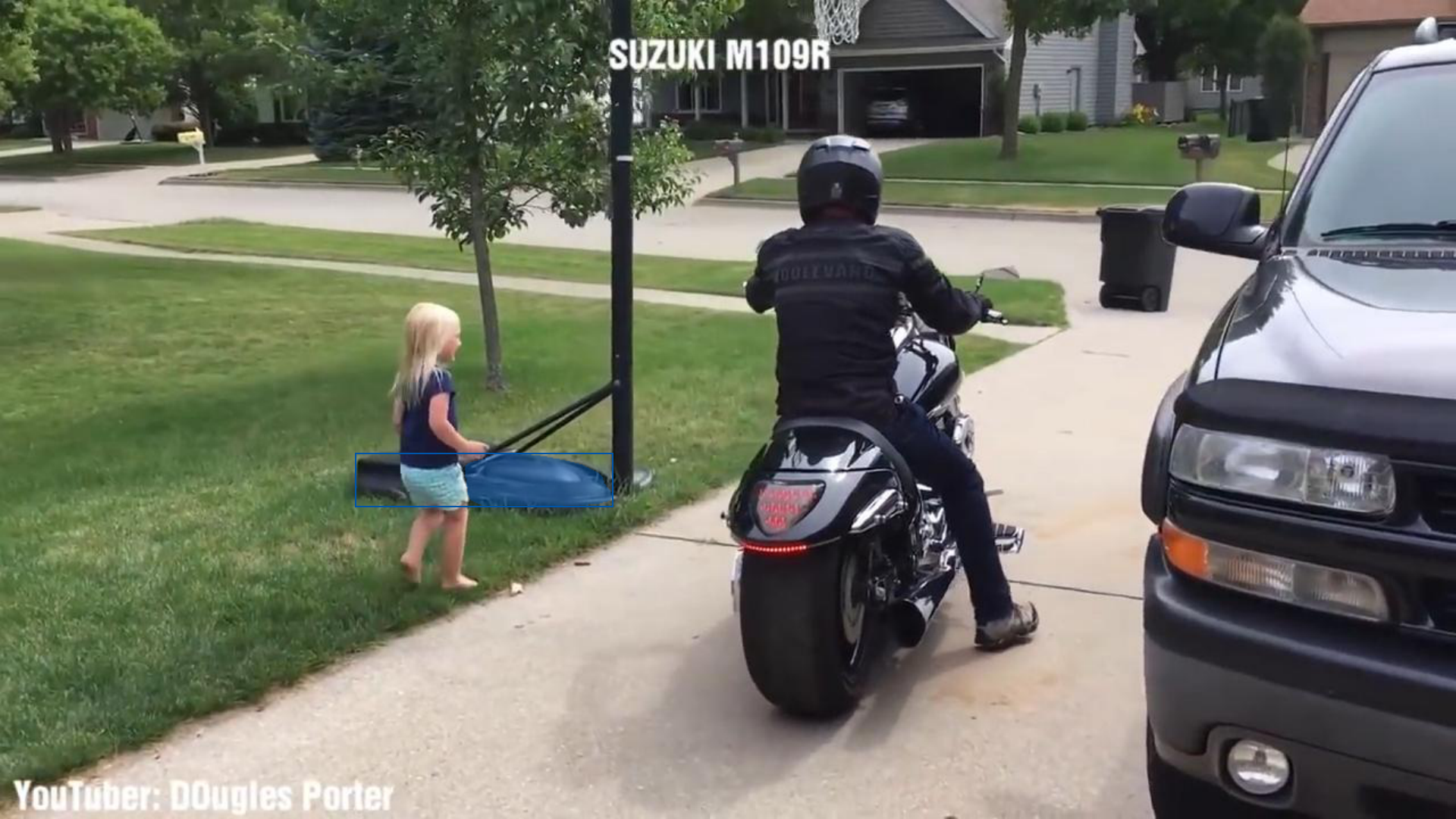}
        \vspace{-16pt}
    \end{subfigure}
    \begin{subfigure}[b]{0.32\linewidth}
     \includegraphics[width=\mysize\linewidth]{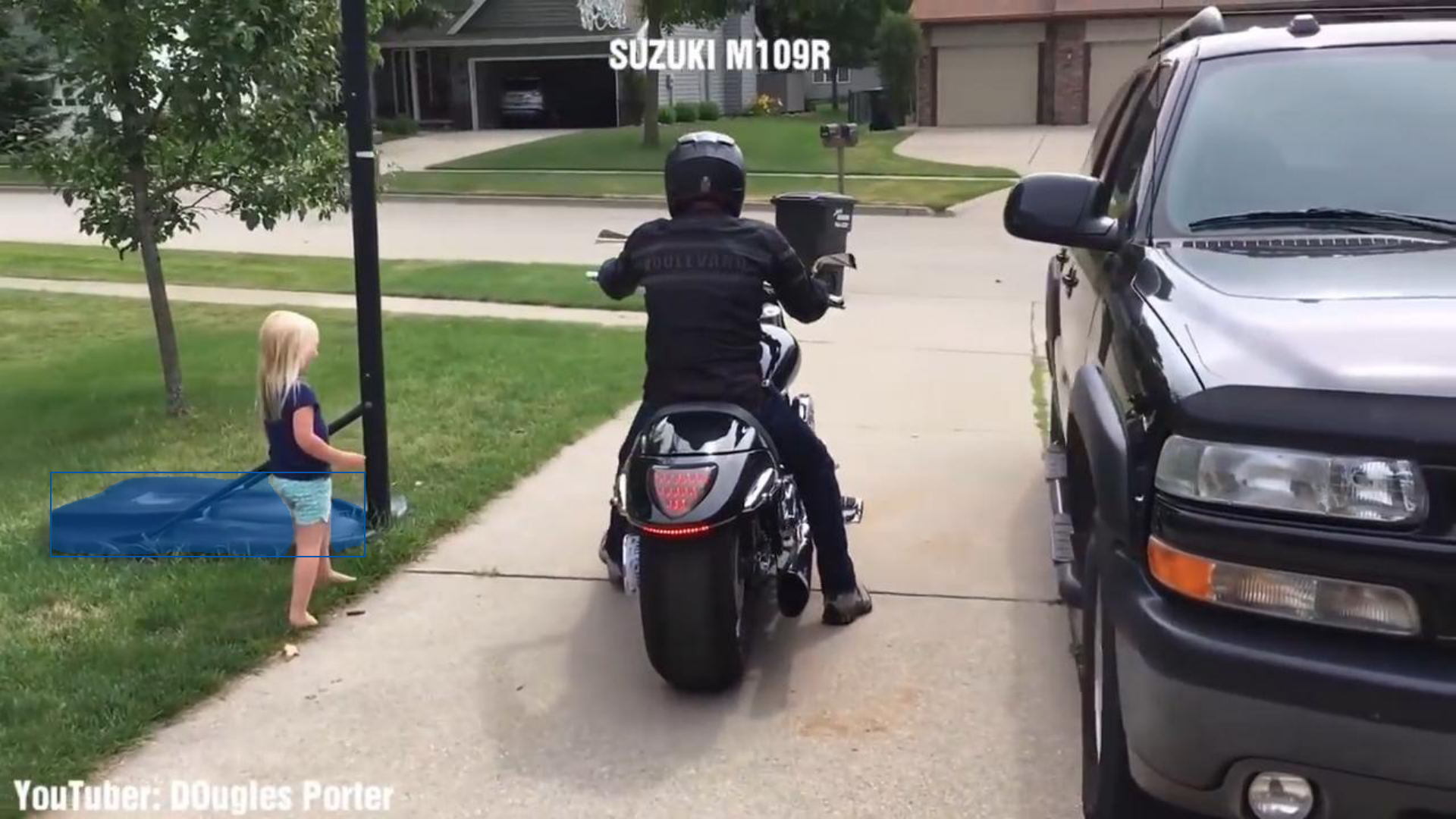}
        \vspace{-16pt}
    \end{subfigure} 
    \vspace{5pt}
	
    \begin{subfigure}[b]{0.32\linewidth}
        \includegraphics[width=\mysize\linewidth]{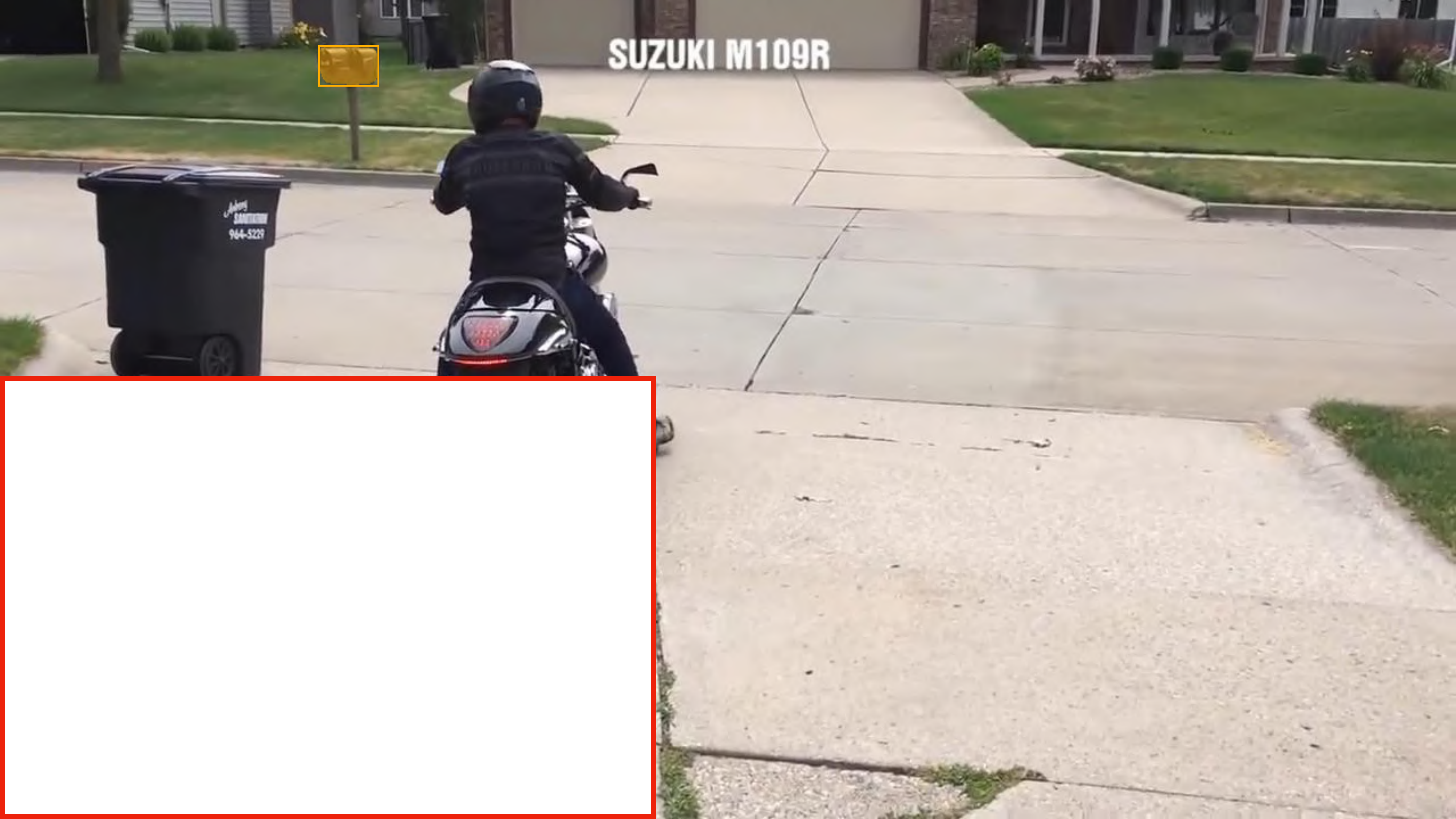}
        \vspace{-16pt}
    \end{subfigure}
    \begin{subfigure}[b]{0.32\linewidth}
        \includegraphics[width=\mysize\linewidth]{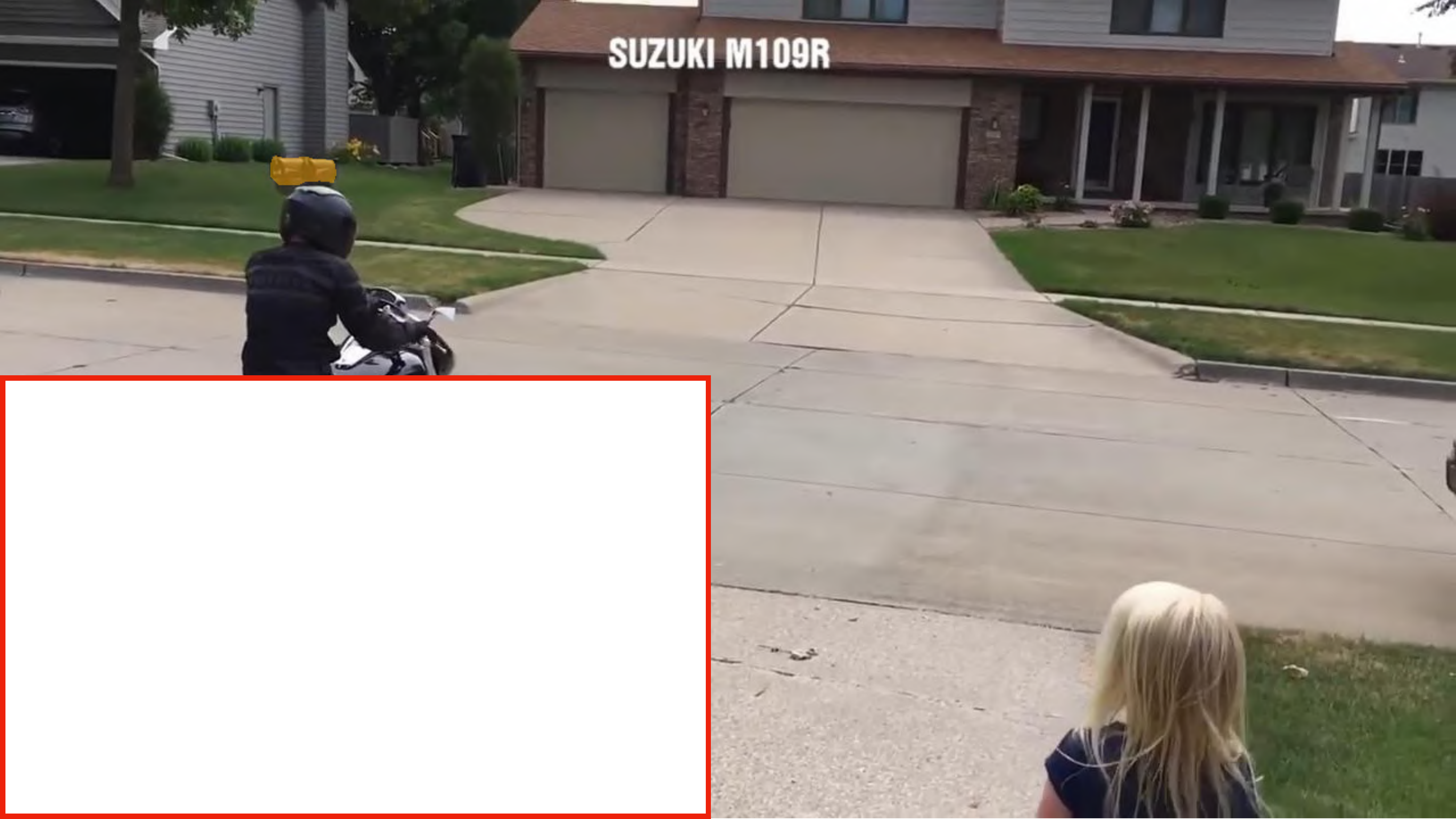}
        \vspace{-16pt}
    \end{subfigure}
        \begin{subfigure}[b]{0.32\linewidth}
        \includegraphics[width=\mysize\linewidth]{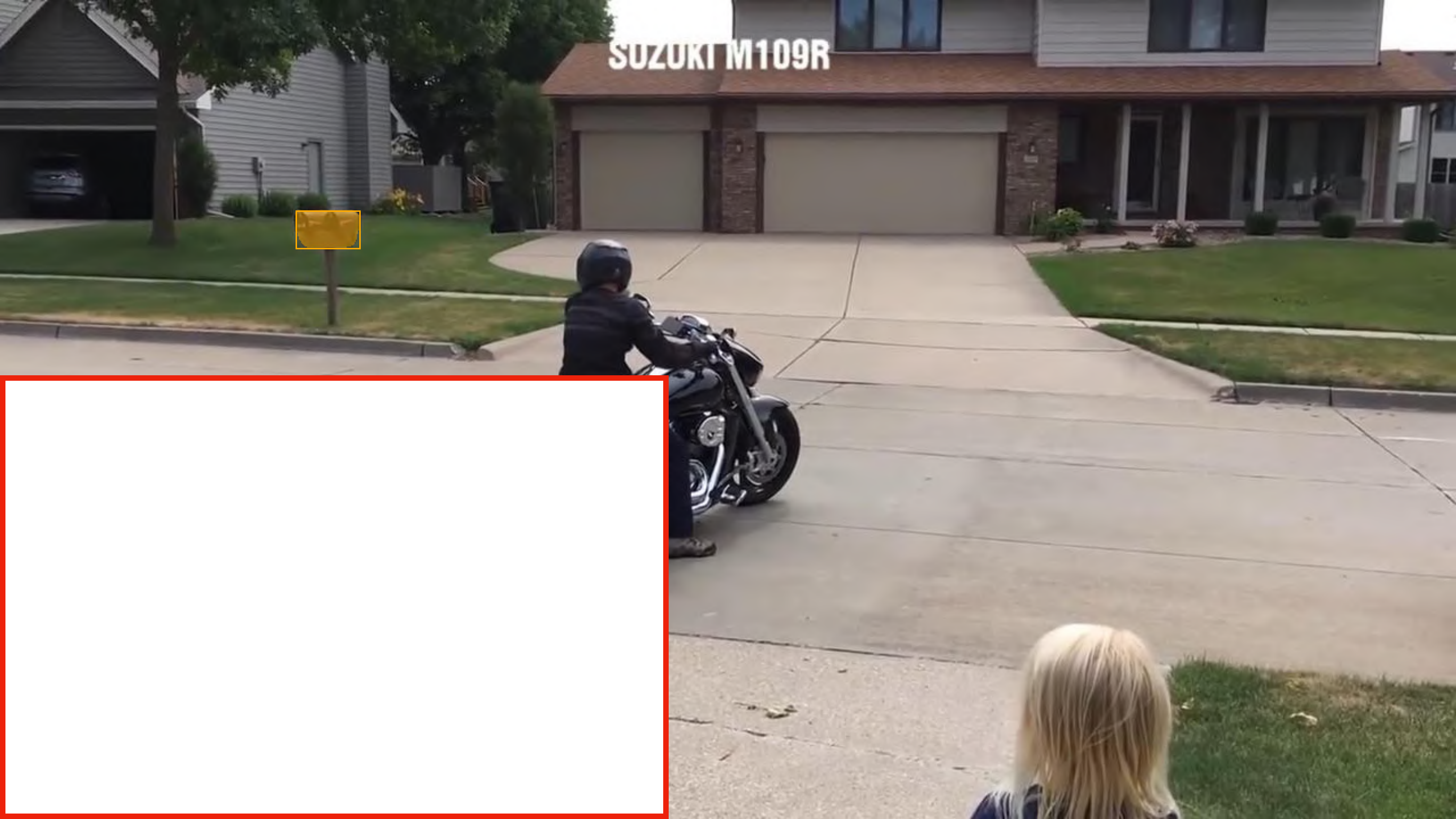}
        \vspace{-16pt}
    \end{subfigure}
    \vspace{5pt}
	
    \begin{subfigure}[b]{0.32\linewidth}
        \includegraphics[width=\mysize\linewidth]{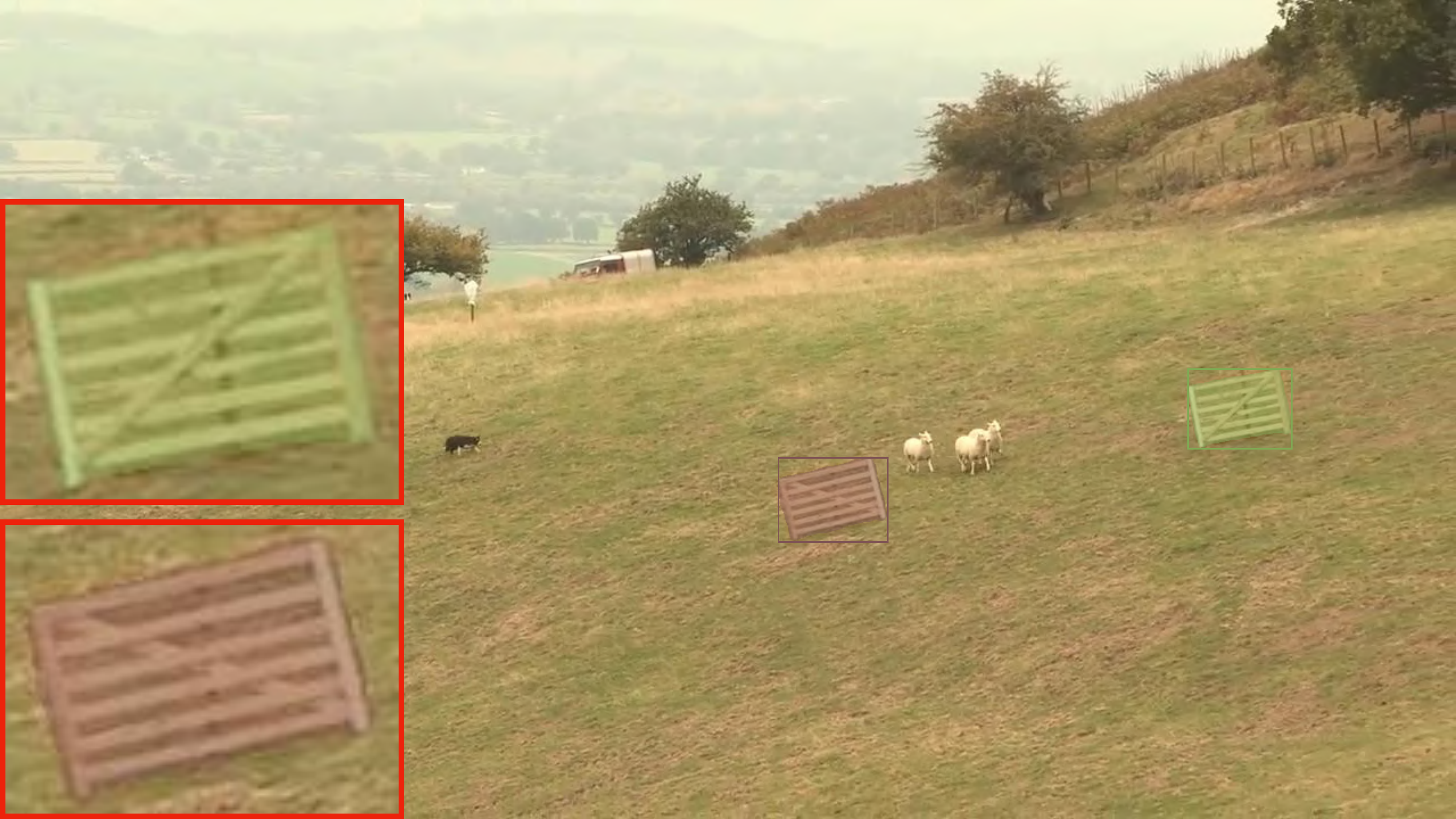}
        \vspace{-16pt}
    \end{subfigure}
    \begin{subfigure}[b]{0.32\linewidth}
        \includegraphics[width=\mysize\linewidth]{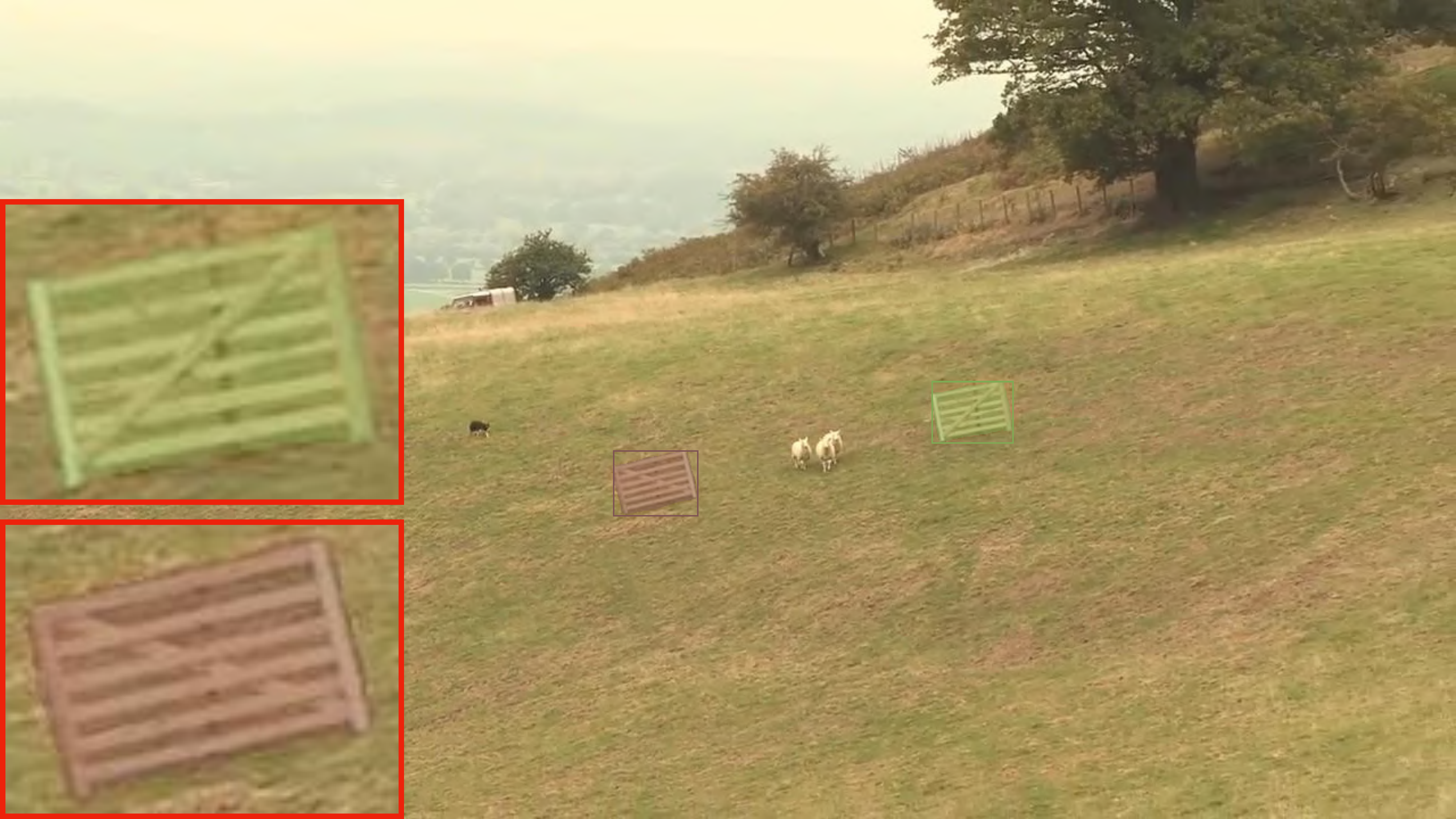}
        \vspace{-16pt}
    \end{subfigure}
        \begin{subfigure}[b]{0.32\linewidth}
        \includegraphics[width=\mysize\linewidth]{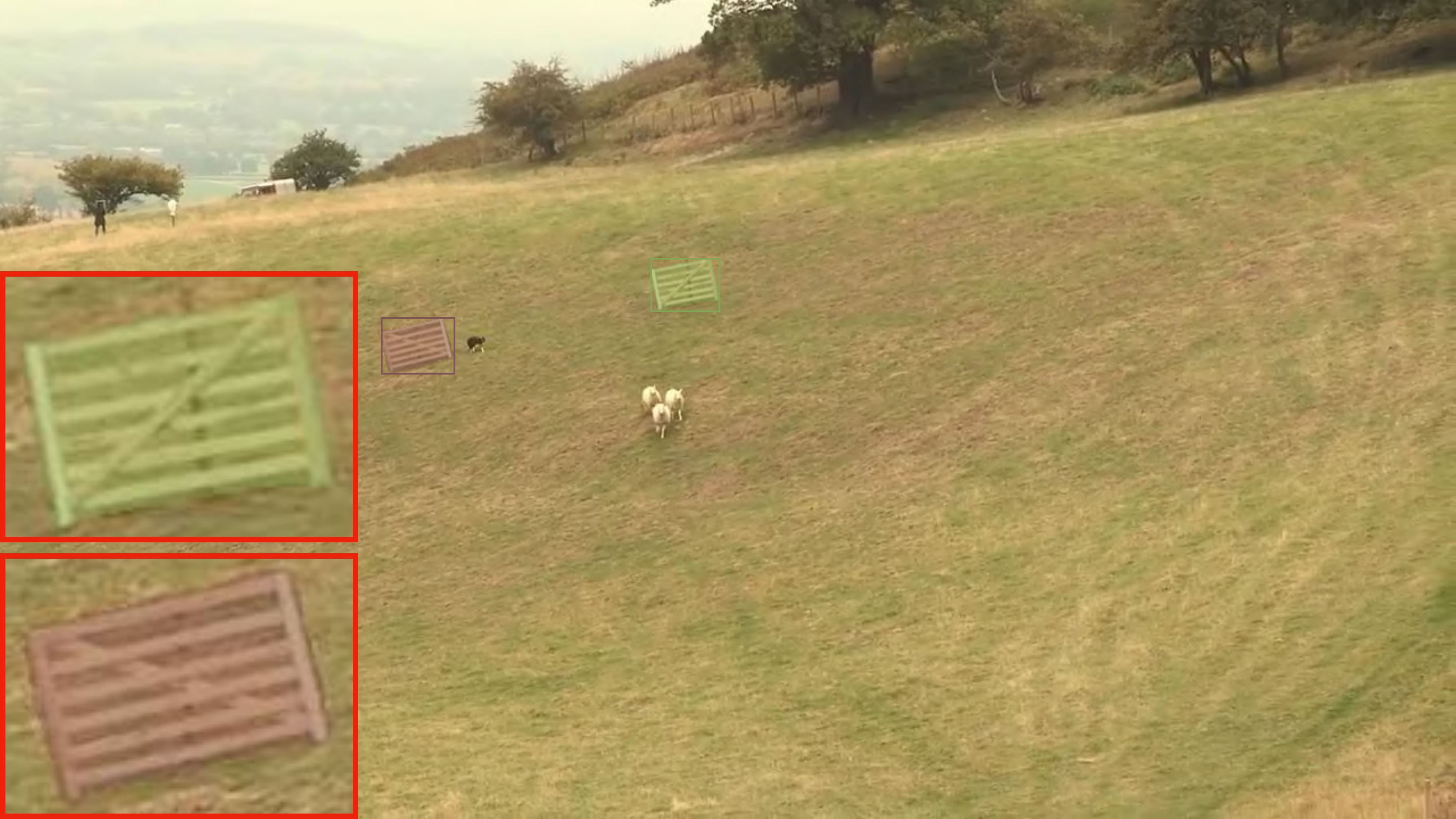}
        \vspace{-16pt}
    \end{subfigure}
    \vspace{5pt}
    
    \begin{subfigure}[b]{0.32\linewidth}
        \includegraphics[width=\mysize\linewidth]{images/additional_non_gt/LaSOT_swing-20/00000901.pdf}
        \vspace{-16pt}
    \end{subfigure}
    \begin{subfigure}[b]{0.32\linewidth}
        \includegraphics[width=\mysize\linewidth]{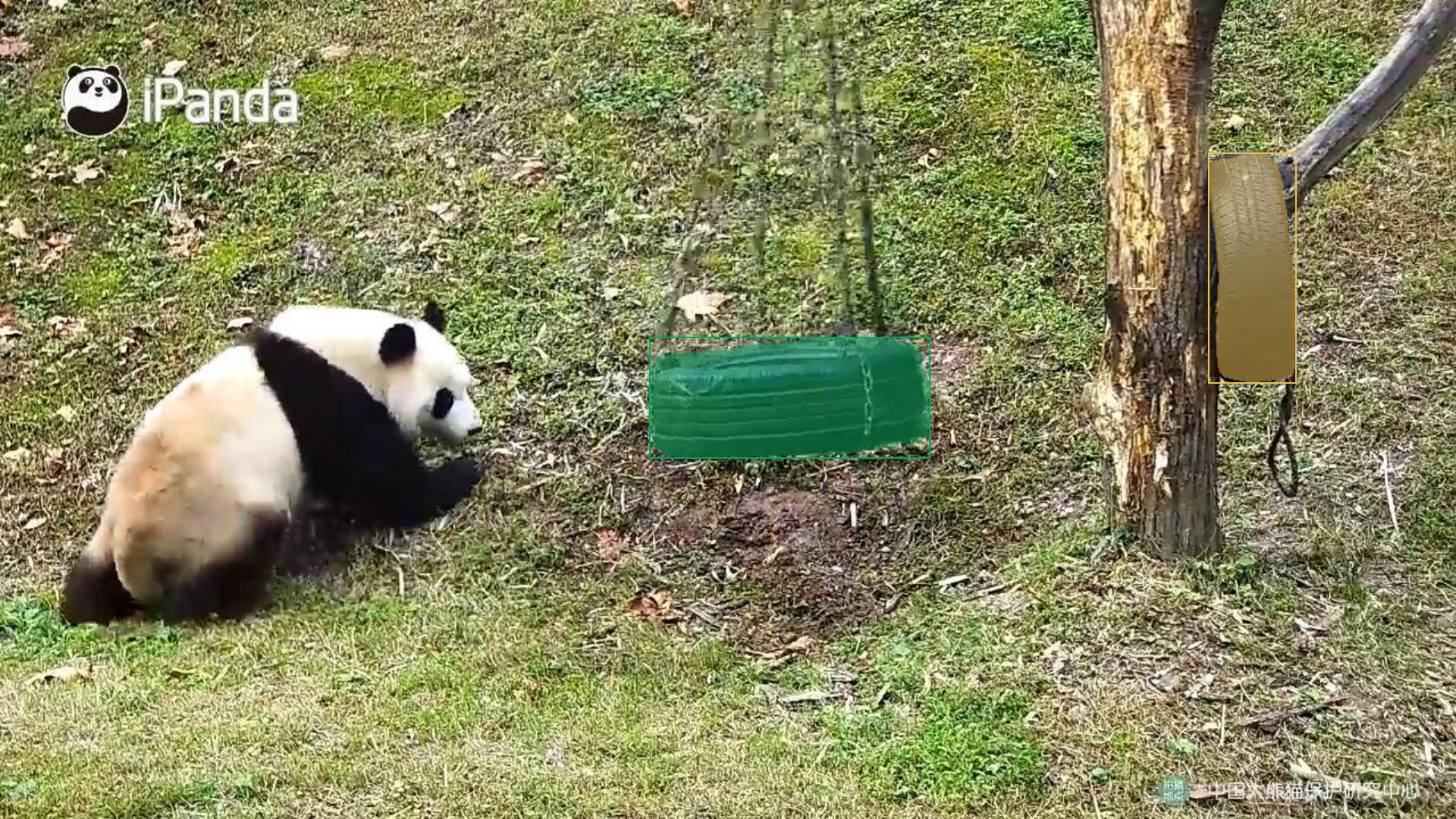}
        \vspace{-16pt}
    \end{subfigure}
    \begin{subfigure}[b]{0.32\linewidth}
     \includegraphics[width=\mysize\linewidth]{images/additional_non_gt/LaSOT_swing-20/00000991.pdf}
        \vspace{-16pt}
    \end{subfigure} 
    \vspace{5pt}
    
    \begin{subfigure}[b]{0.32\linewidth}
        \includegraphics[width=\mysize\linewidth]{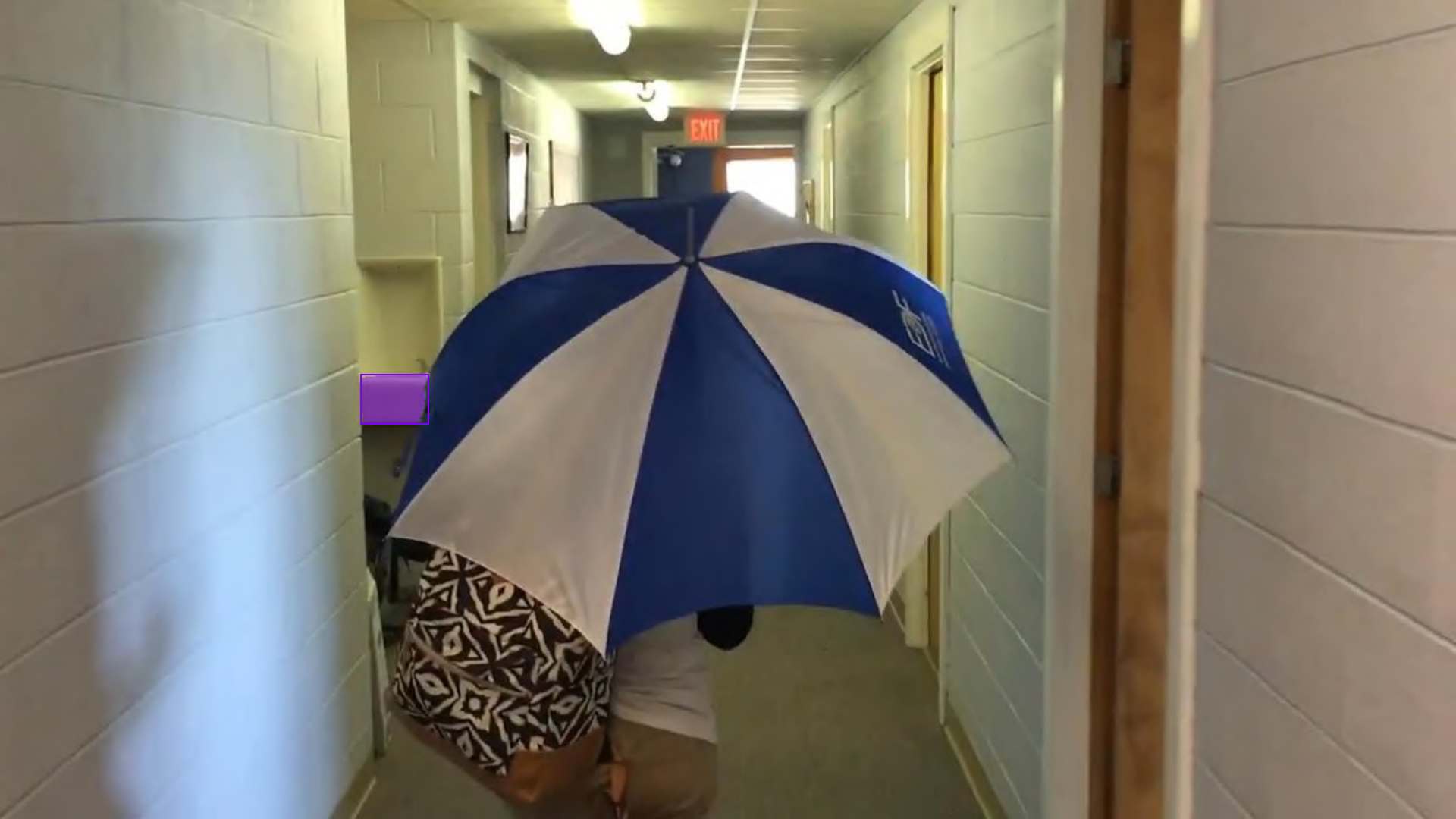}
        \vspace{-16pt}
    \end{subfigure}
    \begin{subfigure}[b]{0.32\linewidth}
        \includegraphics[width=\mysize\linewidth]{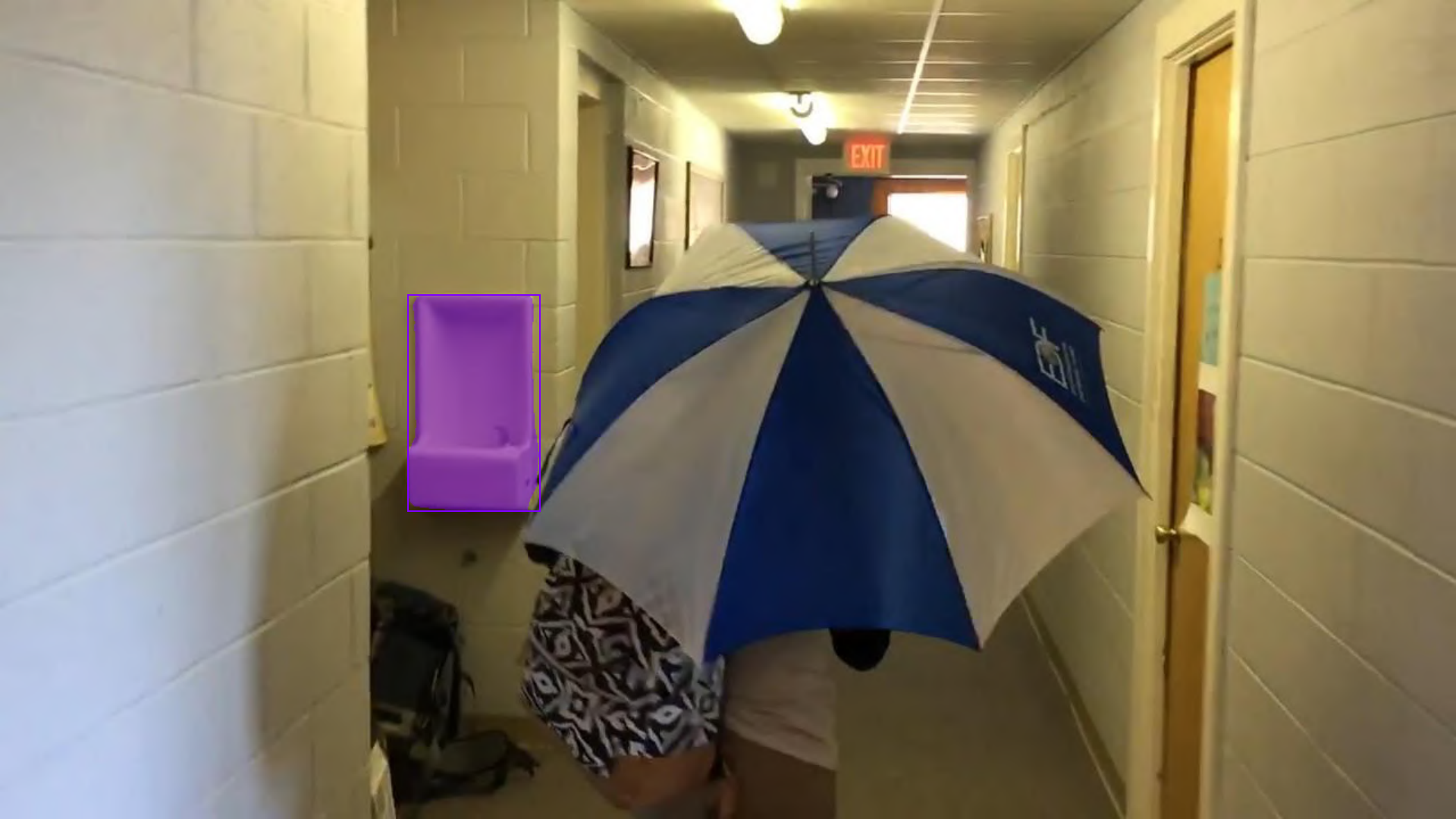}
        \vspace{-16pt}
    \end{subfigure}
        \begin{subfigure}[b]{0.32\linewidth}
        \includegraphics[width=\mysize\linewidth]{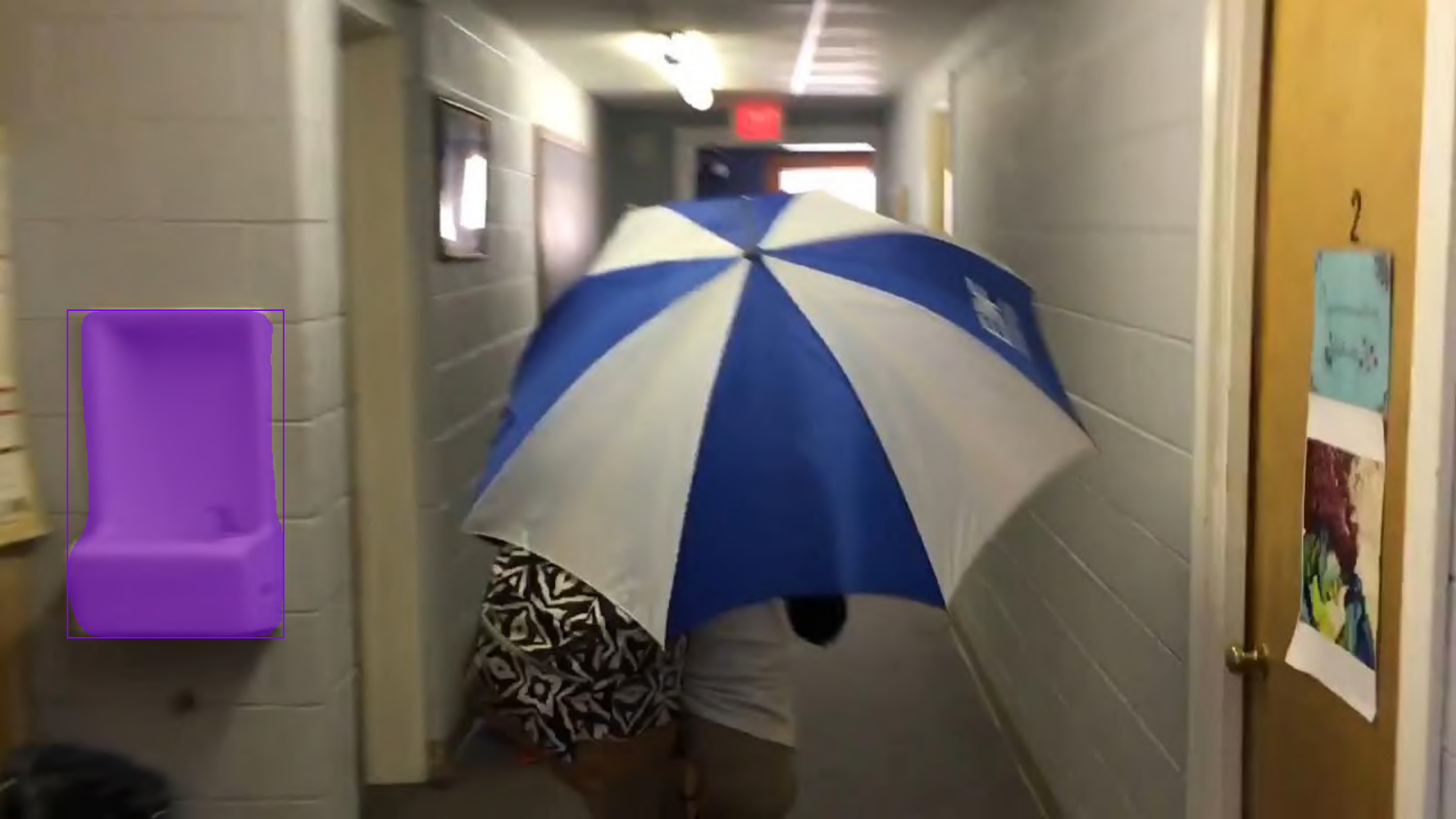}
        \vspace{-16pt}
    \end{subfigure}
    \vspace{5pt}
    
    \begin{subfigure}[b]{0.32\linewidth}
        \includegraphics[width=\mysize\linewidth]{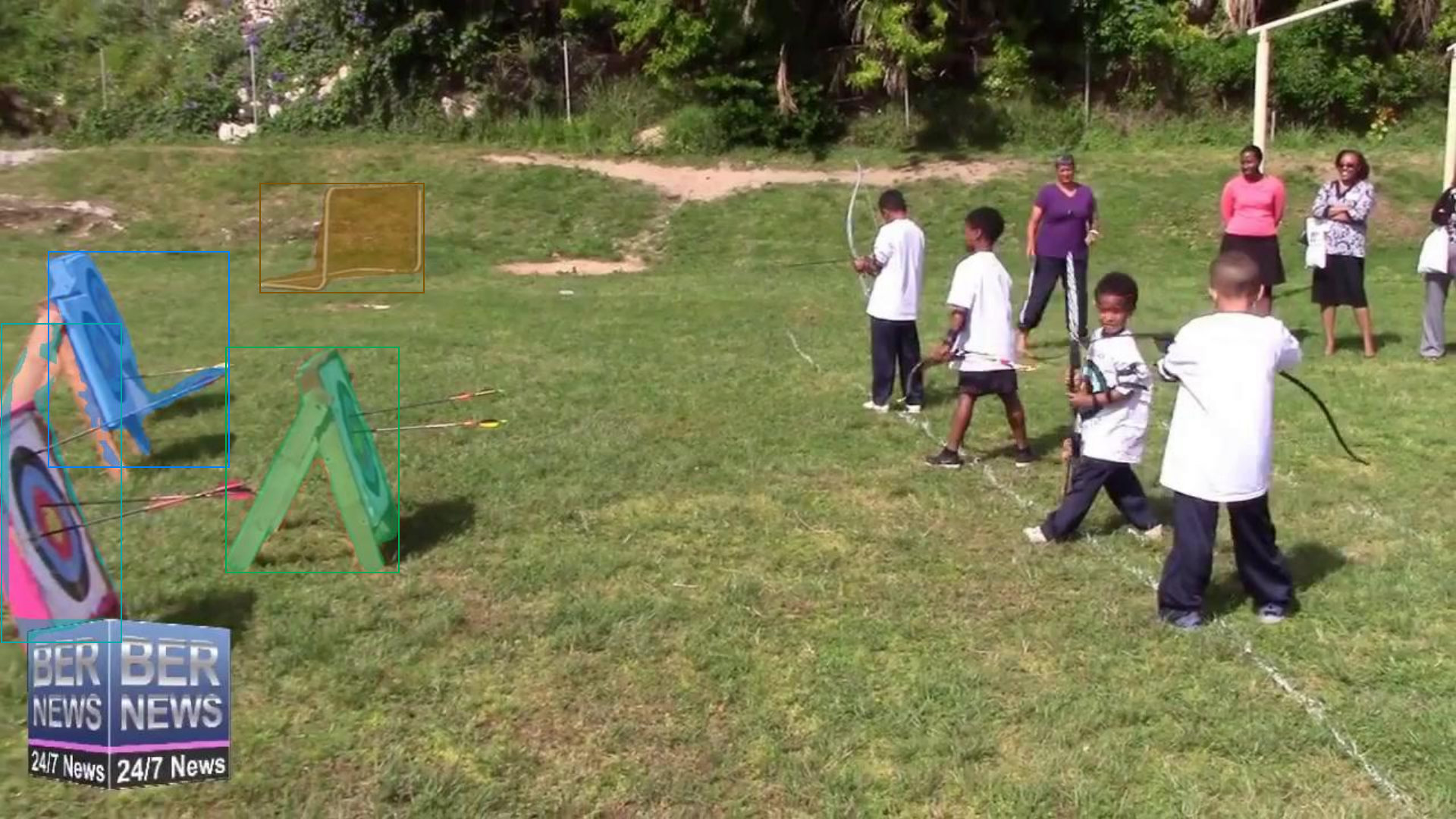}
        \vspace{-16pt}
    \end{subfigure}
    \begin{subfigure}[b]{0.32\linewidth}
        \includegraphics[width=\mysize\linewidth]{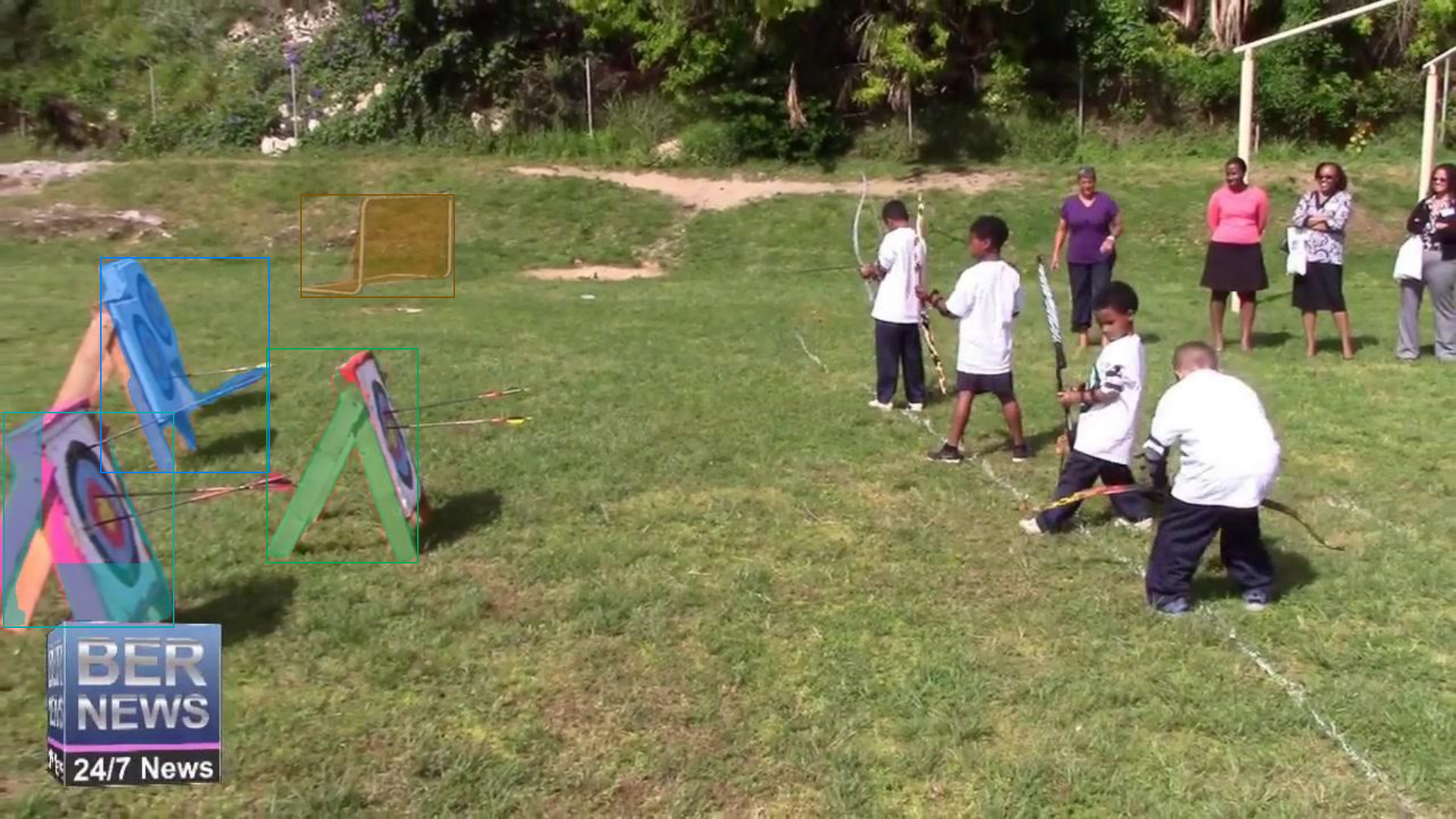}
        \vspace{-16pt}
    \end{subfigure}
        \begin{subfigure}[b]{0.32\linewidth}
        \includegraphics[width=\mysize\linewidth]{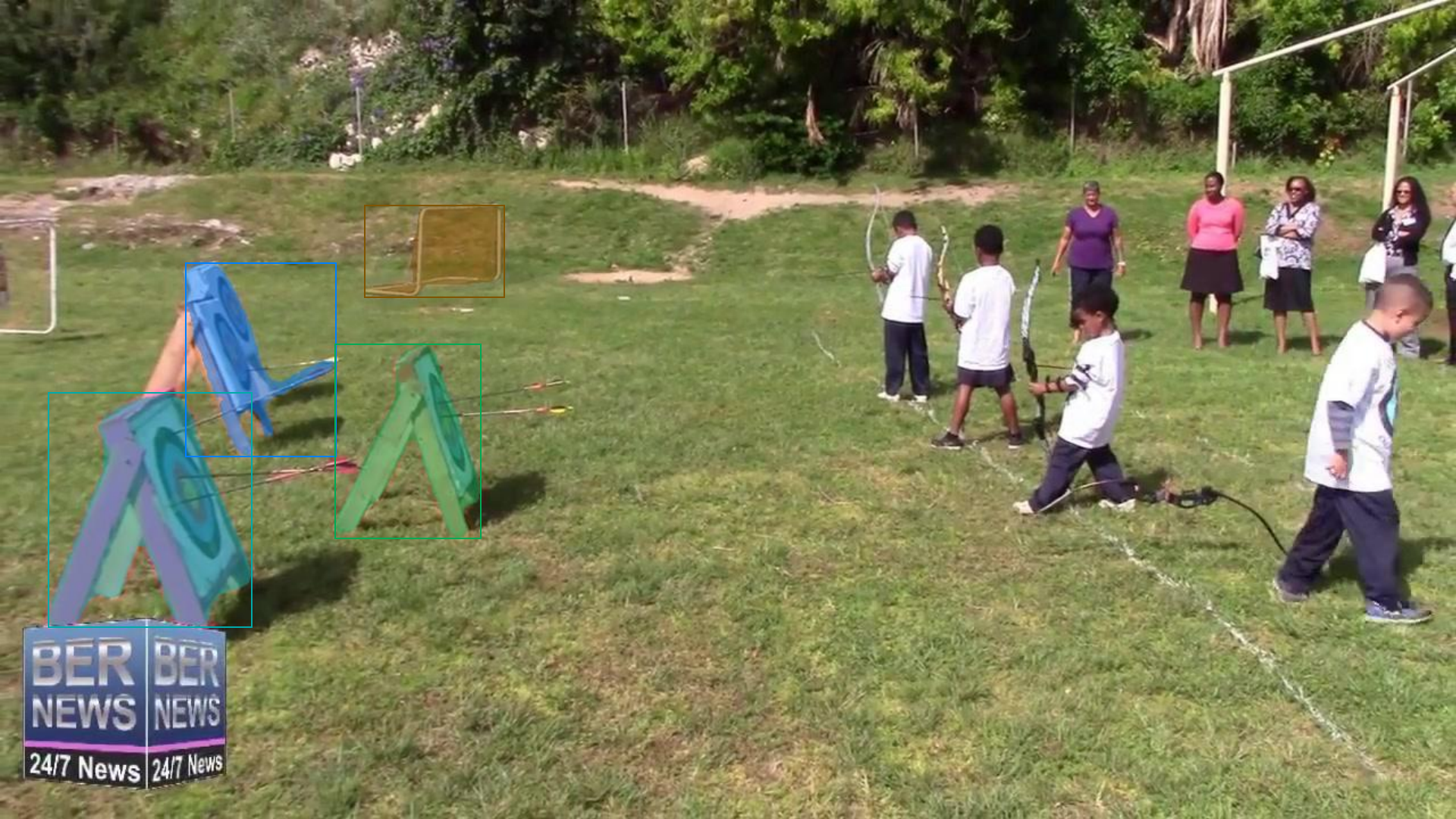}
        \vspace{-16pt}
    \end{subfigure}
    \vspace{5pt}
    
    \begin{subfigure}[b]{0.32\linewidth}
        \includegraphics[width=\mysize\linewidth]{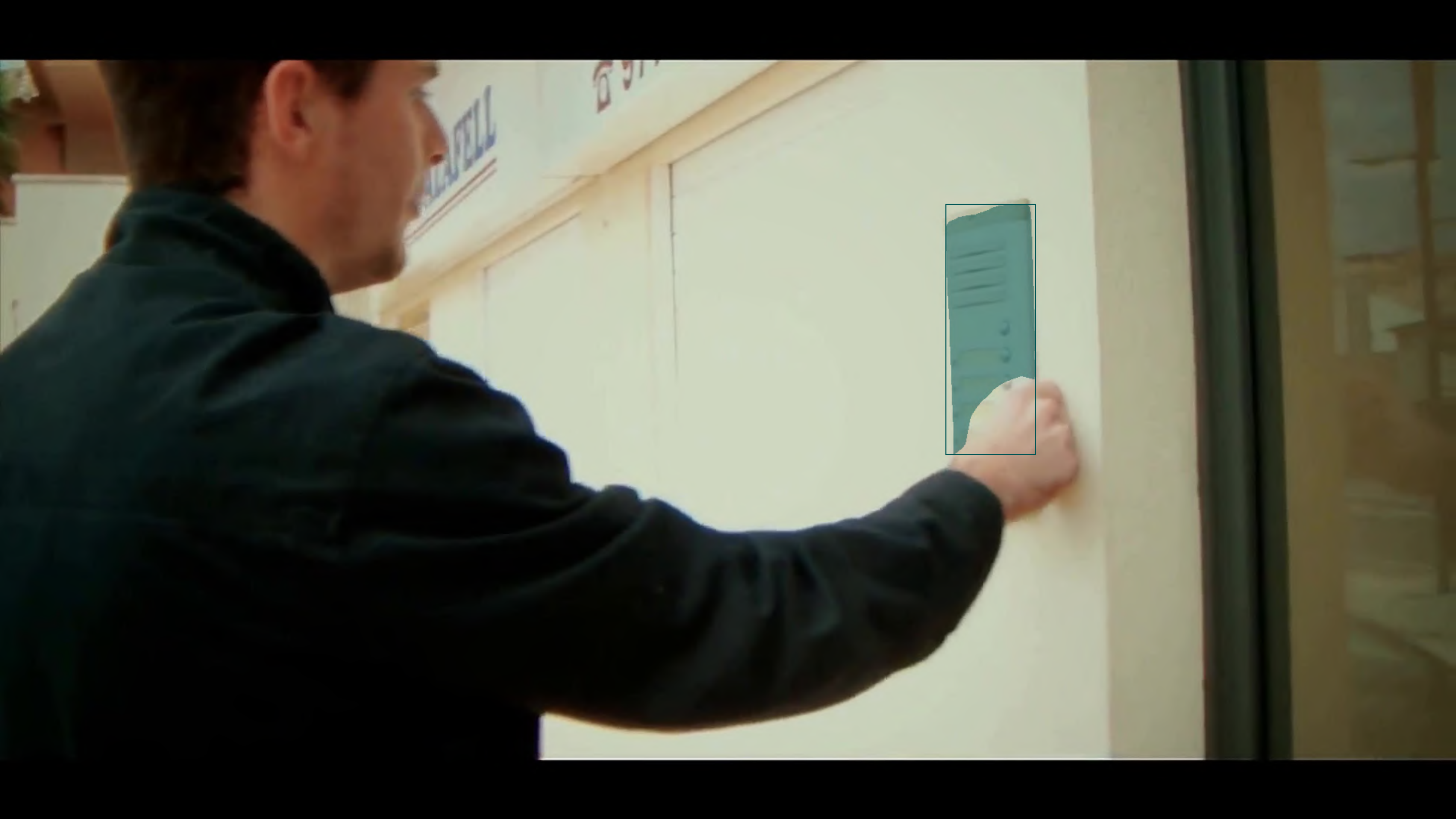}
        \vspace{-16pt}
    \end{subfigure}
    \begin{subfigure}[b]{0.32\linewidth}
        \includegraphics[width=\mysize\linewidth]{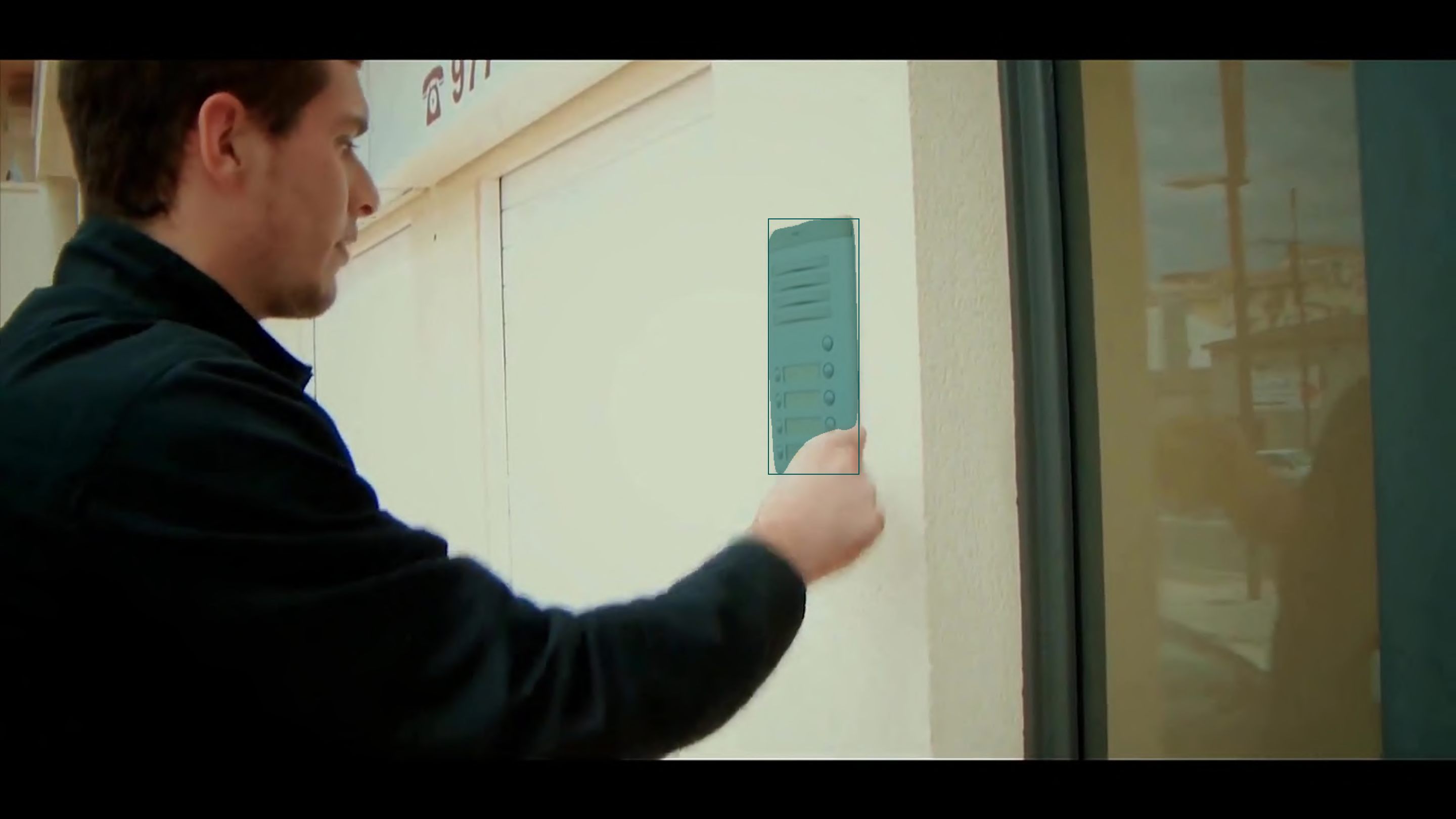}
        \vspace{-16pt}
    \end{subfigure}
        \begin{subfigure}[b]{0.32\linewidth}
        \includegraphics[width=\mysize\linewidth]{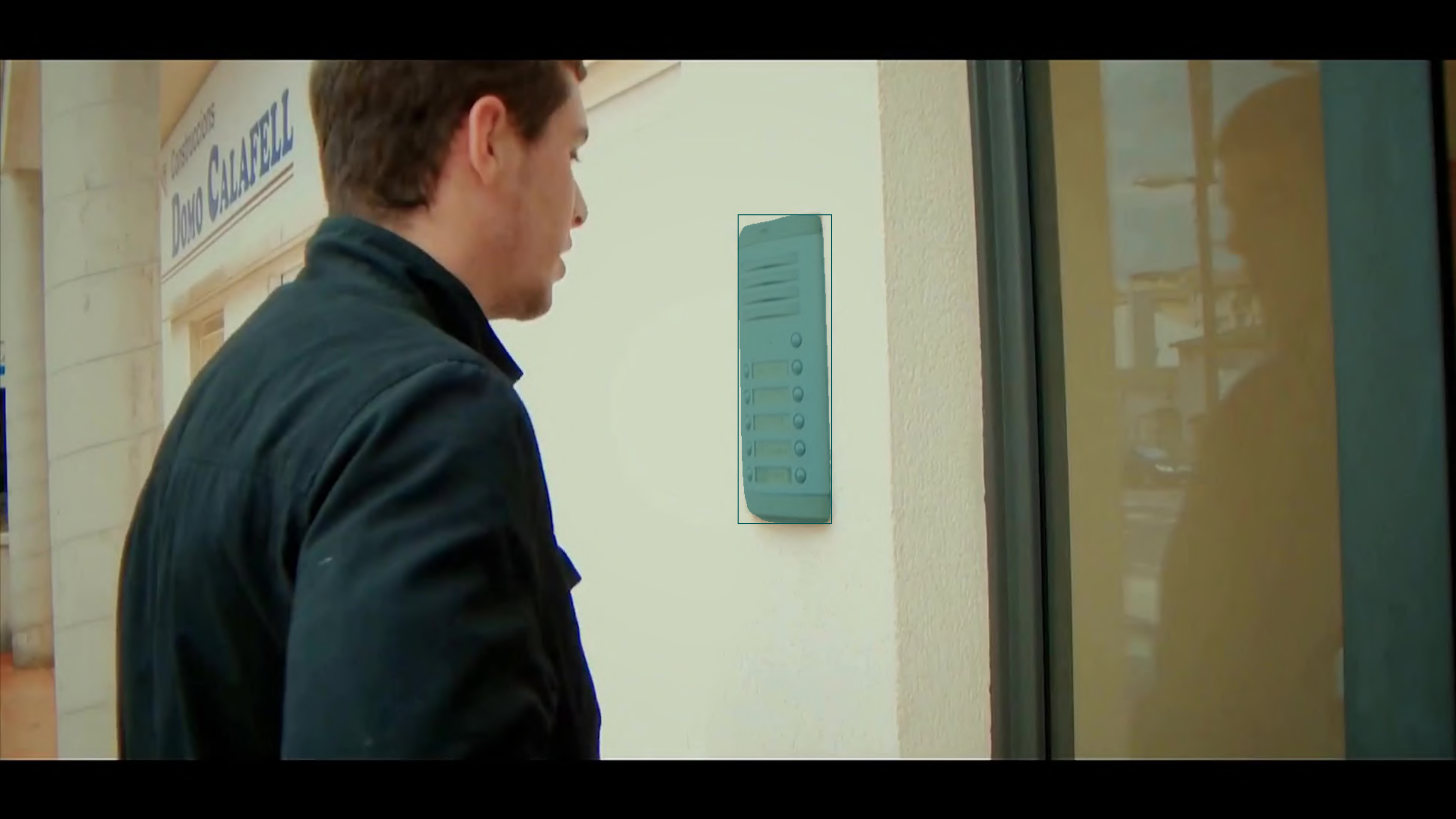}
        \vspace{-16pt}
    \end{subfigure}
    \vspace{5pt}
    
    \caption{\textbf{Tracking results for \textit{unknown unknowns}.} Examples of unlabeled objects outside of the TAO~\cite{dave20ECCV} vocabulary which are correctly tracked by OWTB. OWTB performs well even for small objects (\textit{second} and \textit{third row}).}
    \label{fig:additional_non_gt_tracking}
\end{figure*}

\end{document}